\documentclass[draft]{agujournal2019}
\usepackage{url} 
\usepackage{lineno}
\usepackage[inline]{trackchanges} 
\usepackage{soul}
\usepackage{booktabs} 
\usepackage{tabularx}  
\usepackage{amsfonts}  
\usepackage[authoryear, round]{natbib}
\usepackage{titlesec}
\usepackage[fleqn]{amsmath}
\usepackage{lmodern}
\usepackage[breaklinks=true]{hyperref}
\usepackage{makecell}
\usepackage{threeparttable}
\usepackage[table]{xcolor}
\definecolor{groupbg}{gray}{0.92}
\usepackage[table]{xcolor}
\newcommand{\grouprule}{\arrayrulecolor{black!30}\midrule[0.3pt]\arrayrulecolor{black}}
\usepackage{amssymb}

\usepackage{array}
\newcolumntype{P}[1]{>{\raggedright\arraybackslash}p{#1}}

\titleformat{\subsection}
  {\normalfont\normalsize\bfseries}{\thesubsection}{1em}{}
\titlespacing*{\subsection}
  {0pt}{1.5ex plus .5ex minus .2ex}{0.5ex plus .1ex}
\titlespacing*{\subsubsection}
  {0pt}{1.5ex plus .5ex minus .2ex}{0.5ex plus .1ex}  

\draftfalse

\journalname{Journal of Advances in Modeling Earth Systems (JAMES)}

\usepackage{fancyhdr}
\usepackage{graphics}
\usepackage{xcolor}

\makeatletter
\def\@oddhead{}\def\@evenhead{}
\makeatother

\begin{document}

\title{WeatherDreamer: Learning a World Model from Multimodal Earth Observations}

\authors{Kate Duffy\affil{1}\correspondingauthor{Kate Duffy}{kate@myzeus.ai},
Elise Penn\affil{1},
Tristan Ballard\affil{1},
August Posch\affil{1},
Thomas J. Vandal\affil{1}
}

\affiliation{1}{Zeus AI, Inc., Cambridge, Massachusetts}

\begin{keypoints}
\item A world model learns to forecast weather and corrects itself hourly as observations arrive using data from 18 observing systems and no reanalysis 
\item Surface temperature, humidity, and wind forecasts beat the US global ensemble out to 12 hours and estimate uncertainty more reliably than leading ensembles
\item The model infers rainfall, solar radiation and cloud top height from other observations and removes satellite viewing angle effects
\end{keypoints}

\begin{abstract}
Machine learning weather models now rival operational global forecasts but still start from analyses produced by separate, expensive data assimilation (DA) systems, which for global models are typically updated every six hours. Observation-to-observation models avoid this dependence, but most omit the recursive filtering cycle that lets DA systems carry a background state through sparse, irregular observations. We introduce WeatherDreamer, a world model that learns this cycle end-to-end from 18 global observing systems, without reanalysis. It infers a latent state from observations, carries it forward hour by hour with learned dynamics, corrects it each hour with new observations, and can predict what any of its observing systems would measure at any location. At held-out stations, its ensembles beat NOAA's Global Ensemble Forecast System (GEFS) for temperature, dew point and wind out to 12 h and are better calibrated than GEFS and the ECMWF ensemble. Precipitation, cloud-top height and solar irradiance, never inputs at inference, are inferred with skill matching or exceeding operational products at short leads. Against the first guess of NOAA's Gridpoint Statistical Interpolation (GSI) DA system, WeatherDreamer's short-range departures are within 10\% of GSI's at surface stations and within 25\% for moisture-sensitive microwave channels, but up to five times larger for temperature-sounding channels. Never given observation errors, its learned update assumes errors close to known instrument and retrieval errors of microwave sounders and scatterometer winds. WeatherDreamer also separates viewing geometry from atmospheric state, recovering 83-96\% of observed limb darkening and producing seamless nadir-view imagery. Producing an hourly global analysis in about 1.5 minutes on one GPU, WeatherDreamer shows that assimilation and forecasting can be learned together directly from observations.
\end{abstract}

\section*{Plain Language Summary}
Weather forecasts start from a best estimate of the current state of the atmosphere, made by combining a short forecast with new measurements from satellites, weather balloons and surface stations. These systems are expensive to build and run, and global estimates are usually updated only every six hours. We developed WeatherDreamer, a machine learning model that learns this process directly from observations: every hour it predicts how the atmosphere will change, then corrects that prediction with the measurements taken during that hour.
It learns from 18 types of instruments, without relying on existing weather analyses or hand-built rules about how much to trust each one. At weather stations withheld from the model, its forecasts of temperature, humidity and wind over the next 12 hours were better than those of the US global forecast ensemble, with more realistic estimates of their own uncertainty. It also estimated rainfall, sunlight reaching the ground and cloud heights without receiving them as inputs, and learned to remove the distortion satellites see when viewing Earth at an angle. It came close to a leading DA system near the ground but was less accurate higher in the atmosphere. Learning directly from observations could make more frequent weather updates faster and cheaper.

\section{Introduction}
\label{sec:into}
\noindent 
Machine learning weather prediction (MLWP) models routinely match or exceed operational Numerical Weather Prediction (NWP) skill at a fraction of the computational cost  \citep{lam2023learning, bi2023accurate, lang2024aifs, Price2025}.  
Most, however, remain forecast-only engines that rely on analyses produced by separate data assimilation (DA) systems. 
DA combines a short-range model forecast (the background) with new observations, weighted by their estimated error, to produce the best estimate of the current atmospheric state from which the next forecast is launched.
Since observations are sparse and often indirect (satellites, for example, measure radiances rather than temperature or humidity), DA is essential for turning them into a complete initial state, and advances in DA have driven much of the improvement in forecast skill over recent decades \citep{r2015quiet}. 

Operational DA systems are expensive to develop and run.
Every observation type needs a hand-built observation operator, bias correction, and error model \citep{bannister2017review}, observations are thinned heavily, partly because correlated errors are hard to represent \citep{liu2002interaction}, and each cycle requires either repeated tangent-linear and adjoint integrations (4D-Var; \citealp{rabier2000ecmwf, rawlins2007met}) or an ensemble of short-range forecasts \citep{Kleist_Ide_2015_Part1, Kleist_Ide_2015_Part2}. 
Global analyses are therefore typically produced every 6\,h and arrive hours late.
Regional systems such as the Rapid Refresh and HRRR cycle hourly to support nowcasting and short-range forecasting \citep{benjamin2016rapid, dowell2022hrrr},  but only over limited domains.
WeatherNext3 \citep{weathernext3_2026} ingests raw satellite observations hourly but still initializes from an NWP analysis that lags them by nearly 6\,h.

Several end-to-end systems forecast directly from raw observations, without NWP dependence at inference.
Aardvark \citep{allen2025end} encodes observations into a gridded state, with its encoder and processor pre-trained on ERA5, while EarthNet \citep{vandal2025global}, GraphDOP
\citep{alexe2024graphdop}, AIFS-DOP \citep{pinnington2026aifs} and the
observation-native world model Earth-o1 \citep{gong2026eartho1} map
observations to future states with no reanalysis in training or inference.
These models, however,  build each forecast from a fixed window of observations in a single forward pass, discarding the recursive filtering that lets DA carry a background state through sparse, delayed or missing observations; \citet{alexe2024graphdop} note that a recurrent prior could help for this reason.

A second line of work keeps the recursive filtering but requires reanalysis for learning.
FengWu-4DVar \citep{xiao2023fengwu} embeds the FengWu forecast model in a 4D-Var cost function but assimilates observations simulated from ERA5.
FuXi Weather \citep{Sun2025FuXiWeather} and FuXiWeather2 \citep{xu2026fuxiweather2} cycle learned assimilation and forecasting on real observations, but their forecast and assimilation models are trained on ERA5.
Training against an analysis rewards a system for reproducing the reanalysis model's biases along with its information.

\begin{figure}[t!]
  \centering
  \includegraphics[width=\textwidth]{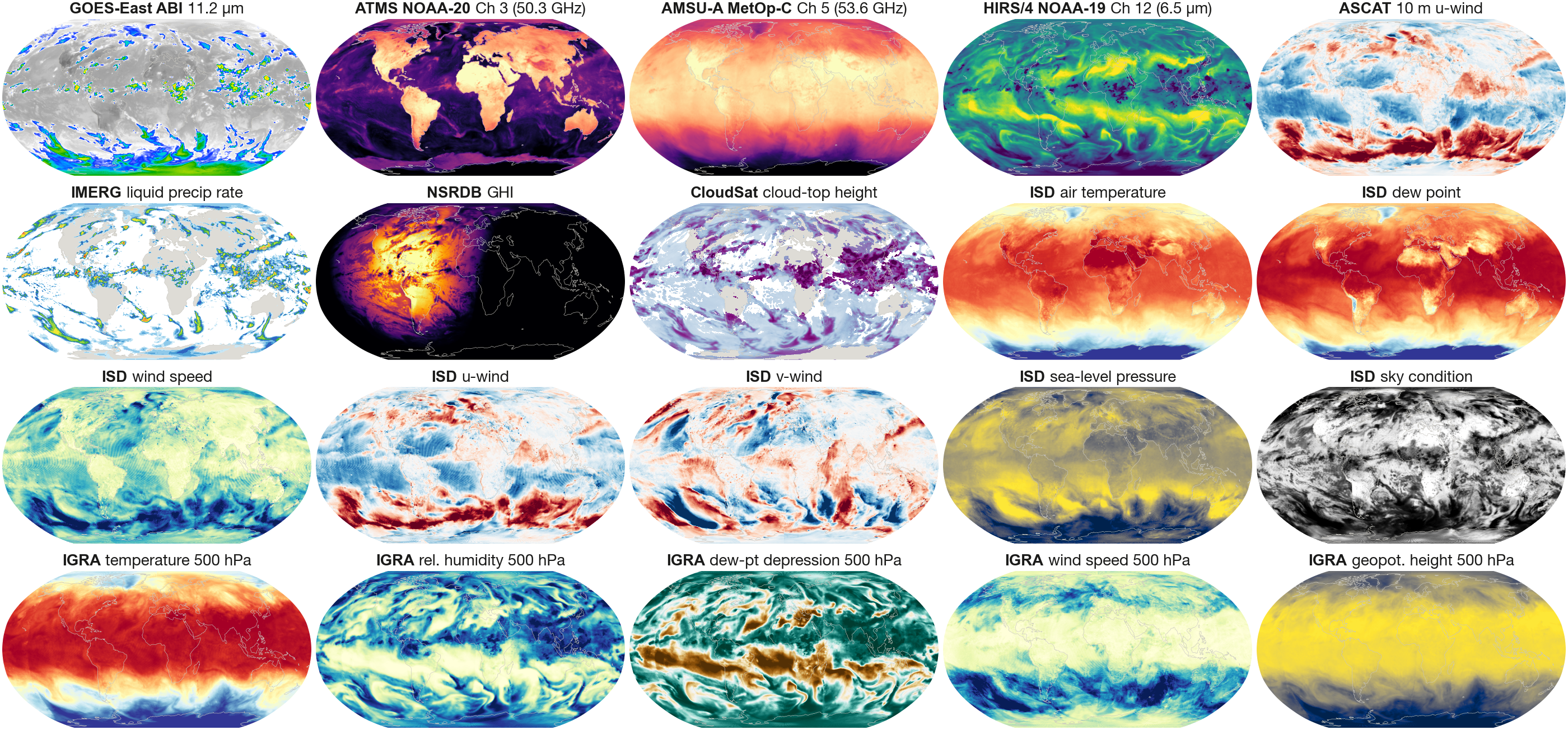}
  \caption{
\textbf{Reconstructed physical variables decoded from a unified, compact latent representation.} 
WeatherDreamer maps diverse, highly heterogeneous global observation streams into a low-dimensional shared latent space. 
Lightweight coordinate decoders generate full, gap-filled global fields across all modalities. 
Surface station networks (ISD) and radiosonde profiles (IGRA) are learned directly from sparse, irregular point measurements and smoothly decoded into continuous space-time representations without requiring spatial grid interpolation.
}
  \label{fig:outputs}
\end{figure}

In this work, we introduce \textbf{WeatherDreamer}, a generative world model that learns a data assimilation and forecast cycle directly from observations.
Following the recurrent state-space models developed for model-based reinforcement learning \citep{hafner2019dream, hafner2020mastering, hafner2023mastering}, WeatherDreamer maintains a learned latent belief state that is predicted forward and corrected by new observations each hour.
The model operates in a latent space from which any of its observation types can be predicted at arbitrary locations, times and viewing geometries (Fig. \ref{fig:outputs}). 
Trained jointly on 18 observing systems, it learns how much to trust each observation, how viewing geometry affects each instrument and how instruments relate to one another, without prescribed error covariances, hand-built observation operators or bias corrections. 
An ensemble CRPS objective \citep{Lang2024crps, alet2025skill} yields probabilistic forecasts.

We evaluate WeatherDreamer as an end-to-end assimilation and forecasting system, entirely against observations, with operational DA and ensemble forecast models as baselines.
At held-out stations, its ensembles beat NOAA's
Global Ensemble Forecast System (GEFS) for near-surface temperature, dew point and wind out to 12\,h and are better calibrated than both GEFS and the ECMWF ensemble (IFS ENS; Section \ref{sec:isd}). 
Precipitation, solar irradiance and cloud-top height, not provided as inputs, are inferred with skill matching or exceeding operational baselines at short leads (Section \ref{sec:unobserved}).
The learned assimilation cycle weights observations without prescribed error models, and its short-range departures approach those of a leading DA system at the surface and for humidity-sensitive channels while lagging for temperature sounding (Section \ref{sec:departures}). 
Finally, the model separates viewing geometry from atmospheric state, yielding seamless nadir-view imagery across satellites (Section \ref{sec:limb}).

\begin{table*}[!t]
\centering
\footnotesize
\setlength{\tabcolsep}{4pt}
\renewcommand{\arraystretch}{1.15}
\caption{\textbf{Observing systems used by WeatherDreamer.} All streams are binned to hourly slots on a HEALPix grid at the listed $N_\mathrm{side}$ (128\,$\approx$\,51\,km, 256\,$\approx$\,25\,km, 512\,$\approx$\,12.7\,km) and patch-embedded onto the model's $N_\mathrm{side}{=}128$ latent grid with the given patch size. \emph{Encoding}: Tok., pre-compressed with a frozen spherical VAE (Text S3); Grid, raw values patch-embedded directly; Points, one token per observation (per level for profiles) at its native coordinates. $T_b$, brightness temperature; $z$, geopotential height.}
\label{tab:datasets}
\begin{tabularx}{\linewidth}{@{}
  l
  >{\hsize=0.85\hsize\raggedright\arraybackslash}X
  >{\hsize=1.15\hsize\raggedright\arraybackslash}X
  l c l@{}}
\toprule
System & Platforms & Variables (no. channels) & Native sampling & $N_\mathrm{side}$ & Encoding \\
\midrule
\multicolumn{6}{@{}l}{\textit{Geostationary imagers}} \\
ABI, AMI     & GOES-16/19, GOES-18, GK2A  & IR $T_b$, 3.9--13.3\,\textmu m (10)              & 2\,km, 10\,min    & 512 & Tok., $4{\times}4$ \\
SEVIRI       & Meteosat (0\textdegree)    & IR $T_b$, 1.6--13.4\,\textmu m (9)               & 3\,km, 15\,min    & 512 & Tok., $4{\times}4$ \\
\grouprule
\multicolumn{6}{@{}l}{\textit{Polar-orbiting sounders}} \\
ATMS         & NOAA-20, S-NPP             & MW $T_b$ (22)                                    & 16--75\,km swath  & 256 & Grid, $2{\times}2$ \\
AMSU-A       & MetOp-B/C                  & MW $T_b$ (12)                                    & 48\,km swath      & 256 & Grid, $2{\times}2$ \\
HIRS/4       & NOAA-18/19                 & IR $T_b$ (18)                                    & 10\,km swath      & 128 & Grid, $1{\times}1$ \\
\grouprule
\multicolumn{6}{@{}l}{\textit{Other satellite}} \\
ASCAT        & MetOp-A/B/C                & 10\,m $u$, $v$ over ocean (2)                    & 12.5\,km swath    & 512 & Grid, $4{\times}4$ \\
GPS-RO       & MetOp-B/C (GRAS)           & Bending angle, 0--60\,km (2)                     & $\le$247 levels   & --  & Points \\
CPR, CALIOP  & CloudSat, CALIPSO          & Cloud-top height, cloud mask (2)                 & 1.1\,km, 2020     & --  & Points \\
\grouprule
\multicolumn{6}{@{}l}{\textit{In situ}} \\
ISD          & SYNOP, METAR               & $T_\mathrm{2m}$, $T_{d,\mathrm{2m}}$, $p_\mathrm{msl}$, 10\,m $u$, $v$, speed, sky cover (7) & Irregular, hourly & -- & Points \\
IGRA         & Radiosondes                & p, $T$, RH, $T{-}T_d$, $z$, $u$, $v$, wind speed, direction (8) & 00/12\,UTC & -- & Points ($\log p$) \\
\grouprule
\multicolumn{6}{@{}l}{\textit{Gridded products}} \\
IMERG V07B   & GPM                        & Liquid precip.\ rate (1)                         & 0.1\textdegree, 30\,min & 256 & Grid, $2{\times}2$ \\
NSRDB v4     & GOES (Americas)            & GHI, DNI, DHI (3)                                & 2\,km             & 512 & Grid, $4{\times}4$ \\
\bottomrule
\end{tabularx}
\end{table*}

\section{Datasets}
\label{sec:datasets}
\subsection{Observing Systems}
\label{sec:obs}

\noindent WeatherDreamer is trained on observations from 18 observing systems and two derived gridded products (Table \ref{tab:datasets}). 
The systems differ in sampling geometry, from full-disk geostationary imagery every 10 to 15 minutes to polar-orbiting swaths that observe a given location roughly twice daily, irregular surface networks (about 11,000 ISD stations), and radiosonde profiles twice daily (about 650 soundings at each of 00 and 12 UTC). 
All observations serve as both inputs and reconstruction targets, enabling any-to-any translation among modalities \citep{mizrahi20234mmassivelymultimodalmasked}.
However, CloudSat, IMERG, and NSRDB (Americas only) are always withheld as inputs at inference time and inferred from other variables only.
Additionally, CloudSat is available for only 2020 of our training period.
All datasets are placed on global HEALPix \citep{gorski2005healpix} grids at resolutions near their native sampling, on a common hourly time base (Text S1).

\subsection{Forcing Fields}
\label{sec:forcings}
\noindent In addition to observations, the model receives fields that are known in advance. 
Solar geometry (zenith and azimuth angles) and clear-sky global horizontal irradiance are computed analytically at each grid cell and hour (Text S2).
The solar fields are encoded as a unit direction vector concatenated with normalized clear-sky irradiance, giving four smooth channels that, together with surface elevation taken from the Copernicus 90 m digital elevation model (COP90), form the forcing $\mathbf{f}_t$.

\subsection{Verification Datasets and Baselines}
\label{sec:evaldata}
\noindent We verify WeatherDreamer's analyses and forecasts against observations and
compare them with operational systems, listed in Table S8. 
534 ISD stations (5\% of the network) are excluded from the input at inference and used only for verification.
As an assimilation benchmark we use the Gridpoint Statistical Interpolation (GSI), the analysis component of NCEP's Global Data Assimilation System (GDAS).
GSI is the core DA system behind operational models such as Global Forecast System (GFS), NAM (North American Mesoscale), and RAP (Rapid Refresh).
We use diagnostic files from a 3D-Var coupled reanalysis run, which cycled every 6\,h and assimilated essentially the operational observing system (Text S6.9).
Forecasts are compared with GFS as a deterministic baseline, the GEFS and IFS ENS ensembles, and, for solar irradiance, HRRR.

\section{Model}
\label{sec:method}

\subsection{From Data Assimilation to Latent World Models}
\label{sec:da}
\noindent Classical data assimilation and sequential filtering models in machine learning \citep{hafner2019dream, hafner2020mastering} share the goal of tracking the unobserved state of a dynamical system from noisy, incomplete, and irregularly sampled observations. 
The same problem is posed in predictive-coding accounts of perception, in which the brain maintains an internal model that predicts its sensory input and updates its belief state from the mismatch between prediction and observation \citep{huang2011predictive, friston2010free, clark2013whatever}. 
The global observing system faces a similar structural challenge, where no single instrument observes the full state of the atmosphere.
Each provides a sparse, noisy, and indirect view.
NWP assimilates observations to constrain a physical forecast engine in a predict-correct cycle \citep{kalman1960new, evensen2009data}, while WeatherDreamer learns a similar filtering cycle end-to-end from observations.
We use the term world model in its machine-learning sense: a learned model of an environment that infers a latent state from partial, noisy observations, predicts how that state evolves, and predicts the observations themselves, without access to the true state
\citep{ha2018world, hafner2019dream}.

In DA, methods such as the Kalman Filter or variational frameworks (3D/4D-Var) systematically alternate between a physical \textit{predict} phase and a statistical \textit{correct} phase \citep{courtier1998ecmwf, rabier2000ecmwf}. 
The predict step propagates the previous analysis $\mathbf{x}^a_{t-1}$ with the forecast model $\mathcal{M}$ to give a background $\mathbf{x}^b_t$.
The correct step maps the background into observation space with an observation operator $H$ and weights the departures $\mathbf{y}_t - \mathbf{H}(\mathbf{x}^b_t)$ by prescribed background and observation error covariances $\mathbf{B}$ and $\mathbf{R}$ via the Kalman gain matrix $\mathbf{K}$
\citep{kalman1960new, ide1997unified}:
\begin{align}
  \mathbf{x}^b_t &= \mathcal{M}(\mathbf{x}^a_{t-1}), \label{eq:da_predict}\\
  \mathbf{K}_t &= \mathbf{B}_t \mathbf{H}^\top \left(\mathbf{H} \mathbf{B}_t \mathbf{H}^\top + \mathbf{R}_t\right)^{-1}, \label{eq:da_gain}\\
  \mathbf{x}^a_t &= \mathbf{x}^b_t + \mathbf{K}_t\left(\mathbf{y}_t - \mathbf{H}(\mathbf{x}^b_t)\right). \label{eq:da_update}
\end{align}

\begin{figure}[t!]
\noindent\includegraphics[width=\textwidth]{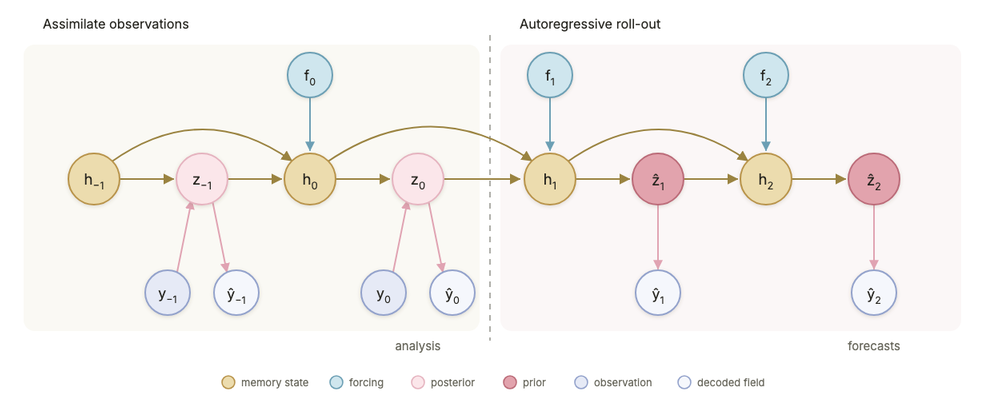}
\caption{
\textbf{Temporal unrolling of the state-space world model.} 
The network tracks atmospheric dynamics across time by updating a hidden memory state ($h_t$) that accumulates history and boundary forcings ($f_t$). 
During \textit{assimilation}, the posterior latent state ($z_t$) is updated by conditioning prior expectations against incoming multi-modal observation streams ($y_t$) to capture the current state estimate.
During \textit{forecasting} latent ensembles are generated by injecting spatially correlated noise into the prior ($\hat{z}_t$) prediction, enabling
Markovian trajectory rollouts conditioned on external forcings.
}
\label{fig:rollout}
\end{figure}

WeatherDreamer replaces each of these components with a learned counterpart within a recurrent state-space model (RSSM) in the style of Dreamer \citep{hafner2019dream, hafner2020mastering, hafner2023mastering}. 
Instead of a physical state, it carries a memory $\mathbf{h}_t$ and a latent state $\mathbf{z}_t$ that holds a unified representation of all observed modalities:
\begin{align}
  \text{Memory:} \quad & \mathbf{h}_t = f_{\mathrm{dyn}}(\mathbf{h}_{t-1}, \mathbf{z}^*_{t-1}, \mathbf{f}_t), \label{eq:dyn}\\
  \text{Prior (background):} \quad & \hat{\mathbf{z}}_t = p_\theta(\mathbf{h}_t), \label{eq:prior}\\
  \text{Posterior (analysis):} \quad & \mathbf{z}_t = q_\phi(\mathbf{h}_t, \mathbf{y}_t), \label{eq:posterior}\\
  \text{Decoder:} \quad & \hat{\mathbf{y}}^m_t = D_m(\mathbf{z}^*_t), \label{eq:decoder}
\end{align}
where $\mathbf{f}_t$ are known forcings (Section \ref{sec:forcings}) and the active state $\mathbf{z}^*_t$ is the posterior $\mathbf{z}_t$ when observations are available and the prior $\hat{\mathbf{z}}_t$ otherwise.
During assimilation, observations correct the state through the posterior; during forecasting or data denial, the prior is fed back and the model rolls forward on its own (Figure \ref{fig:rollout}).
Each time step $t$ spans one hour: $\mathbf{y}_t$ comprises all observations whose valid time falls within $\pm 30$\,min of the step, each retaining its sub-hourly time offset in its embedding (Section~3.3), so the state is updated once per hour regardless of how many observations arrive within it.

The prior plays the role of the background, the posterior that of the analysis, and the decoders $D_m$ that of the observation operator $H$. 
The posterior network takes the place of the Kalman gain and its prescribed $\mathbf{B}$ and $\mathbf{R}$, learning how much to trust each observation relative to the background. 
The memory carries information through gaps in coverage, providing the background state that window-based observation-to-observation models lack.
Unlike classical DA, the state is latent rather than physical, uncertainty is represented by an ensemble rather than an explicit covariance, and observations are aggregated by a learned operator rather than thinned, so the cost of each step does not grow with the number of observations.

\subsection{Architecture Overview}
\begin{figure}[t!]
\noindent\includegraphics[width=\textwidth]{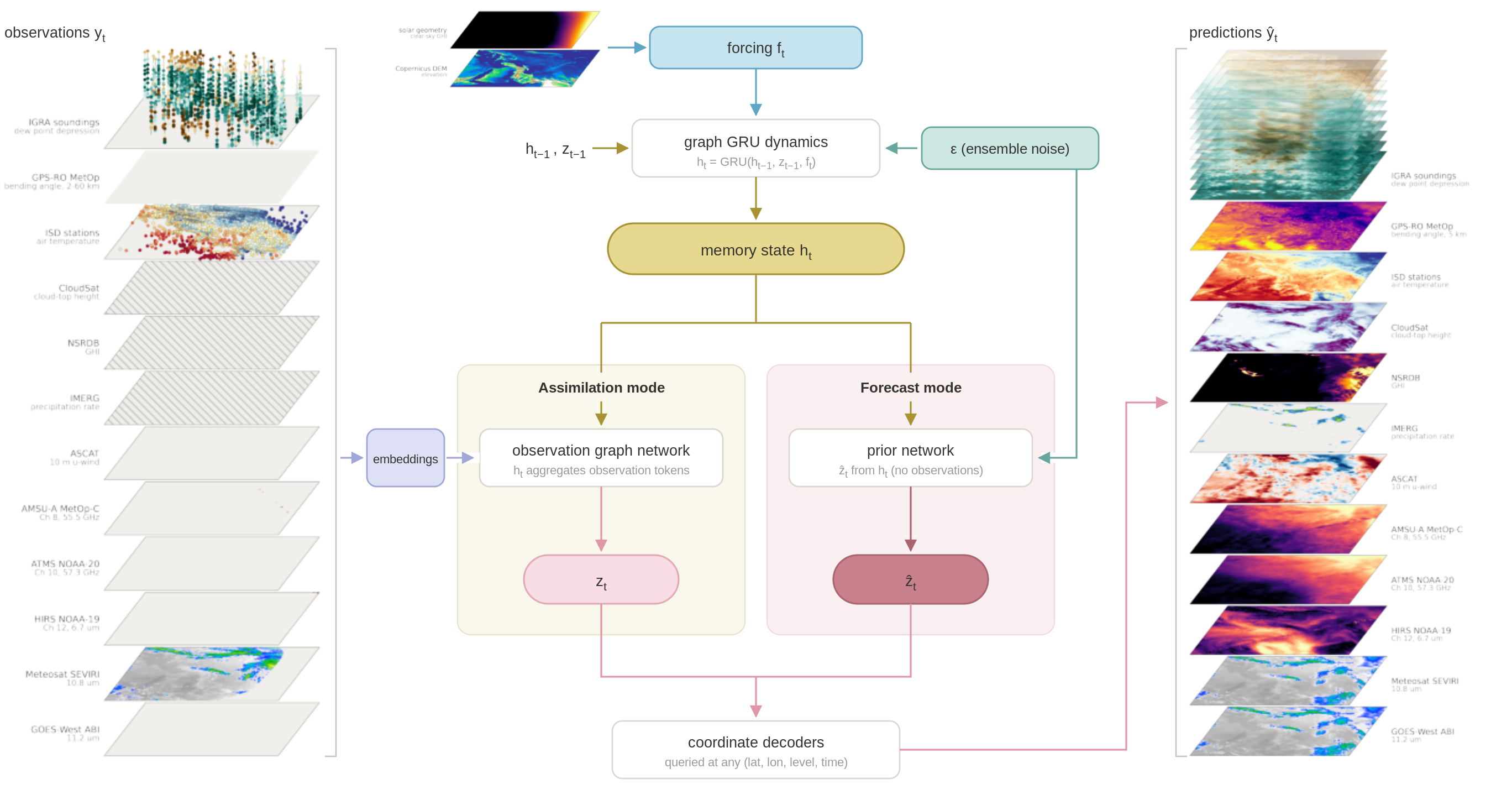}
\caption{
\textbf{Architectural overview and data flow of WeatherDreamer.} 
Multi-modal Earth observations ($y_t$) are embedded and ingested alongside a deterministic memory state $h_t$.
The latent state estimation seamlessly alternates between two modes: assimilation mode, when observations are used to estimate the analysis posterior $z_t$, and forecast mode, where a transition module projects dynamics forward to predict the prior $\hat{z}_t$ from memory and boundary forcing constraints ($f_t$).
Either latent state can be routed to local implicit coordinate decoders to output full physical reconstructions and multi-modal predictions ($\hat{y}_t$) across arbitrary spatial and temporal queries. 
}
\label{fig:overview}
\end{figure}

\noindent Figure \ref{fig:overview} shows how the components of WeatherDreamer implement Eqs. \ref{eq:dyn}--\ref{eq:decoder}.
Observations are converted to tokens with continuous space, time, and modality embeddings (Section \ref{sec:embeddings}).
A graph encoder aggregates tokens onto fixed latent nodes to form the posterior (Section \ref{sec:methods_post}).
Latent nodes are the cells of a HEALPix grid at $N_{\mathrm{side}} = 128$ (about 50\,km spacing), each carrying a memory $\mathbf{h}_t \in \mathbb{R}^{512}$ and a latent state $\mathbf{z}_t \in \mathbb{R}^{64}$. 
The wide memory carries information across many steps, while the narrow latent state limits what each step passes forward (visualized in Figure S3). 
A graph-recurrent cell updates the memory from the previous active state and the forcing and a prior head predicts the latent state from the memory (Section \ref{sec:methods_dyn}).
The model runs at a 1\,h time step on regional patches of $32 \times 32$ latent nodes.
Modality-specific decoders map the active state to any observation type at arbitrary coordinates (Section \ref{sec:decoders}).

\subsection{Encoding Heterogeneous Observations}
\label{sec:embeddings}
\noindent Each observation is mapped to a token in a common  hidden dimension.
Gridded observations are patchified with an invertible convolutional layer with fixed, orthogonal identity weights, so sub-patch structure is preserved before linear projection. 
Geostationary radiances are the exception: they are pre-compressed by frozen modality-specific tokenizers (Text S3; Figure S1), whose latent vectors serve as both inputs and targets. 
Point observations (ISD stations, CloudSat) are ingested as individual tokens at their exact coordinates without rasterization. 
For profiles (radiosondes, GPS-RO), each reported level is a separate token, with its vertical coordinate (normalized log-pressure or height) given a Fourier encoding.
Each token’s projected values are summed with the following embeddings before encoding:

\textit{Position and time:} Continuous coordinates $(t, y, x) \in [-1, 1]^3$ are mapped into a high-frequency harmonic representation  via a bank of sinusoidal projection functions \citep{tancik2020fourier, mildenhall2021nerf}:
\begin{equation}
\mathbf{e}_{\text{pos}}(t, y, x) = \left[ \sin(\boldsymbol{\omega}_k \cdot [t, y, x]), \cos(\boldsymbol{\omega}_k \cdot [t, y, x]) \right]_{k=1}^K
\end{equation}
where $\boldsymbol{\omega}_k$ are log-spaced frequencies.
The time coordinate includes sub-hourly time offsets $\Delta t \in [-0.5, 0.5]$ relative to the model timestep.
The same embedding is used for gridded and point tokens, and it is continuous across HEALPix face boundaries.

\textit{Modality:} A learned embedding $\mathbf{E}_{\text{mod}} \in \mathbb{R}^D$ identifies each observing system so that the encoder can distinguish between sensors and platforms. 

\textit{Viewing angle:} Each satellite token carries its excess air mass $\ell = \sec\theta - 1$, which is zero at nadir and increases continuously toward the limb.
This lets the model separate viewing geometry from atmospheric state.
The decoders are conditioned on $\ell$ in the same way.

\textit{Elevation:} ISD station tokens carry the normalized COP90 elevation at the station location, added through zero-initialized MLPs.

\subsection{Assimilating Observations}
\label{sec:methods_post}

\noindent The posterior network $q_\phi$ maps the observation tokens at time $t$ onto the fixed latent nodes with attention to the memory $h_t$ of what is already known.
It is a bipartite graph from observation tokens to latent nodes, built separately for each modality to handle different sparsity among modalities.
For latent node $i$ and modality $m$, let $\mathcal{N}^m_i$ be the $k = 16$ observation tokens of that modality nearest to the node, measured in the patch's local space–time coordinates, $d^2 = \Delta x^2 + \Delta y^2 + (\lambda \Delta t)^2$ with $\lambda  = 0.25$, so spatial proximity dominates.
Each token $j$ sends a message that depends on its embedding $\mathbf{e}_j$ (Section \ref{sec:embeddings}), its offset $\boldsymbol{\delta}_{ij}$ in space and time from the node, and the node's memory:
\begin{equation}
  \mathbf{m}_{ij} = \mathrm{MLP}_{\mathrm{msg}}
  \left( \left[ \mathbf{e}_j,\; \gamma(\boldsymbol{\delta}_{ij}),\; \mathbf{h}_{t,i} \right] \right),
\end{equation}
where $\gamma(\cdot)$ is the Fourier encoding. Messages are combined using learned attention weights:
\begin{equation}
\begin{aligned}
\alpha_{ij} &= \operatorname{softmax}_{j \in \mathcal{N}_i}\!\left(s_{ij} + b\,\mathbf{1}[j \text{ co-located with } i]\right),
\qquad \mathcal{N}_i = \textstyle\bigcup_m \mathcal{N}_i^m,\\
\mathbf{a}_i &= \sum_{j \in \mathcal{N}_i} \alpha_{ij}\,\mathbf{m}_{ij}.
\end{aligned}
\label{eq:attn}
\end{equation}
where $s_{ij}$ is a learned score computed from $\mathbf{m}_{ij}$ and $b$ is a learnable self-edge bias that lets the node weight an observation in its own cell above its neighbors. 
The neighbor sets of all modalities are pooled, so a single softmax lets modalities compete for weight at each node.
The attention weights depend on the node's memory, so the weight given to each observation can depend on what the model already knows.
The aggregate is augmented with a learned projection of the node's observation coverage (the fraction of valid neighbor slots and the distance to the nearest observation). 
Where a node has no observations of any modality, $\mathbf{a}_i$ is replaced by a learned no-observation embedding $\mathbf{e}_\varnothing$, so the model distinguishes ``unobserved" from a zero value.
The result is added residually to the node state, normalized, and linearly projected to give the posterior.
Message passing costs $O(N k M)$ for N latent nodes and M modalities, independent of the number of observations once neighbors are found; observations are binned to the model grid but are otherwise not thinned.

\subsection{Forecasting with Learned Dynamics}
\label{sec:methods_dyn}
\noindent The dynamics cell $f_{\mathrm{dyn}}$ (Eq. \ref{eq:dyn}) advances the memory by one hour, and the prior head $p_\theta$ (Eq. \ref{eq:prior}) predicts the latent state from the updated memory alone. 
Together they form the forecast model of the system, used at every step during assimilation and as the only update during rollout.

\paragraph*{Inputs.}
At step $t$, each latent node $i$ receives the previous active state $\mathbf{z}^{*}_{t-1,i}$ and the forcing $\mathbf{f}_{t,i}$. 
Each forcing field (Section \ref{sec:forcings}) is patch-embedded to the latent grid, and the results are concatenated and linearly projected to $\mathbf{f}_t \in \mathbb{R}^{[d_f]}$ per node. 
The cell input is $\mathbf{u}_{t,i} = [\mathbf{z}^{*}_{t-1,i}, \mathbf{f}_{t,i}]$. 
Because the forcing is analytic or static, it is available by definition at every step of a rollout.
It conditions the update but is never a reconstruction target.

\paragraph*{Graph-recurrent cell.}
The cell is a gated recurrent unit (GRU) whose linear maps are replaced by
$R = 4$ rounds of message passing $G(\cdot)$ on a graph linking each latent
node to its $k = 8$ nearest neighbors:
\begin{align}
  \mathbf{r}_t &= \sigma\!\left( G_r(\mathbf{u}_t, \mathbf{h}_{t-1}) \right), \qquad
  \mathbf{g}_t = \sigma\!\left( G_g(\mathbf{u}_t, \mathbf{h}_{t-1}) \right), \\
  \tilde{\mathbf{h}}_t &= \tanh\!\left( G_h(\mathbf{u}_t, \mathbf{r}_t \odot \mathbf{h}_{t-1}) \right), \\
  \mathbf{h}_t &= (1 - \mathbf{g}_t) \odot \mathbf{h}_{t-1} + \mathbf{g}_t \odot \tilde{\mathbf{h}}_t ,
\end{align}
The update gate $\mathbf{g}_t$ lets the memory persist through gaps in
coverage, while message passing moves information up to about 200\,km
per hour, faster than typical upper-tropospheric advection.

\paragraph*{Prior head.}
The prior is computed independently at each node by a two-layer MLP: $\hat{\mathbf{z}} = \mathrm{MLP}(\mathbf{h})$.
Because it sees only the memory, the prior is the model's background: its best estimate of the current state before time-$t$ observations are assimilated. 
During training it is pulled toward the posterior by the prior-matching loss, which teaches the dynamics to predict what the observations will show.

\paragraph*{Forecast mode.}
The active state $\mathbf{z}^{*}_t$ is the posterior when observations are present and the prior otherwise, and it is fed to the next step. 
Rollout therefore uses the same cell as assimilation, with no
separate forecast model, and each step costs the same regardless of how many steps came before. 
The memory and latent state are initialized to zero at the start of each sequence, in training and at inference, so the first steps of the 12\,h assimilation window serve as spin-up. 
All results in this paper therefore come from cycles of 13 assimilation steps. 
Continuous cycling over longer periods, in which the state is never reset, is left to future work.
For ensemble forecasts, a spatially correlated perturbation modulates every message-passing round and the prior head through feature-wise linear modulation (FiLM) layers.

\subsection{Decoding the Latent State to Observation Space}
\label{sec:decoders}
\noindent The active latent state ($\mathbf{z}_t$ during assimilation, $\hat{\mathbf{z}}_t$ during rollout) is mapped back to observation space by lightweight modality-specific decoders. 
Because decoders are independent given the latent state, only the modalities of interest need to be evaluated at inference. 
Each decoder is a local implicit function \citep{chen2021learning}: for any query coordinate $(t, y, x)$, it combines the nearest latent tokens with a Fourier encoding of the query's offset from them (Text S4.1, Figure S2). 
Fields can therefore be decoded at arbitrary resolution and at the exact locations of stations and soundings without regridding. 
Profiles are decoded in two steps, a horizontal decode of a column feature followed by a vertical MLP evaluated at each reported level (Text S4.2). 
Precipitation and cloud-top height, which contain significant counts of zero, use decoders that predict occurrence and magnitude separately, with the magnitude loss applied only where the observed value is non-zero (Text S4.3).

\paragraph*{Viewing-angle and elevation conditioning.}
Geostationary and microwave sounder decoders are conditioned on the excess air mass $\ell$ of the view being decoded through feature-wise linear modulation (FiLM; \citealp{perez2018film}) of their hidden features:
\begin{equation}
  \mathbf{x} \leftarrow \mathbf{x} \odot \left(1 + \gamma(\ell)\right) + \beta(\ell),
  \label{eq:film_angle}
\end{equation}
where $\gamma$ and $\beta$ are bias-free MLPs with zero-initialized final layers, so that $\gamma(0) = \beta(0) = 0$ and the unmodulated decoder corresponds exactly to a nadir view. 
The angle effect is learnable because the same location is observed at different angles, by overlapping geostationary satellites and across polar-orbiter scan positions, which breaks the collinearity between position and viewing angle (Section \ref{sec:limb}).
Station decoders are conditioned in the same way on the elevation difference between the station and its nearest latent node, so they can account for terrain below the latent grid resolution.

\subsection{Training and Ensemble Inference}
\label{sec:training}
\noindent WeatherDreamer is trained end-to-end from scratch in a single, joint optimization phase across all observation modalities simultaneously. 
The model has 50 million trainable parameters, with an additional 99 million tokenizer parameters that are frozen during training.
Training is executed on 12-step sequences of spatial patches extracted directly from the HEALPix grid at an hourly cadence ($\Delta t = 1\text{ hour}$).

\paragraph*{Masking.}
\label{sec:masking}
Three masking strategies teach the posterior to infer what it cannot see.
With probability $p_{mod}=0.5$ per sequence, a random subset of observing systems is removed from every step of the 12-hour assimilation window and retained only as reconstruction targets, forcing the posterior to infer them from the remaining sensors.
With probability $p_{block}=0.5$, a contiguous spatial block of tokens is withheld, drawn once per window, encouraging learning gap-filling given partial spatial modality coverage.
Finally, we mask a fraction $f$ of surface stations drawn from a continuous uniform distribution  $f \sim U(0.05, 0.30)$ from each sequence.
The hidden values remain as loss targets, so the model must learn to reconstruct a station from its neighbors rather than from the station's own history.

\paragraph*{Training objectives.}
\label{sec:losses}
The model is trained end to end on a weighted sum of three losses, weighted $\beta_{\text{rec}} = 1.0$, $\beta_{\text{prior}} = 0.5$, and $\beta_{\text{rollout}} = 1.0$:
\begin{equation}
\mathcal{L}_{\text{total}} = \beta_{\text{rec}} \mathcal{L}_{\text{recon}} + \beta_{\text{prior}} \mathcal{L}_{\text{prior}} + \beta_{\text{rollout}} \mathcal{L}_{\text{rollout}}
\end{equation}

The reconstruction loss $\mathcal{L}_{\mathrm{recon}}$ scores each modality decoded from the posterior during the assimilation window against its observations. 
For physical targets (point observations and directly embedded grids) it is the sum of absolute and squared errors; for pre-tokenized modalities, whose targets are unit-norm tokens, it is the cosine distance:
\begin{equation}
  \ell^m_t =
  \begin{cases}
    \lVert \hat{\mathbf{y}}^m_t - \mathbf{y}^m_t \rVert_1
    + \lVert \hat{\mathbf{y}}^m_t - \mathbf{y}^m_t \rVert_2^2
      & \text{physical targets}, \\[2pt]
    1 - \hat{\mathbf{y}}^{m\top}_t \mathbf{y}^m_t
      & \text{token targets},
  \end{cases}
\end{equation}
averaged over valid observations, with zero-inflated fields using the loss in Text S4.3. 
Each term is weighted by how the target was hidden from the encoder: $w = 1$ for observations the encoder also saw, $w = 2$ for block-masked targets, and $w = 3$ for modality-masked targets, so that inferring unseen data counts more than reproducing seen data.

The prior-matching loss $\mathcal{L}_{\mathrm{prior}}$ trains the prior to predict the posterior,
\begin{equation}
    \mathcal{L}_{\mathrm{prior}} = \sum_{t}\sum_{i} \big\lVert \hat{\mathbf z}_{t,i} - \operatorname{sg}(\mathbf z_{t,i}) \big\rVert^2 ,
\end{equation}
where sg(·) is the stop-gradient operator.  
The posterior is treated as a fixed target, so this term trains the prior network and the recurrent dynamics but not the encoder. 
This loss is what teaches the dynamics to forecast: the prior can match the posterior only by anticipating the observations about to arrive.

The rollout loss applies after the 12\,h assimilation window.
Observations are withheld and the model runs on its prior for $K$ steps ($K = 6$, extended to $K = 12$ for the final training steps), with $M = 2$ ensemble members,
\begin{equation}
  \mathcal{L}_{\mathrm{rollout}} = \sum_{k=1}^{K} \frac{1}{|\mathcal{M}_k|} \sum_{m \in \mathcal{M}_k}
  \mathrm{CRPS}_\alpha\!\left(\{\hat{\mathbf y}^{m,(j)}_{T+k}\}_{j=1}^{M},\; \mathbf y^m_{T+k}\right),
\end{equation}
where $T$ is the last assimilation step, $\mathcal{M}_k$ the modalities observed at step $T+k$, and $\mathrm{CRPS}_\alpha$ the almost-fair CRPS of Eq.~\ref{eq:afcrps}.
For token targets the cosine distance is added to the CRPS, and zero-inflated fields keep their occurrence term.
Because gradients flow through the whole rollout, the dynamics are trained directly on the error growth they will produce at inference.

\begin{figure}[t!]
\noindent\includegraphics[width=\textwidth]{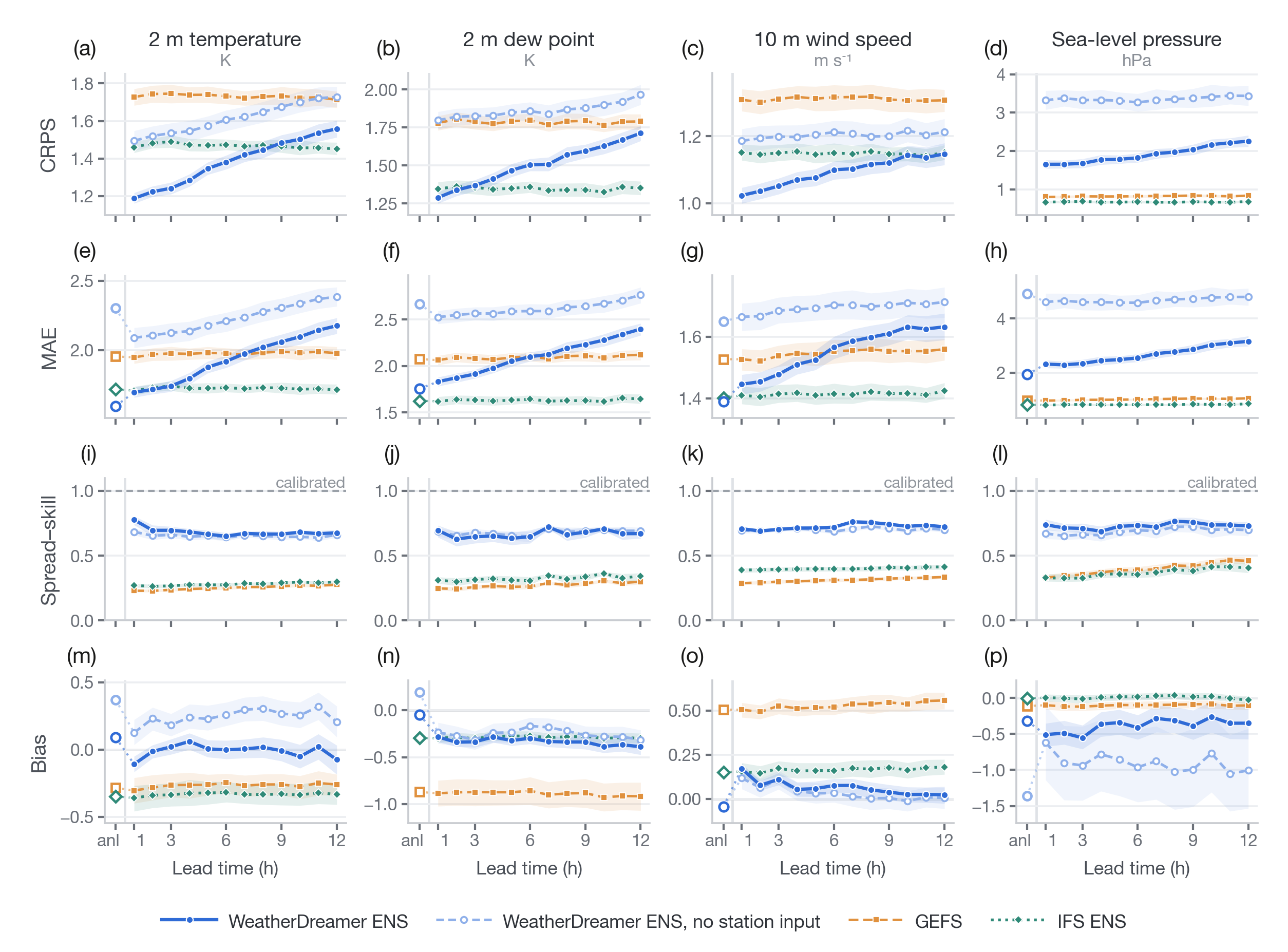}
\caption{
\textbf{Probabilistic short-range forecast skill against surface weather stations.} 
WeatherDreamer ensemble forecasts are verified against surface observations and compared with the GEFS (31 members) and IFS ENS (51 members) operational ensembles.
Two WeatherDreamer ensembles show performance at held-out ISD stations while other ISD stations are assimilated (blue, solid line) and performance when no ISD stations are assimilated, relying only on other sensing systems (light blue, dashed line).
Shading is the 95\% bootstrap interval from resampling 50 initializations.
}
\label{fig:weatherbench}
\end{figure}

\paragraph*{Ensemble training with a CRPS objective.}
\noindent To represent forecast uncertainty, WeatherDreamer is trained as a generative ensemble model with a proper scoring rule, the almost-fair CRPS (afCRPS; \citealp{Lang2024crps}), which interpolates between the standard and the fair (unbiased) ensemble CRPS for $M \geq 2$ members:
\begin{equation}
  \mathrm{CRPS}_\alpha = \frac{1}{M} \sum_{j} |x_j - y| - \frac{1 - \epsilon}{2M(M-1)} \sum_{j \neq k} |x_j - x_k|,
  \quad \epsilon = \frac{1 - \alpha}{M}, \quad \alpha = 0.95,
  \label{eq:afcrps}
\end{equation}
for members $x_j$ and observation $y$.

During the autonomous rollout phase each training window is expanded into  $M=2$ members that share the assimilation state and differ only through injected AR(1) noise (Text S5.1). 
At inference we run zero-noise control and 8 perturbed members from the same analysis.

\paragraph*{Inference.}
\label{sec:inference}
\noindent To create global forecasts from a patch-trained model, at inference, each of the 12 HEALPix base faces is padded with a 16-cell halo from its neighbors and tiled with $32 \times 32$ patches at 50\% overlap. 
Each patch carries its own memory and latent state through the assimilation window and forecast, and patch outputs are blended into global fields with a two-dimensional Hann window, which removes seams at patch edges. 
The analysis is produced by running the assimilation recursion over the 12\,h before initialization and the initialization hour.
Forecasts then roll out on the prior with a zero-noise control and 8 perturbed members using the EMA weights.

\section{Results}
\label{sec:results}

\noindent We evaluate WeatherDreamer both as a short-range forecast system and as a data assimilation system, verifying against observations throughout (Text S6).
We evaluate on \emph{Set A} (50 initializations between August 2024 and August 2025, run with operational latency; Table S9) and \emph{Set B} (30 initializations between January and October 2024, run without operational latency for comparison with the NCEP GSI; Table S10).

\subsection{Near-Surface Weather Forecasts}
\label{sec:isd}
\noindent We verify 2 m temperature, 2 m dew point, 10 m wind speed and sea-level pressure forecasts at the 534 held-out ISD stations over the 50 Set A initializations (Text S6.4), against GEFS and IFS ENS under operational latency (Fig. \ref{fig:weatherbench}). 
WeatherDreamer's ensemble is more skillful than GEFS for temperature, dew point and wind at every lead to 12\,h and better calibrated than both operational ensembles.
IFS ENS becomes more skillful for temperature and dew point after the first hours, and sea-level pressure is a clear weakness.

\paragraph*{Probabilistic skill.}
WeatherDreamer outperforms GEFS for temperature, dew point and wind speed at every lead time out to 12 h as measured by CRPS (Fig. \ref{fig:weatherbench}).
For 10 m wind speed it is the most skillful of the three ensembles at every lead, with a CRPS of 1.09 versus 1.31 (GEFS) and 1.15 m\,s$^{-1}$ (IFS ENS) averaged over leads 1--12 h. 
For 2 m temperature it is the most skillful ensemble for the first 8 h.
Its CRPS grows with lead while those of the operational ensembles stay nearly constant, so IFS ENS becomes the more skillful forecast after about 8 h.
For 2 m dew point, WeatherDreamer outperforms IFS ENS over the first 2 h (1.29 vs.\ 1.35 K at +1 h) and improves on GEFS at all lead times (1.51 vs.\ 1.79 K averaged), but IFS ENS is the most skillful thereafter. IFS ENS is also the only one of the three whose data assimilation uses near-surface humidity and temperature observations over land.

\paragraph*{Ensemble-mean accuracy.}
The ensemble-mean MAE is closer between systems than the CRPS. WeatherDreamer's ensemble mean is more accurate than GEFS's for temperature and dew point over the first 6 h and for wind speed over the first 5 h, and more accurate than IFS ENS's for temperature over the first 3 h; otherwise IFS ENS has the lowest MAE.
WeatherDreamer's larger lead in CRPS than in MAE therefore comes largely from its ensemble spread, which matches the station-level uncertainty considerably better. 
At the analysis time, WeatherDreamer's posterior state, a single deterministic decode, is more accurate for temperature than the operational forecasts valid at the same time (MAE 1.60 K, against 1.95 and 1.72 K for GEFS and IFS ENS), and comparable for wind speed, at stations it hasn't observed.
Decoded on a global grid, the station decoder produces surface analyses broadly consistent with the GFS analysis, although they are least constrained far from stations, notably over the Southern Ocean and Antarctica (Fig. S4).

\paragraph*{Calibration.}
Measured against point observations, all three ensembles are under-dispersive, but WeatherDreamer much less so: its spread--skill ratio is 0.63--0.78 at all lead times and for all four variables, against 0.23--0.47 for GEFS and 0.26--0.41 for IFS ENS. 
The operational ensembles describe uncertainty in grid-box averages, and their spread does not account for the sub-grid representativeness error of a single station. 
WeatherDreamer decodes directly at station locations, and its stochastic members carry part of that local variability. 
No observation-error term is added to any ensemble, so these ratios are lower bounds on calibration for all three systems.

\paragraph*{Contribution of station observations.}
To isolate the contribution of surface stations, we repeated all forecasts with ISD withheld entirely (Fig. \ref{fig:weatherbench}, light blue dashed line), so that every station is unseen and the state is constrained only by satellites, radiosondes and the other observing systems.
Removing station input degrades every variable at every lead time, increasing the CRPS averaged over leads 1--12 h by 10\% for wind speed, 17\% for temperature, 24\% for dew point and 76\% for sea-level pressure.
Even without any surface observations, WeatherDreamer remains more skillful than GEFS for 2 m temperature out to 11 h and for 10 m wind speed at every lead time, although IFS ENS is better for all four variables.
The benefit of station input is largest at the analysis and decays with lead as the forecast relaxes toward the state constrained by the remaining observations.
Sea-level pressure depends most on stations: without them the analysis MAE rises from 1.9 to 4.9 hPa, consistent with the weak constraint that satellite observations place on surface pressure.
Ensemble calibration is nearly unchanged, so withholding stations makes the ensemble less accurate but not overconfident.

\paragraph*{Biases and limitations.}
WeatherDreamer's mean biases are small for temperature ($-0.11$ to $+0.09$ K, against $-0.25$ to $-0.36$ K for the
operational ensembles) and for wind speed.
Its dew-point bias is negligible at the analysis ($-0.05$ K) but grows
to about $-0.3$ to $-0.4$ K at forecast leads, much smaller than that of GEFS ($-0.9$ K) and similar to that of IFS ENS ($-0.3$ K).
Sea-level pressure is the clear weakness.
The analysis fits assimilated stations about as closely as GEFS and IFS, but the correlation between forecast and observed SLP change is about 0.45 at 6–12 h, compared with 0.6–0.9 for GEFS and IFS.

\begin{figure}[t!]
\noindent\includegraphics[width=\textwidth]{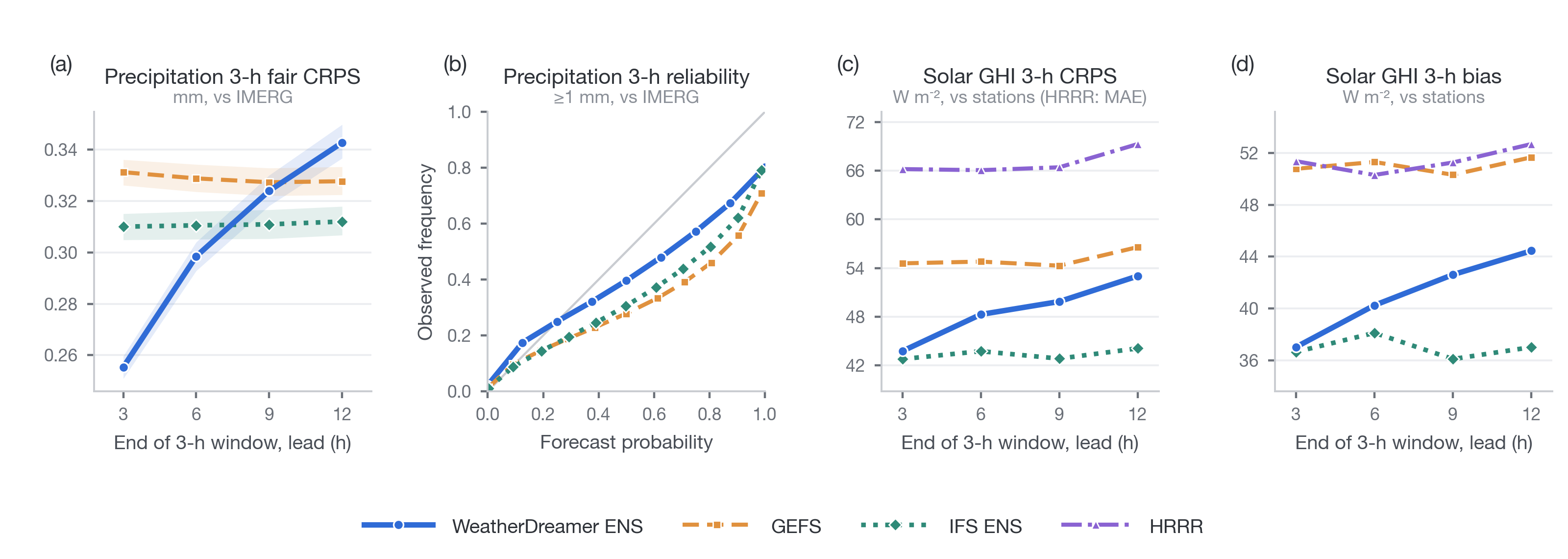}  
\caption{\textbf{Short-range precipitation and solar irradiance forecasts.}
Neither variable is a WeatherDreamer input at inference. The WeatherDreamer ensemble (8 members) is compared with GEFS (31 members), IFS ENS (51 members) and, for irradiance, HRRR (deterministic).
(a, b) Liquid precipitation between 60\textdegree S and 60\textdegree N, verified against IMERG Final V07B: (a) fair CRPS of 3-h accumulations; (b) reliability of the 3-h probability of at least 1\,mm, pooled over leads (gray line: perfect reliability).
(c, d) Global horizontal irradiance at 116 USCRN and 7 SOLRAD stations over the contiguous United States, daylight windows only: (c) fair CRPS of 3-h mean GHI (MAE for HRRR); (d) mean bias.
Shading in (a) is the 95\% bootstrap interval from resampling initializations.
}
\label{fig:precip_solar}
\end{figure}

\subsection{Inferring Precipitation, Solar Irradiance and Cloud-Top Height}
\label{sec:unobserved}
\noindent Precipitation, surface irradiance and cloud-top height are never model inputs at inference, so everything it predicts about them at inference comes from the other observing systems through the shared latent state. 
Their skill is therefore a test of how much physically related information the latent state carries beyond the instruments it was given.

\subsubsection{Precipitation}
\label{sec:precip}
\noindent We verify precipitation against IMERG Final V07B over liquid-precipitation regions between 60$^\circ$S and 60$^\circ$N, comparing hourly rates with GFS and 3-h accumulations with GEFS and IFS ENS (Text S6.5). 
Through +6\,h WeatherDreamer is the most skillful and most reliable of the three ensembles and places intense rain better than GFS, but its rain is too light and too smooth.
IMERG is itself derived partly from ATMS and geostationary infrared observations that WeatherDreamer ingests up to the analysis hour, so the comparison is not fully independent.

\textit{Probabilistic skill.}
For 3-h accumulations, WeatherDreamer is the most skillful of the three ensembles through +6 h (fair CRPS 0.255 mm at +3 h, against 0.310 for IFS ENS and 0.331 mm for GEFS; Fig. \ref{fig:precip_solar}a). 
It falls behind IFS ENS by +9 h and behind both by +12 h.
The Brier skill score for $\geq$1 mm per 3 h follows the same course, slightly above IFS ENS at +6 h (0.27 vs.\ 0.25), and below IFS ENS by +12 h (Fig. S5c).
WeatherDreamer's 3-h rain probabilities are also the most reliable: where it forecasts 50\%, rain is observed 40\% of the time, against 31\% for IFS ENS and 28\% for GEFS (Fig. \ref{fig:precip_solar}b).
At lower probabilities WeatherDreamer slightly under-forecasts.

\textit{Placement.}
We score the WeatherDreamer ensemble mean with frequency-matched thresholds, because it
covers too little area above any fixed threshold (frequency bias 0.67 at +1 h for
1 mm h$^{-1}$), so a fixed threshold would mostly measure this coverage deficit rather
than placement.
For intense rain ($\geq$5 mm h$^{-1}$) WeatherDreamer's frequency-matched Equitable Threat Score (ETS) (higher is better) is more than twice GFS's at short leads, 0.22 against 0.10 at +1 h (Fig. S5a).
The two converge at +6 h, and GFS has higher ETS from about +7 h.

\textit{Amount and structure.}
WeatherDreamer is dry, with a mean rate about two-thirds of IMERG's at every lead,
including the analysis hour, whereas GFS is within 6\% of IMERG. 
The deficit is already present in the analysis rather than accumulating with lead.
Its fields are also too smooth (Fig. S5b): wavelengths below 1000 km carry 33\% of the zonal power at +1 h, against 56--57\% for IMERG and 56\% for GFS.
Predictions at short leads therefore come with underestimated small-scale variability, which grows with lead.

\subsubsection{Surface Solar Irradiance}
\label{sec:solar}
\noindent We verify 3-h mean global horizontal irradiance (GHI) in daylight windows at 116 USCRN and 7 SOLRAD stations across the contiguous United States against GEFS, IFS ENS and HRRR (Text~S6.6).
WeatherDreamer reproduces station irradiance most closely at the analysis and is more skillful than HRRR and GEFS at every lead, while IFS ENS is most skillful from +6\,h.

\paragraph*{Analysis.}
At the analysis hour, WeatherDreamer reproduces station irradiance more closely than any
baseline. 
Against station 1-h means, its correlation is 0.96 and its RMSE 86 W\,m$^{-2}$, compared with correlation 0.91--0.93 and RMSE 109--122 W\,m$^{-2}$ for GEFS, IFS ENS
and HRRR.

\paragraph*{Forecasts.}
WeatherDreamer's CRPS is lower than HRRR's at every lead, by 16--22 W\,m$^{-2}$
(Fig. \ref{fig:precip_solar}c). 
It is also lower than GEFS's at every lead, by 10.8 [6.1, 15.9] W\,m$^{-2}$ at +3 h, narrowing to 3.6 [0.6, 6.4] W\,m$^{-2}$ at +12 h.
IFS ENS is the most skillful forecast: indistinguishable from WeatherDreamer at +3 h
and better from +6 h onward, by 4.5--8.9 W\,m$^{-2}$.
All systems overestimate GHI, consistent with too little cloud
(Fig. \ref{fig:precip_solar}d). WeatherDreamer's bias grows from $+37$ to
$+45$ W\,m$^{-2}$ with lead, close to that of IFS ENS at +3 and +6 h and below those of
GEFS and HRRR ($+50$ to $+53$ W\,m$^{-2}$) at every lead.
As at surface stations, WeatherDreamer's ensemble is better calibrated than the operational ensembles (spread--skill ratio 0.61--0.62, against 0.47--0.52).

\begin{figure}[t!]
\noindent\includegraphics[width=\textwidth]{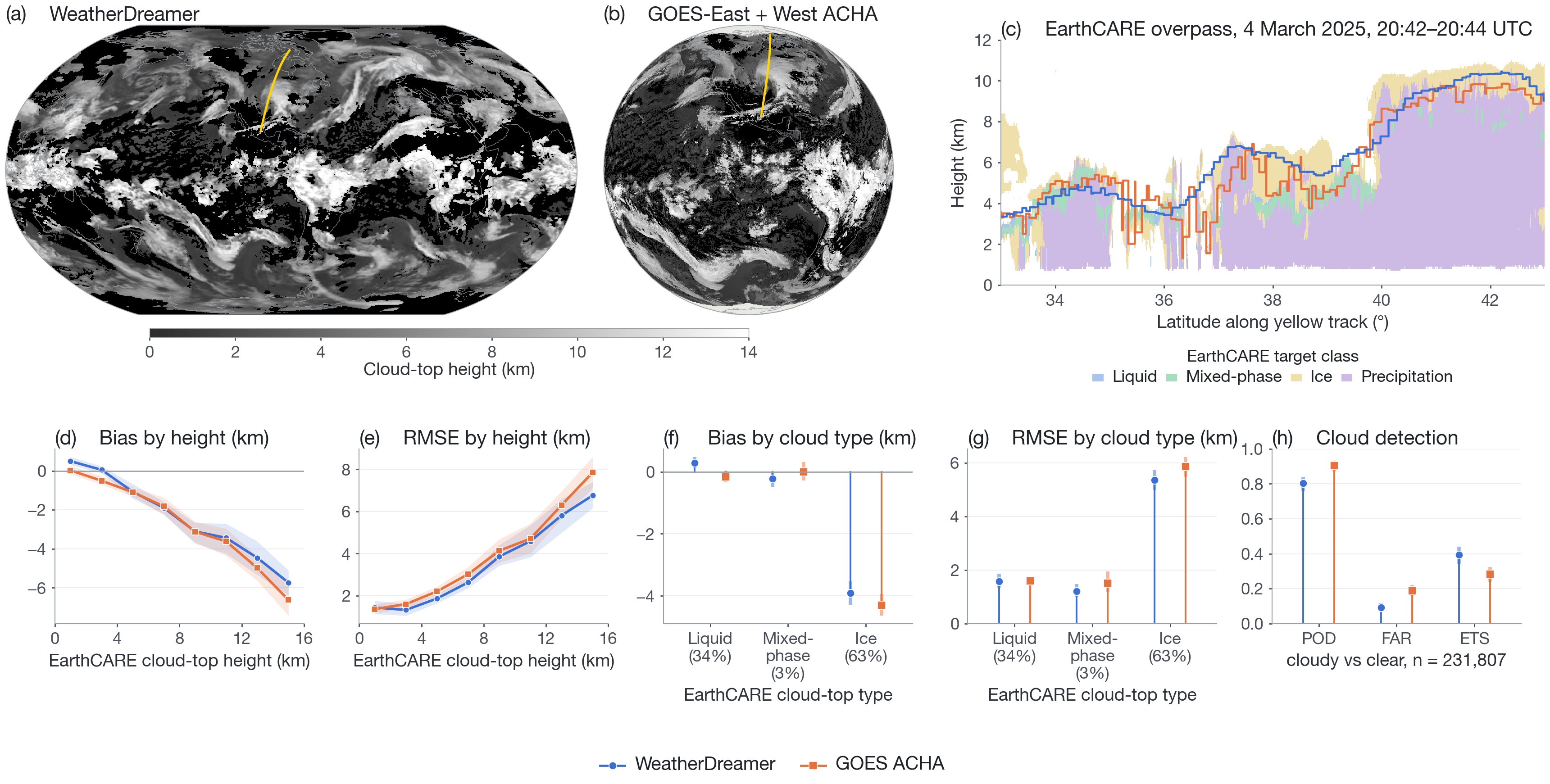}
\caption{
\textbf{Cloud-top height verified against EarthCARE.} 
For March 4, 2025 at 21 UTC:
(a) WeatherDreamer analysis. 
(b) GOES-East and GOES-West ACHA.
(c) EarthCARE classification (AC\_TC\_2B) along the yellow track at 20:42–20:44 UTC between 33° and 43°N, with WeatherDreamer and GOES cloud-top heights overlaid.
(d–h) Scores against EarthCARE for all analyses with an overpass in the GOES-East disk (28 analyses; n = 154,860 cloudy footprints for height, 231,807 for detection).
(d, e) bias and RMSE by EarthCARE cloud-top height in 2 km bins;
(f, g) bias and RMSE by EarthCARE cloud-top type: liquid, mixed-phase, and ice.
(h) cloudy-vs-clear detection, with WeatherDreamer cloudy where cloud probability $\geq$ 0.5 and GOES cloudy where ACHA reports a cloud top.
Bars and shading: 95\% interval from resampling whole initializations.
}
\label{fig:cth}
\end{figure}

\subsubsection{Cloud-top Height}
\label{sec:cloud}
\noindent Cloud-top height is observed directly only by active sensors
along narrow nadir curtains, while geostationary retrievals infer it
indirectly from passive radiances. 
WeatherDreamer learns it from CloudSat, which is never an input at inference, and we verify its analyses against EarthCARE AC\_TC\_2B footprints, an active instrument never seen in training, over 28 analyses in the GOES-East disk (Text~S6.7; Fig.~\ref{fig:cth}a--c). 
WeatherDreamer's heights are more accurate than the operational GOES ABI retrieval (ACHA), mainly for ice-topped cloud, and it detects cloud with half the false-alarm ratio.

\paragraph*{Height accuracy.}
Over 154,860 cloudy footprints, WeatherDreamer reaches $r = 0.73$ and RMSE
4.4\,km, against $r = 0.70$ and 4.8\,km for ACHA, with lower RMSE in 18 of
the 27 analyses with at least 50 collocated footprints. 
RMSE is similar below 4\,km (1.4\,km for both) and lower for WeatherDreamer at 4--8\,km (2.3 vs 2.7\,km) and above 12\,km (6.8 vs 7.6\,km; Fig. \ref{fig:cth}d,e). 
The advantage comes from ice-topped cloud (63\% of the sample; RMSE 5.4 vs 5.9\,km); for liquid-topped cloud the two are equivalent (1.6\,km; Fig.~\ref{fig:cth}f,g).

\paragraph*{Cloud detection.}
As a cloudy/clear classification on all 231,807 footprints
(Fig.~\ref{fig:cth}h), WeatherDreamer's equitable threat score (higher is better) is 0.39 against 0.28 for ACHA,
It detects less cloud (probability of detection 0.80 vs 0.90) with half the false-alarm ratio (0.09 vs 0.19).

\paragraph*{Limitations.}
Both products are biased low against EarthCARE, by 2.4\,km
(WeatherDreamer) and 2.8\,km (ACHA), almost entirely in ice-topped cloud, consistent with active sensors detecting ice tops above the level that passive retrievals report.
ACHA was developed in the context of atmospheric motion vectors and has recently aimed to improve the height assignment of cirrus clouds \citep{heidinger2014noaa}. 
ACHA requires a NWP background state from GFS or IFS. 
WeatherDreamer also ingests geostationary radiances at the analysis hour, so its heights are not independent of the GOES observations underpinning ACHA.

\begin{figure}[t!]
\noindent\includegraphics[width=\textwidth]{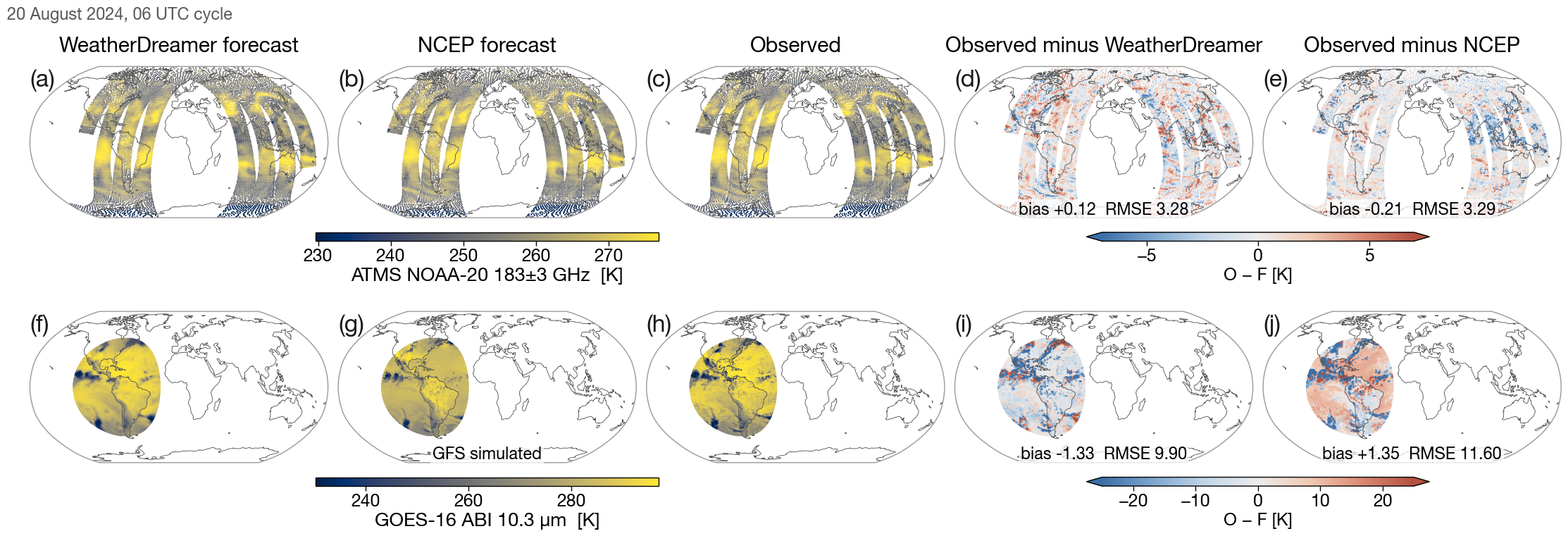}
\caption{
\textbf{WeatherDreamer and NCEP forecasts compared with identical observations.}
Observations of the 06 UTC GSI cycle on 20 August 2024 (ATMS valid 03–09 UTC; GOES-16 valid 06 UTC) are compared with WeatherDreamer and NCEP forecasts initialized at 00 UTC.
WeatherDreamer is decoded at the observed viewing geometry. 
For ATMS, NCEP is the first guess from the GSI diagnostic files; for GOES-16, it is GFS 6-h simulated imagery, because GSI assimilates only clear-sky ABI radiances.
Columns: WeatherDreamer forecast, NCEP forecast, observation, and the observation-minus-forecast departures for WeatherDreamer and NCEP, with mean (bias) and RMSE in each panel.
Rows: ATMS NOAA-20 channel 20 (183±3 GHz; n = 11,340 observations) and GOES-16 ABI 10.3 µm under all-sky conditions (n = 36,060).
}
\label{fig:departure_maps}
\end{figure}

\subsection{Evaluating the Learned Data Assimilation Cycle}
\label{sec:departures}
\noindent The forecast skill in the preceding sections depends on the learned assimilation cycle, which we examine in three steps: how forecast skill improves with more hours of observations (Section \ref{sec:departures_warmup}), how well the short-range background matches the GSI first guess (Section \ref{sec:departures_f}), and how the learned update weights observations (Section \ref{sec:departures_err}).

\subsubsection{Skill Grows With Assimilated History}
\label{sec:departures_warmup}
\noindent We initialized forecasts after assimilating 1, 3, 6 or 12\,h of observations and verified them at held-out stations (Fig.~S6) and against radiosondes and radio occultation.
At the surface, a longer window improves temperature and dew-point skill at every lead: with 1\,h of observations WeatherDreamer trails GEFS after 6\,h, whereas with 12\,h it outperforms GEFS for temperature out to 12\,h.
For these variables, ensemble spread shrinks as the window lengthens, in step with the lower error, so the ensemble's uncertainty reflects how much observational information it has received.
Wind speed skill and spread are nearly independent of window length, and wind speed beats GEFS throughout.

Above the surface, the benefit holds for every variable evaluated, including wind.
Pooled over leads 1--12\,h and scored on identical observations, forecasts initialized after 1\,h of assimilation have a CRPS 15\% higher than those after 12\,h for 925\,hPa radiosonde temperature, 6\% higher for dew-point depression, 8--16\% higher for the wind components and 8\% higher for radio-occultation bending angle at 5\,km.
As at the surface, most of the improvement is gained within the first 3--6\,h.

\subsubsection{Comparison With Variational Data Assimilation}
\label{sec:departures_f}
\noindent The standard diagnostic of a background's quality is its departure from new observations before they are assimilated (O$-$B). 
We compare the standard deviation (SD) of these departures, which measures random error separately from bias, because GSI's departures are taken before its variational bias correction, which removes much of their bias in practice.
The NCEP GSI background is a 3--9\,h forecast from the previous 6-hourly cycle, so for the 30 Set B cycles we score WeatherDreamer forecasts at the same 3--9\,h leads, decoded at each observation's viewing geometry, and refer to these as forecast departures (O$-$F). 
Both systems are evaluated on identical observations from the GSI diagnostic files, following the matched-sample procedure of \citet{alexe2024graphdop}, and including cloud- and precipitation-affected as well as clear-sky observations \citep{geer2018all} (Text S6.9; Fig. \ref{fig:departure_maps}).
Benchmarked against a mature system that assimilates many more observation types, WeatherDreamer's background is close to the system's quality near the surface and for moisture but substantially worse for free-tropospheric temperature and wind (Table \ref{tab:departures_surface}).

\begin{table}[b!]
\centering
\small
\caption{\textbf{Standard deviation of observation-minus-forecast departures.} 
The NCEP GSI background and WeatherDreamer (WD, ensemble mean) are compared to identical observations, at matched lead times of 3--9\,h over 30 cycles in 2024.
For multi-channel or multi-level rows, GSI and WD are the root-mean-square of the per-channel SDs and Range gives the spread of per-channel ratios.
}
% Temperature is virtual temperature (see Methods).}
\label{tab:departures_surface}
\begin{tabular}{lcccc}
\toprule
 & GSI & WD & Ratio & Range \\
\midrule
\multicolumn{5}{l}{\textit{ATMS (NOAA-20), ocean}} \\
\quad Window, ch 1--4, 16 (K) & 5.06 & 5.35 & 1.06 & 0.97--1.35 \\
\quad Troposphere T, ch 5--9 (K) & 0.38 & 0.67 & 1.76 & 1.35--4.55 \\
\quad Stratosphere T, ch 10--15 (K) & 0.53 & 0.98 & 1.84 & 1.38--3.64 \\
\quad Water vapour, ch 17--22 (K) & 3.39 & 3.54 & 1.04 & 0.96--1.23 \\
\multicolumn{5}{l}{\textit{HIRS (NOAA-19), ocean}} \\
\quad Temperature, ch 2--7, 13--17 (K) & 4.49 & 3.87 & 0.86 & 0.82--1.21 \\
\quad Water vapour and window, ch 8, 10--12, 18, 19 (K) & 6.83 & 6.14 & 0.90 & 0.88--1.06 \\
\multicolumn{5}{l}{\textit{GPS-RO (MetOp-C)}} \\
\quad Bending angle, 5--30 km (\%) & 4.96 & 5.69 & 1.15 & 1.09--1.52 \\
\multicolumn{5}{l}{\textit{Radiosondes}} \\
\quad Temperature, 1000--100 hPa (K) & 1.38 & 2.47 & 1.79 & 1.19--2.50 \\
\quad Specific humidity, 1000--300 hPa (g\,kg$^{-1}$) & 1.06 & 1.38 & 1.30 & 1.09--1.39 \\
\quad Wind ($u$, $v$), 1000--100 hPa (m\,s$^{-1}$) & 3.01 & 7.89 & 2.62 & 1.27--3.51 \\
\multicolumn{5}{l}{\textit{Surface stations, held out}} \\
\quad Temperature (K) & 2.37 & 2.47 & 1.04 & -- \\
\quad Specific humidity (g\,kg$^{-1}$) & 1.37 & 1.39 & 1.02 & -- \\
\quad Wind speed (m\,s$^{-1}$) & 2.05 & 2.22 & 1.08 & -- \\
\multicolumn{5}{l}{\textit{ASCAT, ocean}} \\
\quad Wind speed (m\,s$^{-1}$) & 1.31 & 2.16 & 1.65 & -- \\
\bottomrule
\end{tabular}
\end{table}

The gap between WeatherDreamer's forecast and the GSI first guess varies widely across the observing system (Table \ref{tab:departures_surface}; Tables S1--S7). 
It is largest for microwave temperature-sounding channels peaking in the mid-to-upper troposphere and lower stratosphere: ATMS channels 7--12 have departure standard deviations 2.7--4.6 times GSI's, and AMSU-A channels 5--11 1.5--2.7 times. 
Here the temperature field is smooth, the GSI background is constrained by a dense sounder network, and decades of radiative-transfer and bias-correction work have brought GSI's departures close to instrument noise. 
WeatherDreamer's biases in these channels are nonetheless small ($|b| \le 0.18$\,K), whereas GSI's uncorrected departures are biased by up to $-0.9$\,K before variational bias correction.
The gap narrows where the signal depends on moisture, the surface or fine-scale structure. 
Departures are 0.96--1.23 times GSI's at 183\,GHz, 0.97--1.35 times in the microwave window channels, and 1.1 times for GPS-RO bending angle at 5\,km impact height, where moisture dominates, rising to 1.4--1.7 times at 10--30\,km. 
Radiosondes show the same vertical structure: humidity departures are 1.2--1.4 times GSI's, temperature 1.5 times at 925\,hPa and 1.8--2.5 times aloft, and winds 1.6--1.7 times at 925--850\,hPa, rising to 3.2--3.5 times at 300--200\,hPa (Table S5).

At the surface, WeatherDreamer's departures at stations held out from its input are 1.02–1.08 times GSI's for temperature, humidity and wind speed, and 1.01--1.08 times at in-sample stations.
Its temperature bias is small ($-$0.28\,K at held-out stations) where GSI's is $+$1.51\,K. 
GSI uses about three quarters of these temperature and humidity reports but only 3\% of the wind reports, so for wind its background is largely unconstrained by them. 
ASCAT ocean-surface winds are the exception, with departures 1.65 times GSI's (2.16 vs 1.31\,m\,s$^{-1}$ in wind speed); GSI's background fits ASCAT almost to its observation error.

Geostationary infrared radiances are assimilated operationally only as clear-sky averages, and GSI assimilates only two water-vapour bands (6.2 and 7.3\,\textmu m), so its departures for these channels exist only at clear-sky locations. 
Scored at those locations over ocean, WeatherDreamer's departures in the water-vapour bands (6.2--7.3\,\textmu m) are 2.0--2.3\,K against GSI's 1.2--1.6\,K.
In the 10.3\,\textmu m window they are 5.0 vs 0.7\,K, with a $+$3.5\,K bias: wherever WeatherDreamer places cloud at a location that turns out to be clear, its forecast is too cold.
In all-sky conditions, which the GSI system does not assimilate, we compare instead with GFS simulated imagery from the same 6-h forecast for one case (2024-08-20 06\,UTC; Fig. \ref{fig:departure_maps}): at 10.3\,\textmu m WeatherDreamer's RMSE is 9.9\,K against 11.6\,K for GFS, and at 6.9\,\textmu m 4.3 vs 5.0\,K.

\begin{figure}[ht!]
\centering
\includegraphics[width=\textwidth]{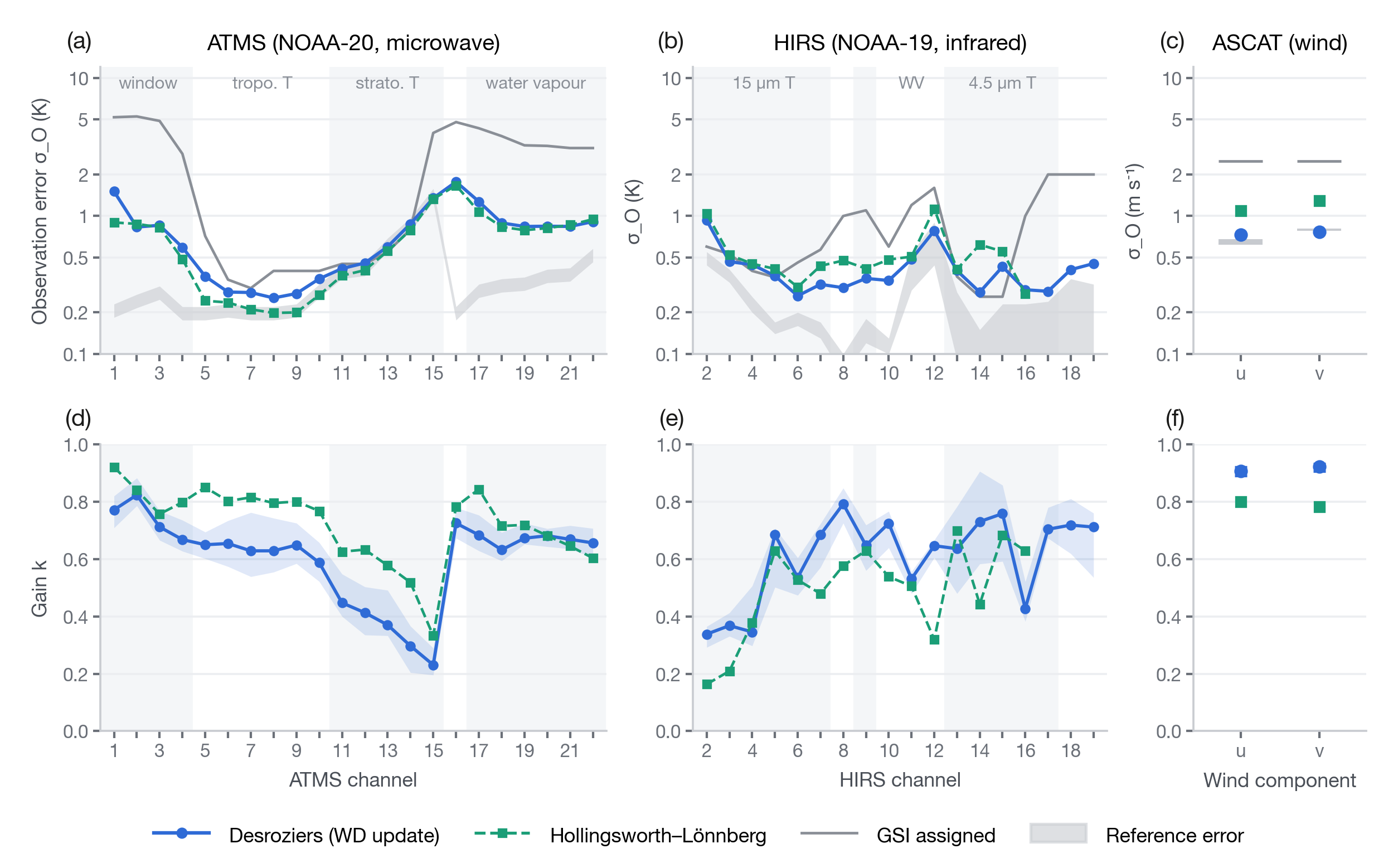}
\caption{
\textbf{Observation-error and gain diagnostics for WeatherDreamer's learned assimilation update.}
Diagnostics are computed for WeatherDreamer's own analysis cycle for  ATMS on NOAA-20 (clear sky), HIRS on NOAA-19 (clear sky) and ASCAT ocean-surface wind components.
\textit{Top row:} observation error $\sigma_O$ using the Desroziers estimate, $\sigma_O^2=\mathrm{cov}(O-A,\,O-B)$ (blue) and the Hollingsworth--Lönnberg (HL) estimate (green).
Gray shading shows reference errors: ATMS on-orbit instrument noise \citep{liu2019}, HIRS specified instrument noise \citep{noaaklm2014}, and ASCAT triple-collocation retrieval error \citep{verspeek2019}.
The gray line is the observation error assigned by NCEP's GSI, for context.
\textit{Bottom row:} The gain the update applies with shaded interquartile range across cycles (blue), and the optimal gain implied by the HL errors (green).
}
\label{fig:gain}
\end{figure}

\subsubsection{Learned Observation Weighting}
\label{sec:departures_err}

\noindent A well-behaved assimilation update should weight each observation according to its error relative to the background. 
WeatherDreamer is never given observation errors, so we diagnose the weighting it has learned from its own analysis cycle (Text S6.10; Fig. \ref{fig:gain}). 
The method of \citet{desroziers2005} gives the effective observation error $\sigma_O$ the update behaves as if it assumes.
Regressing increments on innovations gives the gain $k$ the model applies. 
The method of \citet{hollingsworth1986statistical} independently estimates the errors and the optimal gain $k_{\mathrm{opt}}$.

The update has learned realistic observation precision for several instruments it was never told about. 
For clear-sky ATMS temperature channels, $\sigma_O$ is 0.2--0.4\,K, at the on-orbit instrument noise, and
for ASCAT winds it is close to the triple-collocation retrieval error. 
Its weighting is close to optimal for water-vapor and infrared radiances but conservative for microwave temperature sounding ($k \approx 0.64$ against
$k_{\mathrm{opt}} \approx 0.82$) and too strong for scatterometer winds ($k \approx 0.9$ against 0.8), whose representativeness error it largely ignores. 
Station observations correct the state mainly within about 200\,km, and station wind barely propagates to unobserved locations. 
The learned update applies a substantial fraction of each innovation in the lower troposphere for radiosondes, but the update draws little from upper-air observations: gains for radiosonde temperature and wind fall to 0.1--0.26 at 100--300\,hPa and are near zero for radio-occultation bending angle. 
This under-use matches where WeatherDreamer's departures lag most behind GSI's (Section \ref{sec:departures_f}) and is a main target for improvement (Section \ref{sec:discussion}).

\begin{figure}[ht!]
\centering
\includegraphics[width=\textwidth]{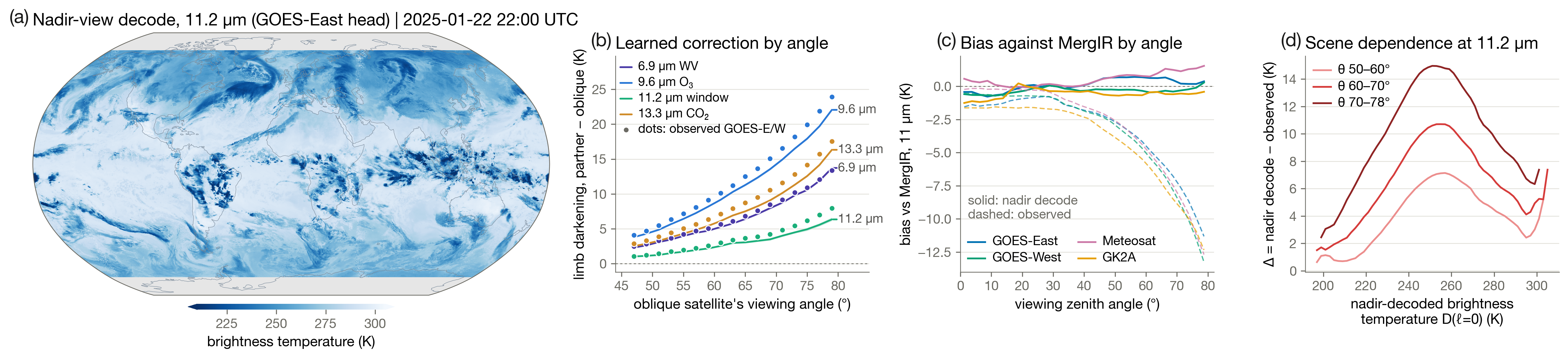}
\caption{
\textbf{Learned viewing-angle correction and harmonized geostationary infrared imagery.}
\textbf{(a)} WeatherDreamer 11.2 µm brightness temperature at 22 UTC on 22 January 2025, decoded at synthetic nadir ($\ell=0$) with the GOES-East head.
\textbf{(b)} Limb darkening in four GOES ABI bands over clear ocean.
Points show the observed difference between collocated GOES-East and GOES-West views of the same box at the same time (near-nadir view, $\theta < 25^\circ$, minus oblique view). 
Lines show the same difference predicted by the model, decoding one latent state at the two satellites' viewing angles. 
Shading and error bars are 68\% bootstrap intervals over analyses.
\textbf{(c)} Mean 11\,µm bias against GPM MergIR as a function of viewing angle for four geostationary imagers: observed brightness temperature (dashed) and the synthetic-nadir decode (solid), 60$^\circ$S--60$^\circ$N.
\textbf{(d)} Learned 11.2\,µm correction, $\Delta$ = nadir decode $-$ observed, as a function of the nadir-decoded brightness temperature (2\,K bins), for three viewing-angle ranges, over ocean pixels pooled across the four imagers.
}
\label{fig:limb}
\end{figure}

\subsection{Learned Limb Correction and Harmonized Satellite Imagery}
\label{sec:limb}

\noindent As a satellite's viewing zenith angle $\theta$ increases, the slant path through the atmosphere lengthens and the sensor sees emission from higher, colder layers.
As a result, apparent brightness temperatures cool toward the limb. 
In clear skies this effect is small in window channels and large in absorbing channels such as water vapor and ozone \citep{elmer2016limb}. 
It leaves cold limbs on geostationary disks, which result in discontinuities where coverage passes from one satellite to the next, and stripes where adjacent polar orbits meet. 
Observation-driven models such as GraphDOP and AIFS-DOP can learn the limb effect from a viewing-angle input \citep{lean2025learning, pinnington2026aifs}. 
WeatherDreamer separates geometry from state explicitly through zero-anchored angle conditioning of its embeddings (Section \ref{sec:embeddings}) and decoders
(Section \ref{sec:decoders}). 
The same posterior state can therefore be decoded at each satellite's observed angle or at any other angle, including nadir, with the difference separating the learned angle effect from reconstruction error.
The learned correction recovers 83--96\% of the observed limb darkening,
scales with absorption and scene temperature as expected, such that decoding at nadir removes the cold limbs and inter-satellite discontinuities of conventional composites.

\paragraph*{Validation against collocated observations.}
Collocated GOES-East and GOES-West views of the same clear-ocean scenes differ mainly through viewing geometry, since both carry the ABI sensor.
Over 0.25$^\circ$ boxes viewed by one satellite at $\theta < 25^\circ$ and by the other at 47--80$^\circ$,  the learned limb darkening follows the observed angle
dependence in all four bands tested at unseen times, recovering 83--96\% of the observed value and differing by at most about 2.5\,K at the limb (Fig. \ref{fig:limb}b).
For the microwave sounders, the observed brightness-temperature difference between NOAA-20 and Suomi-NPP at crossovers grows with the difference in $\sec\theta$, at 13.6\,K per unit secant for channel 7 and 6.1\,K for channel 20. 
The observed-angle decode reproduces these slopes, and the nadir decode reduces them to 0.3 and 0.1\,K, so the two satellites agree once geometry is removed
(Fig. S7).

\paragraph*{Form of the learned correction.}
The learned correction follows the expected dependence on absorption.
Among the four GOES-East bands it is smallest in the 11.2\,\textmu m window, reaching about 6\,K near 80$^\circ$, followed by water vapor (6.9\,\textmu m, 14\,K), CO$_2$ (13.3\,\textmu m, 16\,K) and ozone (9.6\,\textmu m, 22\,K) (Fig. \ref{fig:limb}b).
It also depends on the scene.
At 11.2\,$\mu$m and 70--78$^\circ$, it rises from about 3.5\,K over the coldest scenes, where opaque high cloud tops leave little atmosphere above the emitting layer, to about 15\,K near 250\,K, with a minimum near 295--300\,K (Fig. \ref{fig:limb}d). 
This resembles the empirical correction of \citet{joyce2001latitudinally} used for merged geostationary infrared products, but is learned for all bands jointly.

\paragraph*{Harmonized fields.} 
Merged infrared products such as GPM MergIR \citep{janowiak_joyce_xie_2017} and GridSat-B1 \citep{knapp2011globally}, used for example to propagate microwave rain rates in IMERG, are single-channel and correct viewing geometry empirically for the $\sim$11\,$\mu$m window only.
Decoding WeatherDreamer's latent state at nadir instead removes the angle dependence for every band (Figs. \ref{fig:limb}a and S7).
Against MergIR over 48 analyses (60$^\circ$S--60$^\circ$N), decoding at each pixel's observed angle reproduces the cold limbs and inter-satellite discontinuities of a conventional composite (mean bias $-3.2$\,K, $-7.8$\,K for $\theta > 60^\circ$), whereas the nadir decode reduces the mean bias to 0.0\,K ($-1.1$\,K beyond 60$^\circ$)  (Fig. \ref{fig:limb}c). 
Because MergIR applies its own empirical corrections, this tests consistency with operational practice rather than correctness. 
The angle conditioning corrects radiometric intensity but not position, so parallax displacement of high cloud tops at large angles remains, and part of the residual difference reflects a 1--2\,K offset between raw GOES-East observations and MergIR even near nadir.

\section{Discussion and Conclusions}
\label{sec:discussion}
\noindent WeatherDreamer shows that a predict--correct cycle can be learned end-to-end from observations alone, without reanalysis, hand-built observation operators or prescribed error statistics. 
The learned cycle is competitive with operational systems near the surface and for moisture, but not yet in the free troposphere.
Where the model does well, it behaves as a cycled system should: the benefit of station input is largest at the analysis and decays with lead as the dynamics carry it forward, and skill grows with the length of the
assimilation window. 
The learned cycle is also inexpensive. 
On a single NVIDIA H100, with patch-by-patch inference, one global hourly assimilation step takes about 1.5 minutes, of which roughly 45 seconds is model computation and roughly 45 seconds is reading in observations.
A 12\,h, 8-member ensemble forecast with 1 control member takes about 11 minutes. 
An analysis can therefore be updated every hour within minutes of the observations arriving, in contrast to the 6-hourly, hours-late analyses of global operational DA.
Because each analysis needs only the preceding 12\,h of observations it can be restarted cheaply at any time.
Carrying the state forward continuously, which could let it retain information beyond that window, is a natural next step.

The weaknesses in the upper troposphere reflect the strength of the operational background as well as two main sources of underperformance.
First, WeatherDreamer's update draws little from upper-air observations: its gain is low for upper-tropospheric radiosonde temperature and wind and near zero for radio occultation, so free-tropospheric errors are corrected less at each cycle than those near the surface. 
Effective observation errors and gains, monitored during training, from the model's own innovations, could be used to rebalance the loss toward under-used observing systems.
A coarser latent grid would let each sparse profile constrain a larger fraction of the state. 
Second, the dynamics operate on regional patches of roughly
1600\,km, which limits their ability to represent the large-scale, slowly evolving circulation that governs upper-level winds and synoptic pressure changes. 
Surface pressure, weakly constrained by satellites and sparsely observed over the oceans, is the clearest symptom. 
Global or larger-context dynamics, made more affordable by a coarser grid, would let the model represent this circulation directly.
The comparison with GSI also favors the operational system by construction: GSI assimilates a far larger observing system, including aircraft reports, satellite-derived winds, hyperspectral infrared sounders and radio occultation from many more missions, and benefits from decades of development, whereas WeatherDreamer ingests 18 observing systems, was trained on four years of data, and runs on a single GPU.

Several aspects of the evaluation qualify these results. WeatherDreamer's advantage in probabilistic skill at stations comes largely from its spread rather than its ensemble mean: its ensemble-mean error is lower than IFS ENS's only for temperature in the first 3 h, while its CRPS is lower for temperature over the first 8 h.
Because it decodes directly at station locations, part of its calibration advantage likely reflects sub-grid representativeness rather than a better estimate of the atmospheric state.
Held-out stations are a random 5\% of the network, so many have assimilated neighbors, and the improvement at held-out stations vanishes beyond about 200\,km. 
Station forecasts are also compared only with global ensembles; hourly-cycling regional systems such as HRRR and RAP, which are deterministic and cover only North America, would be a further benchmark over the contiguous United States.
However, even with every surface station withheld, WeatherDreamer's ensemble forecasts at those stations are more skillful than GEFS for 2\,m temperature out to 11\,h and for 10\,m wind speed out to 12\,h, while remaining better calibrated than both operational ensembles.
In the comparison with GSI, WeatherDreamer is scored as its ensemble mean as is standard for ensemble systems, which gives it some advantage over GSI's single deterministic background.

A practical constraint on scaling WeatherDreamer is that compute and data loading grow with the number of observing systems. 
The harmonized fields of Section \ref{sec:limb} offer a way around this. 
Decoded at nadir, all geostationary imagers, or all microwave sounders, become a single spatially continuous, limb-corrected field per group, which could replace their individual inputs, either as latent embeddings or as fields decoded into observation space. 
Grouping similar sensors, like the geostationary imager mosaic used in WeatherNext3 \citep{weathernext3_2026} would hold the cost of each step roughly fixed as platforms are added or retired, and would make room for observing systems not yet included, such as hyperspectral sounders.

The latent state may also be useful beyond forecasting. 
It is a compact, gap-filled, hourly representation of everything the observing system has seen, and new quantities could be decoded from it without retraining the
model, especially to target quantities observed sparsely or only by specialized instruments, such as aerosols, turbulence, or renewable-energy variables. 
Because any observing system can be withheld at inference, the model supports inexpensive data-denial experiments, its innovations could flag anomalous or miscalibrated observations for quality control, and the latent state could serve as an observation-driven initial condition or input for other models without passing through a physical analysis.

Data assimilation has long been the step of operational weather prediction that machine learning has not replaced.
The most skillful learned forecast models still start from analyses produced by physics-based systems.
WeatherDreamer shows that the full cycle can instead be learned directly from observations: without reanalysis, it learns how far to trust each instrument, how viewing geometry distorts what each one sees, and how to carry the atmospheric state forward between them.
The result is already competitive with operational systems near the surface, and its shortfalls aloft point to clear remedies in larger spatial context and better use of upper-air observations.
At about 1.5 minutes per global update on a single GPU, a learned cycle that closes that gap would make hourly, observation-driven global analyses and forecasts practical at a fraction of today's cost.

\section*{Open Research Section}
\noindent The observations used to train and evaluate WeatherDreamer are publicly available.
Geostationary infrared radiances are from GOES-16/-18/-19 ABI \citep{schmit2005introducing},
Meteosat SEVIRI \citep{schmetz2002introduction} and GK2A AMI \citep{kim2021introduction}.
Microwave radiances are from ATMS on NOAA-20 and Suomi NPP \citep{weng2012atms}
and AMSU-A on MetOp-B/C \citep{mo1996amsua}; infrared radiances from HIRS/4 on
NOAA-18/19 \cite{smith1979tovs, burgdorf2020hirs}; ocean surface winds from ASCAT \citep{figa2002advanced};
and radio occultation bending angles from GRAS \citep{luntama2008prospects}.
Cloud-top height is from the CloudSat 2B-GEOPROF product \citep{mace2014rlgeoprof}.
Surface station observations are from the Integrated Surface Database \citep{lott2004integrated}
and radiosonde profiles from the Integrated Global Radiosonde Archive \citep{durre2006overview}.
Precipitation is from IMERG Final Run V07B \citep{huffman2020gpm} and surface irradiance
from NSRDB v4 \citep{sengupta2018national}. Surface elevation is from the Copernicus
90 m digital elevation model \citep{copernicus2021dem}.

Verification uses irradiance observations from USCRN \citep{diamond2013uscrn} and SOLRAD \citep{hicks1996noaa} networks.
Verification also uses EarthCARE AC\_TC\_2B
\citep{illingworth2015earthcare}, GPM MERGIR \citep{janowiak_joyce_xie_2017}, and GOES ACHA
products \citep{goesr_acha_2018}.

Operational baselines were retrieved from public cloud archives that require no registration. 
GFS forecasts (0.25\textdegree{} \texttt{pgrb2.0p25}) are from the NOAA Open Data Dissemination (NODD) program \citep{noaa_gfs_aws} at \url{https://noaa-gfs-bdp-pds.s3.amazonaws.com}.
GEFS v12 forecasts (0.25\textdegree{} \texttt{pgrb2sp25}, control and 30 perturbed members) \citep{noaa_gefs_aws} at \url{https://noaa-gefs-pds.s3.amazonaws.com}.
HRRR forecasts (3 km \texttt{wrfsfc}) \citep{noaa_hrrr_aws} at \url{https://noaa-hrrr-bdp-pds.s3.amazonaws.com}.
IFS ENS forecast data (0.25\textdegree{}, 51 members), \textcopyright{} European Centre for Medium-Range Weather Forecasts (ECMWF), 2025, were retrieved from the ECMWF open-data mirror on AWS at \url{https://ecmwf-forecasts.s3.amazonaws.com} \citep{ecmwf_ifs_ens} and are licensed under CC BY 4.0; we interpolated them to station locations. 
Operational baselines were retrieved for the 50 Set A cycles listed in Table S9, with operational latency rules applied as described in Text S6.2.

NCEP Gridpoint Statistical Interpolation (GSI) observation-space diagnostic files are from the NOAA 3D-Var coupled reanalysis scout run v2 \citep{noaa_gsi_scout_v2}, publicly available without registration through the NOAA Open Data Dissemination program at \url{https://noaa-reanalyses-pds.s3.amazonaws.com/analyses/scout_runs/3dvar_coupledreanl_scoutrun_v2/}.
We used the background (\texttt{ges}) and analysis (\texttt{anl}) diagnostic files contained in the \texttt{radstat} and \texttt{cnvstat} archives for the 30 Set B cycles listed in Table S10.

% A representative subset of the WeatherDreamer analyses and forecasts, comprising the data needed to reproduce all figures and tables, is available at Zenodo via \url{]} under a Creative Commons Attribution 4.0 International (CC BY 4.0) license.
% The complete set of analyses and forecasts (approximately 5 TB) is not publicly archived because of its size.
% The WeatherDreamer model code, inference code, and evaluation and figure scripts are available to reviewers via [private share link].

\section*{Conflict of Interest declaration}
\noindent All authors are employees of Zeus AI, Inc., which develops WeatherDreamer and may use it in commercial products and services. The authors declare no other conflicts of interest.

\acknowledgments
This material is based upon work supported by the U.S. Department of Energy, Office of Science, Office of Biological and Environmental Research, under Award Number DE-SC0024737, the Department of Defense Office of Naval Research under Award Number N6833525C0308, and National Aeronautics and Space Administration (NASA) Award Number 80NSSC25C0437. 
This research used resources of the National Energy Research Scientific Computing Center, a DOE Office of Science User Facility supported by the Office of Science of the U.S. Department of Energy under Contract No. DE-AC02-05CH11231 using NERSC award BER-ERCAP0035070.
Resources supporting this work were also provided by the NASA High-End Computing (HEC) Program through the NASA Center for Climate Simulation (NCCS) at Goddard Space Flight Center. 
Any opinions, findings, conclusions or recommendations expressed in this material are those of the authors and do not necessarily reflect the views of the U.S. Department of Energy or the Office of Naval Research. 
We thank the data providers listed in the Open Research section.

\bibliography{references}

\end{document}

\title{Supporting Information for ``WeatherDreamer: Learning a World Model from Multimodal Earth Observations''}

\authors{Kate Duffy\affil{1}\correspondingauthor{Kate Duffy}{kate@myzeus.ai},
Elise Penn\affil{1},
Tristan Ballard\affil{1},
August Posch\affil{1},
Thomas J. Vandal\affil{1}
}

\affiliation{1}{Zeus AI, Inc., Cambridge, Massachusetts}

% \maketitle

\section*{Contents of this file}
\begin{enumerate}
    \item  Text S1-S6
    \item  Figures S1-S7
    \item  Tables S1-S10

\end{enumerate}

\newpage

\section*{Introduction}
\noindent This supporting information provides details of the observation
preprocessing, forcing fields, tokenizers, decoders and training (Text
S1--S5), the verification methods (Text S6), and additional results,
including full observation-minus-forecast departure statistics against the
NCEP GSI first guess (Tables S1--S7). 

\section{Preprocessing and Gridding}
\label{sec:gridding}
\noindent All gridded and swath observations are remapped to the Hierarchical Equal Area isoLatitude Pixelization (HEALPix) \citep{gorski2005healpix}, which divides the sphere into 12 faces, each subdivided into $N_{\mathrm{side}} \times N_{\mathrm{side}}$ equal-area cells arranged on rings of constant latitude. 
Equal-area cells avoid the polar oversampling of latitude--longitude grids, and the nested hierarchy allows modalities of different native resolution to share one indexing scheme. 
Each modality is stored at a resolution closest to its native sampling, from $N_{\mathrm{side}} = 128$ (about 51~km) to $512$ (about 12~km) (Main Table 1). 
Viewing zenith angle is stored for every satellite observation.
Point observations from ISD, IGRA and CloudSat are not gridded; each is kept at its reported latitude, longitude, time, and, for radiosondes, pressure level.

Observations are assigned to hourly windows centered on the hour. 
The offset of each observation from the hour is stored
and used in the model's continuous time encoding (Section 3.3).
Each variable and channel is standardized using means and standard deviations computed over the training period. 
For radiosonde and GPS-RO profiles, which vary by orders of magnitude with height, standardization is computed by vertical bin (pressure levels and height bins, respectively) and interpolated linearly between bins in continuous decoding application. 
Precipitation is cube-root transformed, and surface irradiance is linearly rescaled.
No additional screening or thinning is applied, so that all observations that pass provider QC are available to the model.

\section{Forcing Fields}
\label{supp:solar}

\noindent To provide the primary thermodynamic driver for diurnal atmospheric cycles, solar geometry and clear-sky irradiance are computed analytically at each grid cell and hourly time step \citep{hartmann2016global, duffie2013solar}. 
For a grid point at latitude $\phi$ and longitude $\lambda$ at time $t$, the solar declination is approximated as $\delta = 23.45^\circ \sin\!\left(\frac{360^\circ}{365}(n - 81)\right)$ for day-of-year $n$. 
Apparent solar time $t_{\mathrm{solar}}$ is obtained from UTC via the longitudinal offset $\lambda/15^\circ$ together with the equation-of-time correction $E(n)$, yielding the local hour angle:
\begin{equation}
h(\lambda, t) = 15^\circ\,(t_{\mathrm{solar}} - 12) .
\end{equation}
The solar zenith angle $\theta_0$ then follows from:
\begin{equation}
\cos \theta_0 = \sin \phi \sin \delta + \cos \phi \cos \delta \cos h ,
\end{equation}
and the solar azimuth $\alpha$ from the corresponding spherical relation. Clear-sky global horizontal irradiance $G_{cs}$ is evaluated with a single-parameter empirical attenuation model \citep{meinel1976applied}:
\begin{equation}
G_{cs} = \max\!\left(0,\; 1098 \, \cos \theta_0 \, \exp\!\left(-\frac{0.057}{\cos \theta_0}\right)\right) \quad [\mathrm{W\,m^{-2}}] ,
\end{equation}
where the exponential term represents bulk atmospheric extinction along the slant path and the leading coefficient absorbs the mean top-of-atmosphere irradiance and clear-sky transmissivity.

The three resulting fields $(\theta_0, \alpha, G_{cs})$ are stored on the HEALPix grid and re-encoded before entering the model. Rather than ingesting the angles directly, where the $2\pi$ wrap in azimuth and the vanishing gradient of $\theta_0$ near the terminator are both poorly conditioned, the solar geometry is presented as a unit direction vector on the sphere, concatenated with the normalized irradiance:
\begin{equation}
f_t = \Big(\sin \theta_0 \cos \alpha, \;\; \sin \theta_0 \sin \alpha, \;\; \cos \theta_0, \;\; \frac{G_{cs}}{G_{\mathrm{ref}}}\Big), \qquad G_{\mathrm{ref}} = 1000\ \mathrm{W\,m^{-2}} ,
\end{equation}
yielding four continuous, everywhere-defined spatial forcing channels whose third component is exactly $\cos \theta_0$.
These channels enter the recurrent state-space model through the action pathway rather than the observation pathway and because they are analytic, they remain available at every step of an autonomous rollout.
Conditioning the transition model on $f_t$ at each imagined step grounds the diurnal forcing of the latent dynamics and suppresses phase drift over forecast horizons.

\section{Hyperspherical Observation Tokenizers}

\noindent The gridded input modalities have spatial resolutions ranging from 6 to 50 kilometers (HEALPix nside 128 to 512) \cite{gorski2005healpix} and spectral bands up to 22 channels.
To manage I/O and during joint multimodal training across many sensors, high-resolution or high-dimensionality gridded modalities are pre-compressed offline using spherical variational autoencoders (SVAE) \cite{davidson2018hyperspherical, xu2018spherical} with vision transformer encoders and Local Implicit Image Functions (LIIF) coordinate-based decoders \citep{chen2021learning}.

\noindent The SVAE encoder partitions an input tensor $\mathbf{x}$ into non-overlapping spatial patches (e.g., $2\times 2$, $4\times 4$, or $8\times 8$) and processes them via a transformer encoder. 
Rather than enforcing a standard Gaussian latent space, the network projects tokens to a hyperspherical manifold $\mathbb{S}^{D-1}$ governed by a von Mises-Fisher (vMF) posterior distribution $\text{vMF}(\boldsymbol{\mu}, \kappa)$ \citep{davidson2018hyperspherical}. 
The projection head emits a unit-norm directional mean vector $\boldsymbol{\mu}$ and a non-negative concentration parameter $\kappa$:  

$$q(\mathbf{z} \mid \mathbf{x}) = C_d(\kappa) \exp\left( \kappa \boldsymbol{\mu}^T \mathbf{z} \right)$$

\noindent where $C_d(\kappa)$ is the $d$-dimensional normalization constant evaluated using scaled Bessel functions. 
Continuous latents are reparameterized via Wood/Davidson rejection sampling $\mathbf{z} \sim \text{vMF}(\boldsymbol{\mu}, \kappa)$ \citep{davidson2018hyperspherical}, allowing gradients to flow cleanly through $\boldsymbol{\mu}$. 
Training follows the standard VAE objective \citep{kingma2013auto} with a reconstruction term consisting of masked mean-squared-error and cosine-similarity losses computed in pixel space, plus a KL divergence between the vMF posterior and a uniform prior $U(\mathbb{S}^{D-1})$ on the sphere:  
$$\mathcal{L}_{\text{SVAE}} = \mathcal{L}_{\text{MSE}}(\hat{\mathbf{y}}, \mathbf{y}) + \mathcal{L}_{\text{cos}}(\hat{\mathbf{y}}, \mathbf{y}) + \beta_{\text{KL}} \mathcal{D}_{\text{KL}}\left( \text{vMF}(\boldsymbol{\mu}, \kappa) \,\Vert\, U(\mathbb{S}^{D-1}) \right)$$
To prevent posterior collapse ($\kappa \to 0$), the concentration head is initialized with a positive bias guard $\kappa_{\text{init}} \approx 5.0$, and the KL weight $\beta_{\text{KL}}$ is activated after an initial reconstruction warm-up.
Compared to Gaussian latent spaces, we found that hyperspherical tokenization achieves $5\times\text{--}10\times$ higher compression at matched reconstruction loss.
To handle the missing data that is common in many modalities due to irregular or incomplete spatial coverage, missing data tokens are masked in attention and excluded from the reconstruction loss \citep{he2022masked}.

\noindent Once trained, tokenizers are frozen. Each tokenized modality is encoded offline and its latent representations, stored as posterior means, are written to disk as HEALPix Zarr archives, which serve as the pre-tokenized inputs to WeatherDreamer. 
To render latent states back into physical grids without patch artifacts or resolution constraints, we replace conventional linear or patch decoders with a local implicit coordinate decoder (LIIF-style) for continuous rendering \citep{chen2021learning}, detailed in Text S4.1.

\begin{figure}[!htbp] 
\noindent\includegraphics[width=\textwidth]{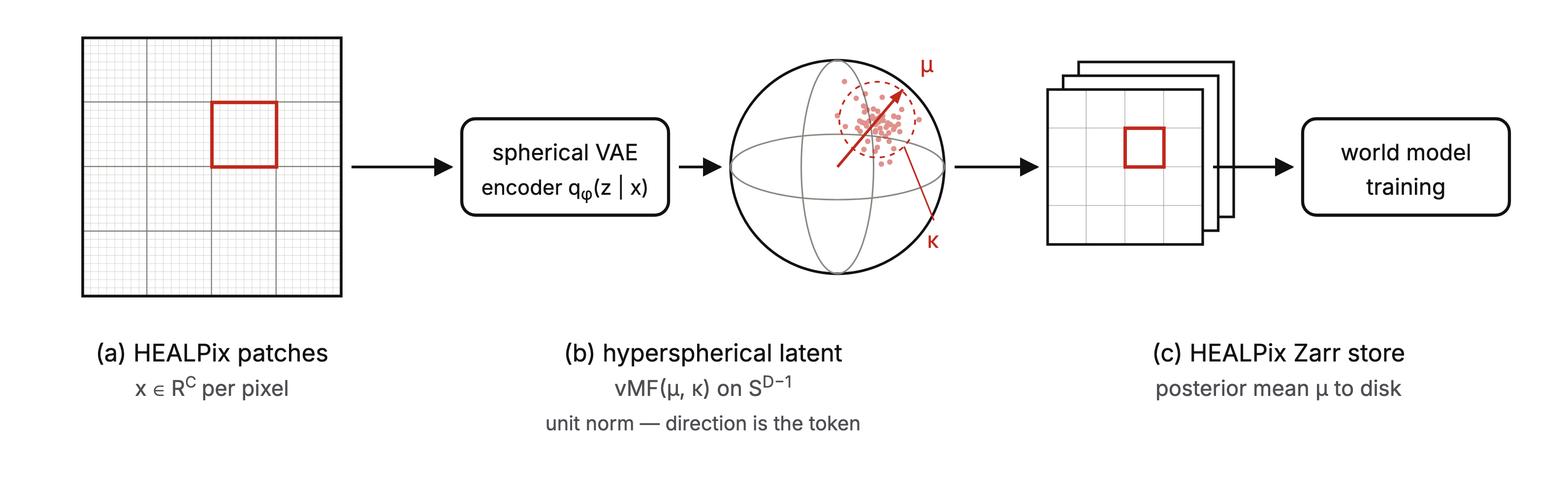}
\caption{\textbf{Offline hyperspherical token compression pipeline.} 
Used for the geostationary infrared imagers (GOES-East, GOES-West, GK2A and SEVIRI). 
All other gridded systems are patch-embedded directly. 
(a) Observation datasets are partitioned into localized HEALPix spatial patches. 
(b) The Spherical VAE encoder $q_\phi$ projects features onto a unit hyperspherical manifold $\mathbb{S}^{D-1}$ parameterized by a von Mises-Fisher distribution $\text{vMF}(\boldsymbol{\mu}, \kappa)$. 
(c) Directional posterior means $\boldsymbol{\mu}$ are archived directly to disk as compressed HEALPix Zarr stores, serving as pre-tokenized inputs for world model training.
  }
\label{tokenization}
\end{figure}

\section{Decoders}
\subsection{Local Implicit Coordinate Decoders (LIIF)}
\label{sec:decoders_liff}
\noindent To render continuous, arbitrary-resolution physical fields without patch artifacts or grid constraints, WeatherDreamer employs a Local Implicit Image Function (LIIF) decoder \citep{chen2021learning}. 
For any spatio-temporal query coordinate $q = (t, y, x) \in \mathbb{R}^3$, the decoder identifies the nearest latent feature grid token $z_k$ and extracts context from its local $3 \times 3$ neighborhood \citep{chen2021learning}. 
The relative coordinate offset $\delta x = q_{\text{spatial}} - x_k$ and query cell dimensions $s = (\Delta y, \Delta x)$ are mapped into a high-frequency representation using a Fourier feature encoder $\gamma(\cdot)$ \citep{tancik2020fourier, mildenhall2021nerf}:
\begin{equation}
\gamma(\mathbf{v}) = \left[ \sin(2^0 \pi \mathbf{v}), \cos(2^0 \pi \mathbf{v}), \dots, \sin(2^{L-1} \pi \mathbf{v}), \cos(2^{L-1} \pi \mathbf{v}) \right]^T
\end{equation}
where the scale parameter $L$ controls the maximum spatial frequency resolution \citep{tancik2020fourier}. The continuous field value $\hat{y}(q)$ at query point $q$ is evaluated via a Multi-Layer Perceptron ($\text{MLP}_\theta$) using area-proportional bilinear ensemble weighting across the four neighboring latent cells $k \in \{1, 2, 3, 4\}$ \citep{chen2021learning}:
\begin{equation}
\hat{y}(q) = \sum_{k=1}^4 \frac{A_k}{A_{\text{total}}} \cdot \text{MLP}_\theta \left( \left[ z_k, \gamma(\delta x_k), \gamma(s) \right] \right)
\end{equation}
where $A_k$ represents the diagonal overlap area between query point $q$ and latent cell $k$ \citep{chen2021learning}.
By parameterizing fields as continuous functions rather than discrete pixel arrays, this implicit formulation enables WeatherDreamer to decode physical state variables at arbitrary spatial resolutions and continuous off-grid coordinates without requiring re-gridding or spatial interpolation \citep{chen2021learning}.

\begin{figure}
\noindent\includegraphics[width=\textwidth]{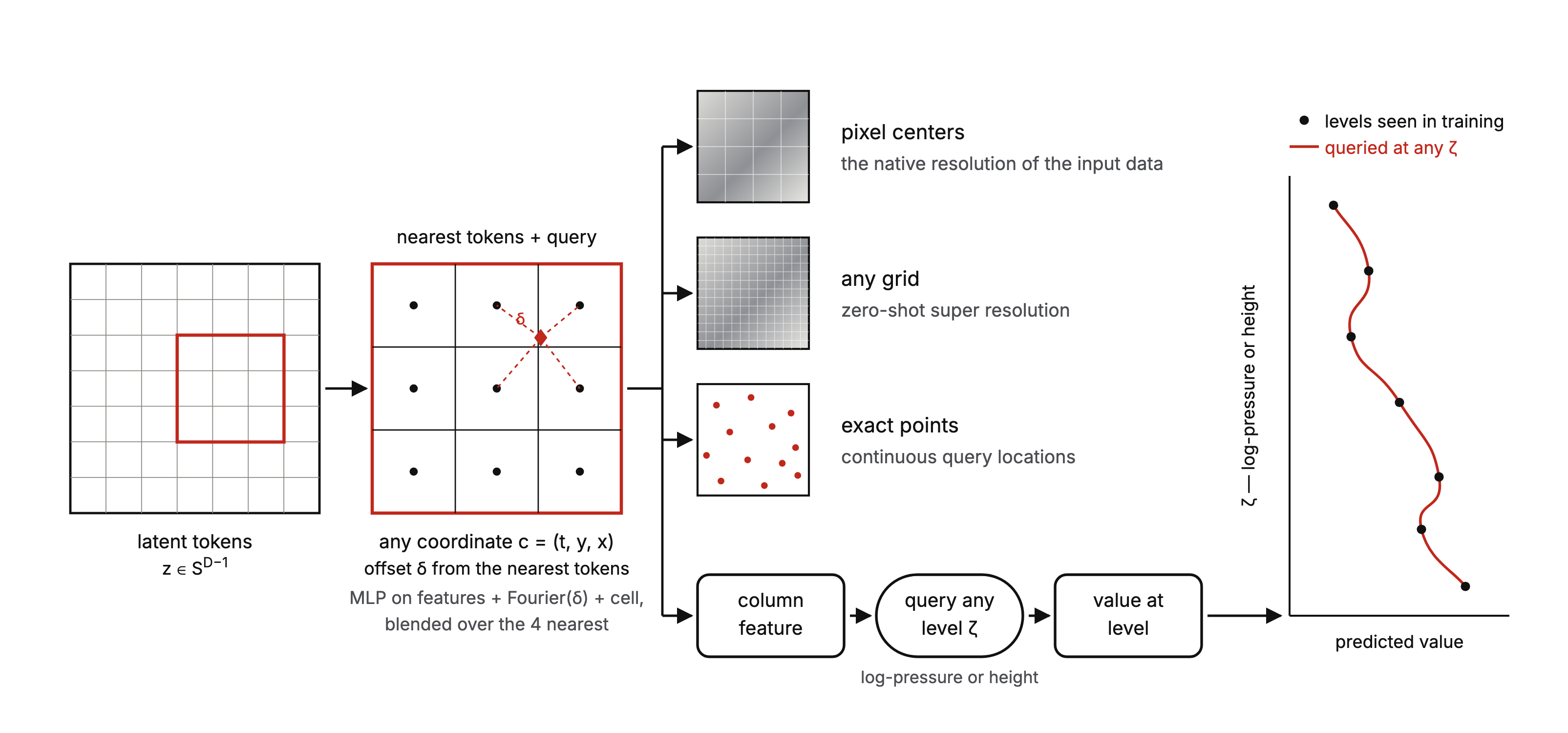}
\caption{
\textbf{Continuous coordinate decoding pipeline.} 
Latent tokens $z \in \mathbb{S}^{D-1}$ are queried at any continuous spatiotemporal coordinate $c = (t, y, x)$. The LIIF decoder \citep{chen2021learning} gathers neighboring token features, applies Fourier relative-offset encodings, and bilinearly blends MLP predictions to guarantee spatially smooth, patch-free representations. The single flexible architecture can simultaneously render native-resolution grids, zero-shot super-resolved fields, exact station point locations, or 3D vertical columns queried at arbitrary vertical levels ($\zeta$).
}
\label{fig:decoding}
\end{figure}

\subsection{Vertical Profile Decoders}
A primary horizontal decoder projects column feature vectors at continuous surface position $(t, y, x)$ \citep{chen2021learning}. 
A secondary vertical MLP then evaluates queries along measured vertical coordinates $p$ using continuous log-pressure coordinates $\phi(p) \in [-1, 1]$ \citep{tancik2020fourier, mildenhall2021nerf}:
\begin{equation}
\hat{y}_{\text{profile}}(t, y, x, p) = \text{MLP}_{\text{vert}} \left( \left[ \text{MLP}_{\text{horiz}}(z(t,y,x)), \gamma(\phi(p)) \right] \right)
\end{equation}
Evaluating loss at exact $3\text{D}$ instrument coordinates allows the unified state to maintain continuous vertical thermodynamic stratification without vertical grid interpolation.

\subsection{Modeling Zero-Inflated Fields}

\noindent For zero-inflated physical variables such as IMERG precipitation rate or cloud top height, standard $L_2$ regression causes feature smoothing and underpredicts extreme events \citep{ravuri2021skilful}. 
To preserve physical intensity, zero-inflated fields utilize a dual-head discrete-continuous coordinate decoder emitting occurrence logits $p(q)$ and magnitude values $m(q)$:
\begin{equation}
\hat{y}_{\text{precip}}(q) = \sigma(p(q)) \cdot \text{ReLU}(m(q))
\end{equation}
where $\sigma(\cdot)$ is the sigmoid function. The occurrence head is trained using Binary Cross-Entropy ($\mathcal{L}_{\text{BCE}}$) across all spatial queries, while magnitude regression is evaluated strictly over non-zero ground-truth pixels ($y(q) > 0$):
\begin{equation}
\mathcal{L}_{\text{precip}} = \mathcal{L}_{\text{BCE}}(\sigma(p(q)), \mathbb{I}_{y(q) > 0}) + \mathbb{I}_{y(q) > 0} \cdot \left( \| m(q) - y(q) \|_1 + \| m(q) - y(q) \|_2^2 \right)
\end{equation}
During rollout steps, the magnitude term is replaced by the ensemble CRPS, while the occurrence BCE is unchanged.
To anchor the initial state to global climatological baselines, the logit layer output bias is initialized to $b_0 = -2.0$ for IMERG precipitation (matching global rain fraction $P \approx 0.12$) and $b_0 = 0.79$ for CloudSat cloud-top fields (matching the observed cloudy fraction $P \approx 0.69$).

\section{Training}
\subsection{Ensemble Noise Construction}
\noindent For each member and step we draw a 32-dimensional field of i.i.d. standard normal values on a coarse $8\times8$ grid over the $32\times32$ latent patch, corresponding to one coarse cell per $4\times4$ latent cell. 
Each latent node takes a Gaussian-weighted combination of its 16 nearest coarse centers, with length scale of $1.3 \times$ the coarse spacing and weights normalized so that $\sum_k w^2_k =1$  which preserves unit variance. 
Successive rollout steps are correlated by an AR(1) process, $\xi_{t+1} = \rho \xi_t + \sqrt{1 - \rho^2} \eta_{t+1}, \quad \rho = 0.9$ so that a member's perturbation persists instead of jittering hour to hour. 
The resulting field $\epsilon_t$ modulates every message-passing round of the dynamics cell and the prior network by feature-wise linear modulation (FiLM), $x \leftarrow x(1+\gamma(\epsilon_t)) + \beta(\epsilon_t)$.

\subsection{Optimization Details}
\label{sec:optim}
\noindent All components are optimized jointly with AdamW using two parameter groups: the dynamics (recurrent cell and prior network) at a peak learning rate of $2 \times 10^{-3}$ and all other modules at $2 \times 10^{-4}$, with weight decay 0.05 and gradient clipping at 1.0. 
The learning rate warms up linearly over 750 steps, is constant until step 35,000, and then decays on a cosine schedule to 0.1$\times$ its peak at step 50,000 and is held constant thereafter.
Training runs for 70,000 steps with a global batch of 32 twelve-hour windows (plus a six-hour rollout, advancing to twelve hours for the last 10,000 steps), sampled as 32×32-latent patches from all 12 HEALPix faces, on 4 NVIDIA GH200 GPUs (4 samples per GPU with 2 batch gradient accumulation). 
As a final step, the model is frozen and only the CloudSat head is fine-tuned on the temporal period of CloudSat coverage, 2020, for 3,000 steps.
We keep an exponential moving average of the trainable weights (decay 0.999) beginning from step ~28,000, and all reported results at step 70,000 use the EMA weights.

\subsection{CloudSat Head Fine-Tuning.} 
CloudSat/CALIPSO cloud-top heights are sparse (a single nadir curtain per orbit) and limited to one year of our training period, so during joint training the cloud-top decoder receives few supervised examples relative to the dense modalities. 
After the main training, we fine-tune only the CloudSat decoder head with all other parameters frozen, so the latent state and every other output are unchanged. 
CloudSat is always withheld as an input and its loss is applied as a cross-modal (unobserved-modality) reconstruction, so the head learns to infer cloud-top height from the remaining observations (principally geostationary IR), which matches how it is used at inference. 
We sample 12-h training windows from 2020 that contain at least 100 CloudSat observations, and train for 3000 steps with AdamW (learning rate $2 \times 10 ^{-4}$, 100 warm-up steps, no decay, weight decay 0.05), batch size 8 (2 GH200 GPUs × 4), with the same rollout and CRPS objective as the main training and an EMA of the head weights (decay 0.998). 
The EMA weights are used at inference. 

\section{Verification Methods}

\subsection{Evaluation Sets}

\noindent WeatherDreamer is trained on 2020--2023, and 2024 and 2025 are used for testing.  
We use two sets of inference runs. 

\emph{Set A} comprises 50 initializations between 20 August 2024 and 15 August 2025, run with operational latency: 12~h of assimilation followed by 12-h forecasts, with fields decoded at nadir.
Initialization times are distributed uniformly across UTC hours (two or three per hour), and the hours are permuted across dates so that time of day is not aligned with season.
Set A supplies the forecast evaluations against surface stations, radiosondes, precipitation, irradiance and cloud-top height. 
Not every reference dataset covers the whole period, so each experiment uses the subset of initializations for which its reference is available.

\emph{Set B} is designed for comparison with the NCEP GSI first guess: 30 GSI cycles between 3 January and 29 October 2024 (the public diagnostic archive ends on 12 November 2024), each initialized 6~h before the cycle time after 12~h of assimilation. 
Like GSI, which applies no latency cutoff, it assimilates without latency, and its 9-h forecasts are decoded at each observation's actual viewing geometry, as GSI's radiative transfer does.
Set B supplies the observation-space departure comparison and the assimilation diagnostics.

\subsection{Operational Latency for WeatherDreamer and Baselines}

To mimic operational latency in set A, polar-orbiting sounder and imager radiances, ASCAT winds and GPS radio occultation are withheld for the final 2h before initialization, and radiosondes for the final 4h.
We refer to the state at the initialization time as the \emph{analysis} (lead 0) and to later states as forecasts at lead $+k$~h. 
Forecasts consist of a zero-noise control and 8 perturbed members, all from the EMA weights. 
Deterministic scores use the ensemble mean unless stated
otherwise.

Operational baselines are presented with operational latency: each uses the newest cycle that would have been disseminated by the initialization time, and its forecast is read at the lead that is valid at each verification time. 
For the 6-hourly global systems this is the cycle at least 4~h old for GFS (0.25$^\circ$, hourly output), 5~h old for GEFS (31 members, 0.25$^\circ$) and 7~h old for IFS ENS (51 members, 0.25$^\circ$ open data); HRRR (3~km, hourly cycles) is taken from the cycle 2~h before initialization. 
GEFS and IFS ENS fields, available every 3~h, are linearly interpolated in time to hourly verification times, except for irradiance and 3-h precipitation, which are scored on their native 3-h windows.

\subsection{Statistical Conventions}
Unless stated otherwise, scores are pooled over all valid observation--lead pairs, and all systems in a comparison are scored on the identical set of pairs. 
Uncertainty intervals are 95\% bootstrap intervals from resampling initializations (2000 resamples for precipitation, 10{,}000 paired resamples for irradiance differences); a difference between systems is called significant when its paired interval excludes zero. 
Set-B departure statistics are computed per cycle and summarized by the median over cycles (Section~\ref{sec:setb}).

\subsection{Verification of Set A Forecasts Against Observations}
\label{sec:supp_setA_verif}

\paragraph*{Surface stations.}
WeatherDreamer surface forecasts are decoded at the exact location of each station in the ISD catalogue (11{,}065 stations) by the point decoder on which station observations are trained. 
Each forecast at lead $L$ is compared with the station report in the hourly bin valid at the initialization time plus $L$\,h. 
Held-out stations (534, 5\% of the network) are never inputs at inference; in-sample stations are inputs up to the initialization time only. GEFS and IFS ENS are interpolated bilinearly to the station locations and linearly in time between their 3-hourly outputs, using the most recent cycle available at initialization (Text S6.2). 
In every panel all systems are scored on identical station--lead pairs: a pair is kept only if every system has a valid value there. Sea-level pressure from ISD is converted from tenths of hPa; temperatures are in K or $^\circ$C as reported, and wind speed in m\,s$^{-1}$.

\paragraph*{Metrics.}
We use the fair continuous ranked probability score \citep{ferro2014fair},
\[
\mathrm{CRPS}_f = \frac{1}{M}\sum_{i=1}^{M}|x_i-y| - \frac{1}{2M(M-1)}\sum_{i=1}^{M}\sum_{j=1}^{M}|x_i-x_j|,
\]
for an ensemble of $M$ members $x_i$ and observation $y$, with $M=8$ for WeatherDreamer (perturbed members; the control is excluded), 31 for GEFS and 51 for IFS ENS. 
Deterministic scores (MAE, bias $=$ forecast $-$ observation) use the ensemble mean. 
The spread--skill ratio is $\sqrt{(M+1)/M}\,\big(\sum \sigma^2/\sum e^2\big)^{1/2}$, where $\sigma^2$ is the ensemble variance (unbiased) and $e$ the ensemble-mean error, pooled over all station--lead pairs at a given lead; a value of 1 indicates a calibrated ensemble. 
No observation-error term is added to any ensemble. 
At the analysis time (lead 0) WeatherDreamer's value is its single deterministic posterior decode.
No lapse-rate adjustment is applied to GEFS or IFS ENS for differences between model and station elevation.

\subsection{Precipitation}
Precipitation analyses and forecasts are verified against IMERG Final Run V07B over the 50 initializations, with half-hourly rates averaged in pairs to hourly. 
IMERG Final is a training target but is withheld at inference, so WeatherDreamer's precipitation is inferred entirely from the other observing systems. 
The model value at valid time $t$ is the mean rate over the hour ending at $t$. Lead 0 is the hour ending at initialization (the analysis hour), and leads +1 to +12\,h are forecasts with no further observations.

To exclude cold, mixed-phase and snowfall regimes, we score only pixels between 60$^\circ$S and 60$^\circ$N in which at least 90\% of the contributing 0.1$^\circ$ IMERG cells are classified as liquid in every half-hour of the verification period (two for hourly scores, six for 3-h accumulations). 
IMERG is averaged onto the model's HEALPix grid at $N_{\mathrm{side}}=256$ (about 25\,km), and all scores except the power spectra are computed there.

WeatherDreamer's precipitation head predicts a rain probability and a conditional amount. 
Each member's expected rate is their product, used without calibration;
deterministic scores use the ensemble mean of these expected rates.
All baselines are taken from the most recent cycle available at initialization.

\textit{Hourly scores.} Hourly rates are compared with GFS (0.25$^\circ$, hourly mean rate). 
The equitable threat score is frequency-matched: each forecast is thresholded at the quantile of its own field that reproduces the observed frequency of the event, so its frequency bias is one by construction and the score measures placement alone. 
Normalized zonal power spectra of the ensemble-mean and single-member field are computed along latitude rows of a 0.25$^\circ$ grid between 20$^\circ$S and 20$^\circ$N.

\textit{3-h accumulations.} Accumulations are formed in windows
$(t_0+L-3\,\mathrm{h},\,t_0+L]$, with $L=3,6,9,12$\,h. 
WeatherDreamer members are summed over the three hourly leads. 
GEFS (31 members) and IFS ENS (51 members) accumulations are interpolated in time where the window does not coincide with their 3-hourly output.
Ensembles are compared with fair CRPS and with the Brier skill score for $\ge 1$\,mm relative to the sample climatology (Fig. S5c). 
Reliability is pooled over the four windows in 11 forecast-probability bins of width 0.1.
Baselines use the most recent cycle available at initialization.

\subsection{Surface Irradiance}
Global horizontal irradiance is verified at 116 USCRN and 7 SOLRAD stations. 
Forecasts are scored as 3-h means in windows ending at multiples of 3\,h UTC so that no system is interpolated in time; at the analysis hour, the 1-h mean centred on the initialization is used. 
WeatherDreamer is sampled at the nearest grid cell, HRRR (cycle two hours before initialization) from its 15-min averages, and GEFS and IFS ENS from their 3-hourly accumulations. 
Only windows with daylight are scored. 
For deterministic HRRR the CRPS reduces to the absolute error. 
Differences in CRPS are tested with a paired bootstrap over initializations.

\subsection{Cloud-Top Height}
Cloud-top height is verified against EarthCARE (ESA L2 products AC\_\_TC\_\_2B and ATL\_CTH\_2A, baseline BA). 
For each set-A analysis hour, EarthCARE frames within $\pm 30$\,min are used; each footprint is matched to the WeatherDreamer analysis at the nearest cell, to the GOES-East ABI cloud-top height (ACHA) from the scan nearest the overpass, and, within the CONUS domain, to the HRRR analysis. 
The EarthCARE reference is the height of the highest bin classified as cloud (AC\_\_TC\_\_2B classes 8--21; precipitation-only, insect and aerosol classes are excluded). 
WeatherDreamer's cloud-top height is the conditional height divided by the cloud probability, $h/\sigma(p)$. 
Height errors are computed only where EarthCARE detects cloud, on footprints where every compared system provides a value; cloud detection is scored separately with probability of detection, false-alarm ratio and equitable threat score, counting WeatherDreamer as cloudy where $\sigma(p)\ge 0.5$.

\subsection{Viewing-Angle Analyses}
To isolate the learned angle correction, the saved analysis state of each set-A initialization is decoded again with only the viewing-angle label changed: at nadir, at the observing satellite's own angle, or at another satellite's angle. 
These re-decodes reproduce the original decodes to within 0.05\,K on average. 
Collocated GOES-East and GOES-West views are averaged on 0.25$^\circ$ boxes over ocean between 60$^\circ$S and 60$^\circ$N, keeping boxes that are clear in the ABI Level-2 clear-sky mask in both views, together with their eight neighbours, and spatially uniform at 11.2\,\textmu m. 
In each box one satellite views at $\theta<25^\circ$ and the other at 47--80$^\circ$. 
The inter-satellite calibration offset, measured on boxes viewed at near-equal angles ($|\Delta\theta|<2^\circ$), is removed before forming the near-nadir minus oblique difference. 
Biases against GPM MergIR are computed at 4\,km against each imager's 11\,\textmu m band, binned by viewing zenith angle. 
For the microwave sounders, crossovers are pairs of NOAA-20 and Suomi-NPP observations of the same cell within 1\,h whose zenith angles differ by more than 20$^\circ$, and the inter-satellite difference is regressed on the difference in $\sec\theta$.

\subsection{Verification of Set B Forecasts Against GSI Reference}
\label{sec:setb}
As a reference we use the background (first-guess) departures of the NCEP Gridpoint Statistical Interpolation (GSI) system, taken from its observation-space diagnostic files.
These files are publicly available for 2024 from the NOAA coupled-reanalysis scout run (GSI 3D-Var), which assimilates essentially the operational observing system. 
Equivalent operational files are not publicly archived.
We use 30 cycles between 3 January and 29 October 2024, seven or eight at each of 00, 06, 12 and 18\,UTC. 
2024 was not used for training.

For each cycle we run WeatherDreamer from an initialization 6h before the cycle time, assimilating the preceding 12h of observations without operational latency (GSI applies no latency cutoff either), and forecast 9\,h ahead. 
Forecasts are decoded at each observation's actual viewing geometry, as GSI's radiative transfer model does. 
The GSI background is a 3-9h forecast from the previous cycle, valid at each observation's time. 
Correspondingly, each observation is compared with the WeatherDreamer forecast valid at the nearest hour, i.e. lead times of 3-9h.

The systems are evaluated on identical observations. 
For every observation in the diagnostic files we sample the WeatherDreamer forecast at the nearest HEALPix pixel and form $O-F$ using GSI's own observed value $O$.
WeatherDreamer forecasts are the ensemble mean of the control and 8 perturbed members. GSI departures are taken before variational bias correction, so neither system is bias-corrected, and both use the sign convention $O-F$. 
Physically implausible values (including the $10^{11}$ fill values present in the files) are removed by range checks applied identically to both systems.

\paragraph*{Screening and matching follow these procedures:}
\begin{itemize}
  \item Radiances: all satellite zenith angles, over ocean (water fraction $>0.99$). For AMSU-A and HIRS/4,
        observations are kept only where WeatherDreamer's input data contained the same pixel in the same
        hour, since elsewhere the forecast carries no viewing-angle information.
  \item Radiosondes: temperature, humidity and wind interpolated in log-pressure from the 11 decoded levels (1000–100 hPa) to each reported pressure; results are binned to the nearest standard level.
  \item Surface reports: matched to the nearest ISD station within 10\,km and decoded at that station;
        results are given separately for stations held out from WeatherDreamer's input and for in-sample stations.
  \item Temperature: GSI radiosonde and surface temperatures are virtual temperatures, so WeatherDreamer
        temperature is converted to virtual temperature using its own humidity forecast.
  \item GPS-RO: each occultation is interpolated to the decoded impact heights (5, 10, 20, 30\,km),
        log-linearly in bending angle; departures are fractional, $100\,(O-F)/O$.
  \item ASCAT: 10-m wind components over the ocean.
\end{itemize}

\paragraph*{Table conventions.}
All tables in this section share the following conventions unless the caption states otherwise.
Departures are $O-F$ (observation minus forecast), so a positive bias means the forecast is too low.
``GSI'' denotes the NCEP background departure and ``WD'' the WeatherDreamer departure of the ensemble mean (control plus eight perturbed members) on the same observations. 
Bias and standard deviation (SD) are computed per cycle and summarized by the median over the 30 cycles. 
$N$ is the total number of matched observations over all cycles. 
The ratio is WD SD divided by GSI SD, so values below 1 mean WeatherDreamer fits the observations more closely. ``All observations'' columns use every observation in the diagnostic files; ``Assimilated by GSI'' columns repeat the same statistics on the subset that passed GSI quality control and was used in its analysis. 
Radiance channels are identified by instrument channel number and central frequency (microwave, GHz) or wavelength (infrared, \textmu m). 
Units are K for brightness and virtual temperature, g\,kg$^{-1}$ for specific humidity, m\,s$^{-1}$ for wind and percent for GPS-RO bending angle. 
Lead times are 3--9\,h, matched to each observation's time.

\subsection{Assimilation Diagnostics}
For each set-B analysis hour we compare the observation $O$, the one-hour prior $B$ and the posterior $A$, both decoded at the observation's viewing geometry. 
The Desroziers estimate of the observation-error variance is $\sigma_O^2=\mathrm{cov}(O-A,\,O-B)$ and the gain is the regression slope $k=\mathrm{cov}(A-B,\,O-B)/\mathrm{var}(O-B)$, computed per cycle and channel and summarized by the median over cycles.
As an independent estimate (Hollingsworth--Lönnberg), the covariance of $O-B$ between pairs of observations from the same hour is binned by separation (25\,km bins to 500\,km); a second-order autoregressive (SOAR) function $s^2(1+r/L)e^{-r/L}$ is fitted to bins with $r\ge 40$\,km, its value at $r=0$ giving the spatially correlated prior-error variance $\sigma_B^2$ and the remainder of $\mathrm{var}(O-B)$ giving $\sigma_O^2$; the optimal gain is $k_{\mathrm{opt}}=\sigma_B^2/(\sigma_B^2+\sigma_O^2)$.
Clear-sky ATMS observations over ice-free ocean ($|\phi|<50^\circ$, no land within $5\times5$ pixels, satellite zenith $\le 63^\circ$) are selected with a cloud-liquid-water retrieval and scattering index (CLW $<0.05$\,mm, SI $\le 9$; \citealp{grody2001determination}). 
Clear-sky HIRS/4 observations require $\ge 98\%$ of the overlapping NOAA ABI Level-2 clear-sky-mask pixels (GOES-16 or GOES-18, whichever views the cell at the smaller zenith angle) to be classified clear. 
ASCAT winds over ocean with $|\phi|<60^\circ$ are used without cloud screening.
Reference errors are on-orbit ATMS noise \citep{liu2019}, scaled by 0.8--1.0 for the averaging of about 1.7 footprints per 25\,km cell; HIRS/4 specified noise \citep{noaaklm2014}, converted to brightness temperature for 250--280\,K scenes; and the triple-collocation retrieval error of the 12.5\,km ASCAT product \citep{verspeek2019}.

For in situ observations, which are assimilated hourly from fixed sites, the prior is not independent of each station's own previous report, so we evaluate the update at held-out stations instead: the ratio $\mathrm{SD}(O-A)/\mathrm{SD}(O-B)$ measures how much assimilating the surrounding network improves the state at unobserved sites, and is also binned by the distance to the nearest assimilated station (0--25, 25--50, 50--100, 100--200 and $>200$\,km).
For radiosondes the gain $k$ is computed per mandatory level from the analysis hours at 00 and 12\,UTC.

\begin{figure}[htbp]
\noindent\includegraphics[width=\textwidth]{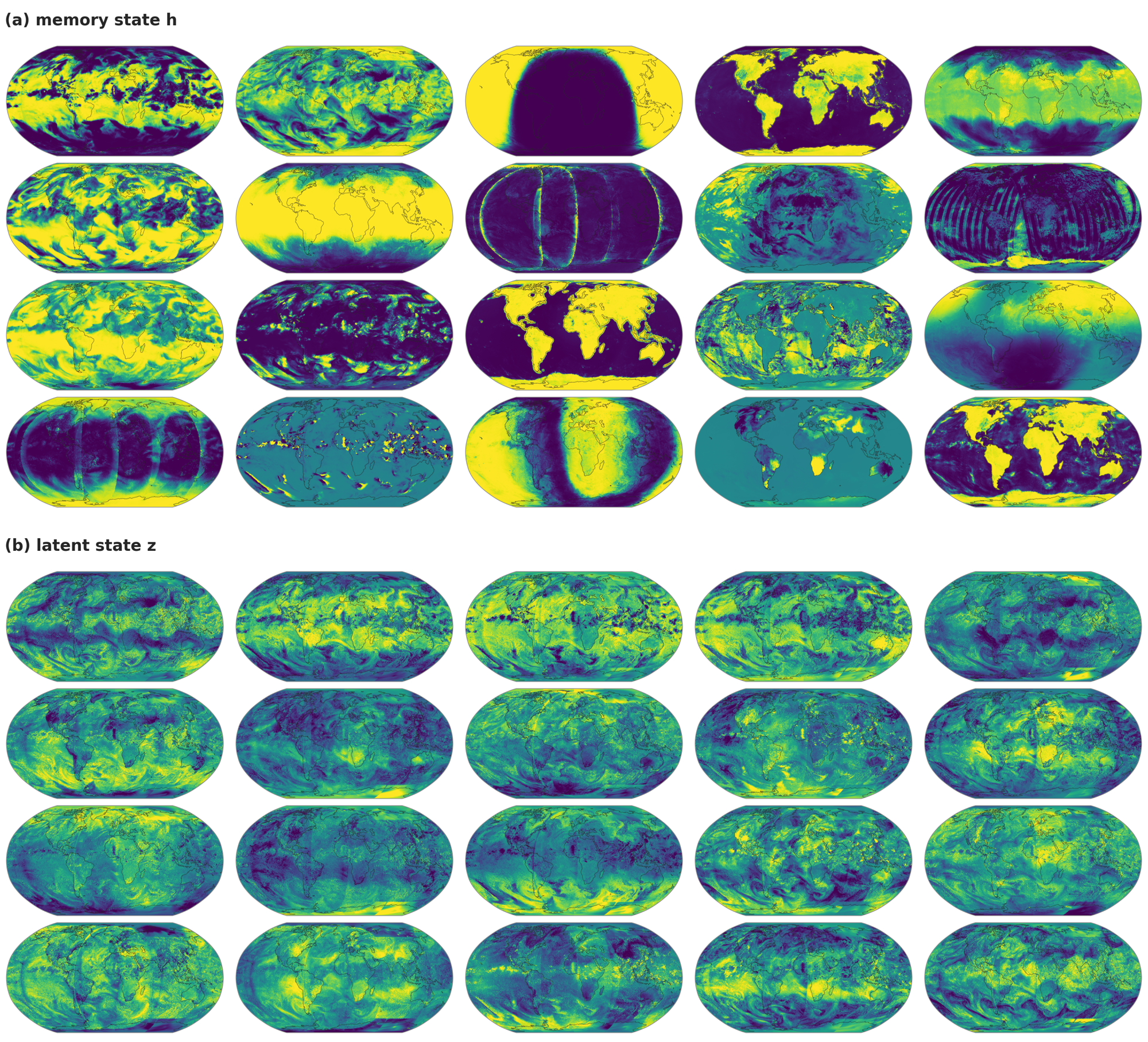}
\caption{
\textbf{Memory and latent state of the world model at one analysis timestep.} 
Channels of (a) the recurrent memory $h_t$ (dim 512) and (b) the latent state $z_t$ (dim 64) read by the decoders at 20 August 2024 00 UTC.
The memory carries forcing (solar terminator, land–sea contrast), meteorological structure, and a record of the observing system, such as geostationary disk footprints and polar-orbiter swaths, analogous to the observation-density information a conventional filter carries in its error covariance. 
The latent state is dominated by meteorological structure with a weak residual imprint at coverage boundaries.
Each panel is colored over its own 1st–99th percentile range.
}
\label{fig:h}
\end{figure}

\begin{figure}[htbp]
\noindent\includegraphics[width=\textwidth]{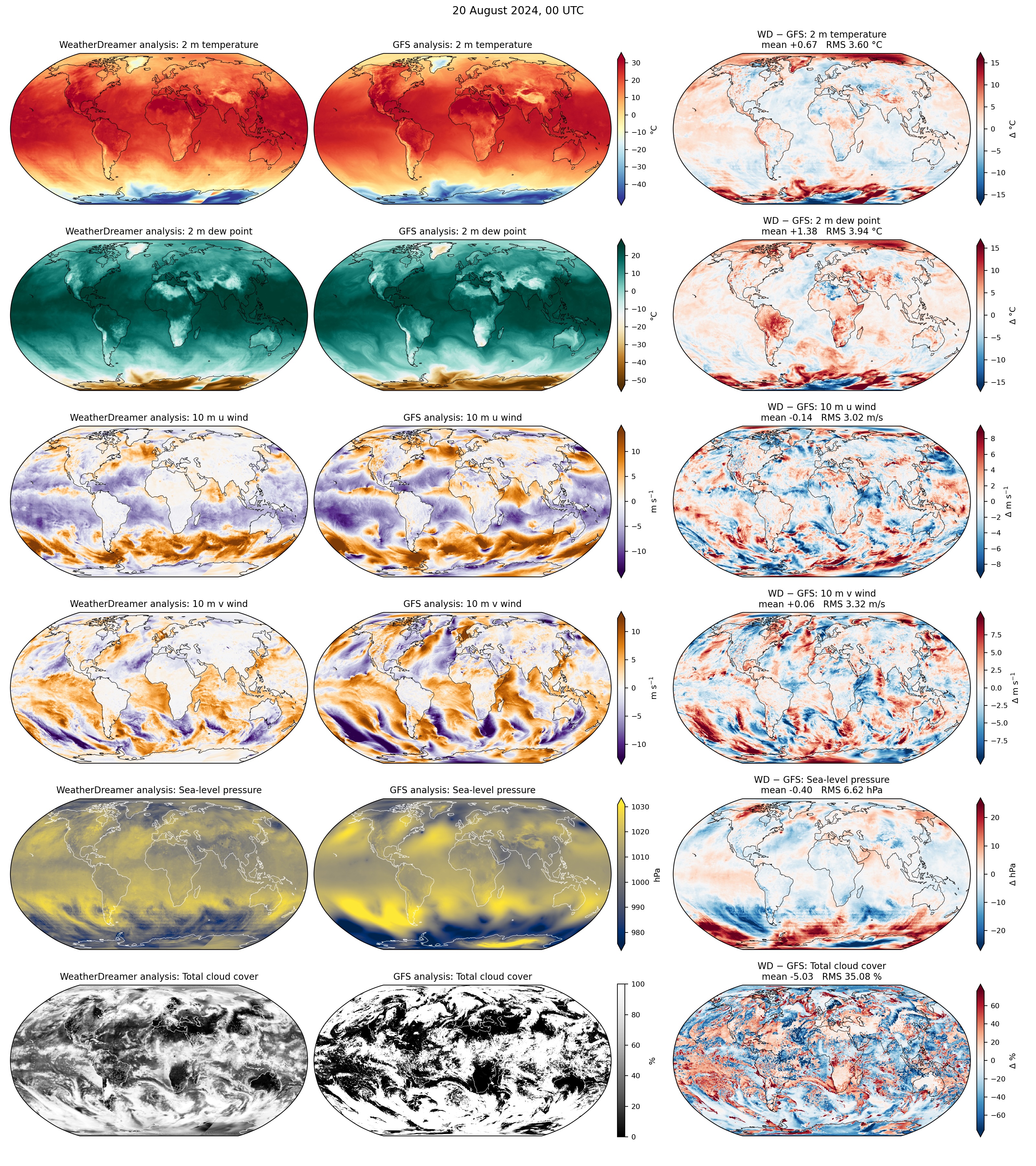}
\caption{
\textbf{WeatherDreamer surface analysis compared with the GFS analysis.} 
WeatherDreamer  0.23$^\circ$analysis, GFS 0.25$^\circ$ analysis, and their difference at 00\,UTC on 20 August 2024, for 2~m temperature, 2~m dew point, 10~m zonal ($u$) and meridional ($v$) wind, sea-level pressure and total cloud cover. 
WeatherDreamer fields are decoded on a 25\,km HEALPix grid from the surface-station decoder, except 10~m winds over ocean, which come from the ASCAT decoder. 
Because the surface-station decoder is trained only at station locations, its gridded fields are least constrained far from stations, notably over the Southern Ocean and Antarctica.
WeatherDreamer cloud cover is the decoded ISD total sky cover, scaled to percent; GFS cloud cover is its total-column cloud fraction, so the two are only approximately comparable.
Difference panels give the area-weighted global mean and RMS.
}
\label{fig:analysisgfs}
\end{figure}

\begin{figure}[htbp]
\noindent\includegraphics[width=\textwidth]{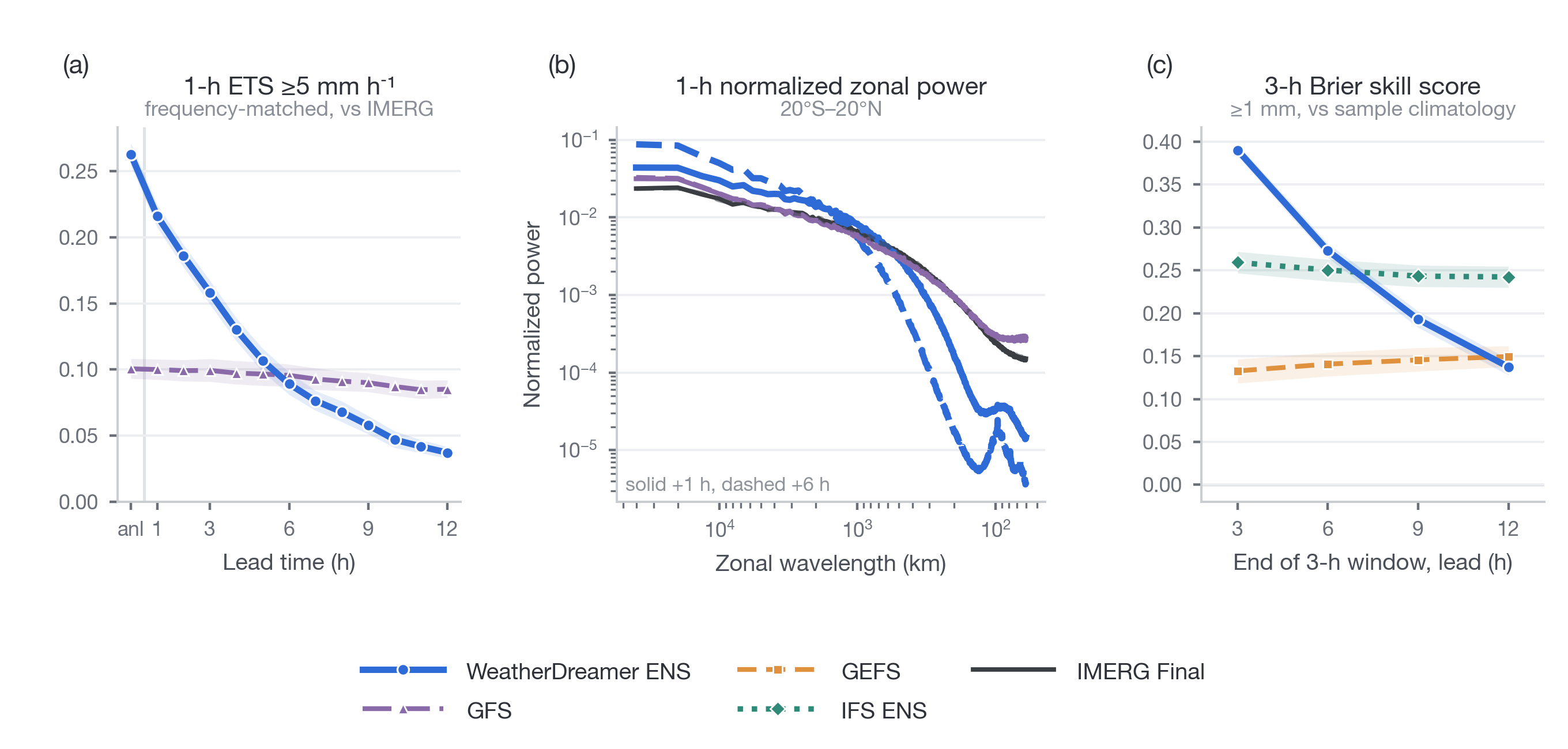}
\caption{
\textbf{Supplementary precipitation diagnostics against IMERG.}
Liquid precipitation between 60$^\circ$S and 60$^\circ$N, verified against
IMERG Final V07B, which is withheld from WeatherDreamer at inference.
(a) Frequency-matched equitable threat score (higher is better) for hourly
precipitation of at least 5\,mm\,h$^{-1}$, for the WeatherDreamer ensemble
mean and GFS, from the analysis hour (anl) to +12\,h.
(b) Normalized zonal power spectra of hourly precipitation between
20$^\circ$S and 20$^\circ$N for WeatherDreamer, GFS and IMERG, at +1\,h
(solid) and +6\,h (dashed).
(c) Brier skill score for 3-h accumulations of at least 1\,mm, relative to
the sample climatology, for the WeatherDreamer ensemble, GEFS and IFS ENS.
Shading in (a) and (c) is the 95\% bootstrap interval from resampling
initializations. Scoring details are given in Text~S6.5.
}
\label{fig:analysisgfs}
\end{figure}

\begin{figure}[ht!]
\centering
\includegraphics[width=\textwidth]{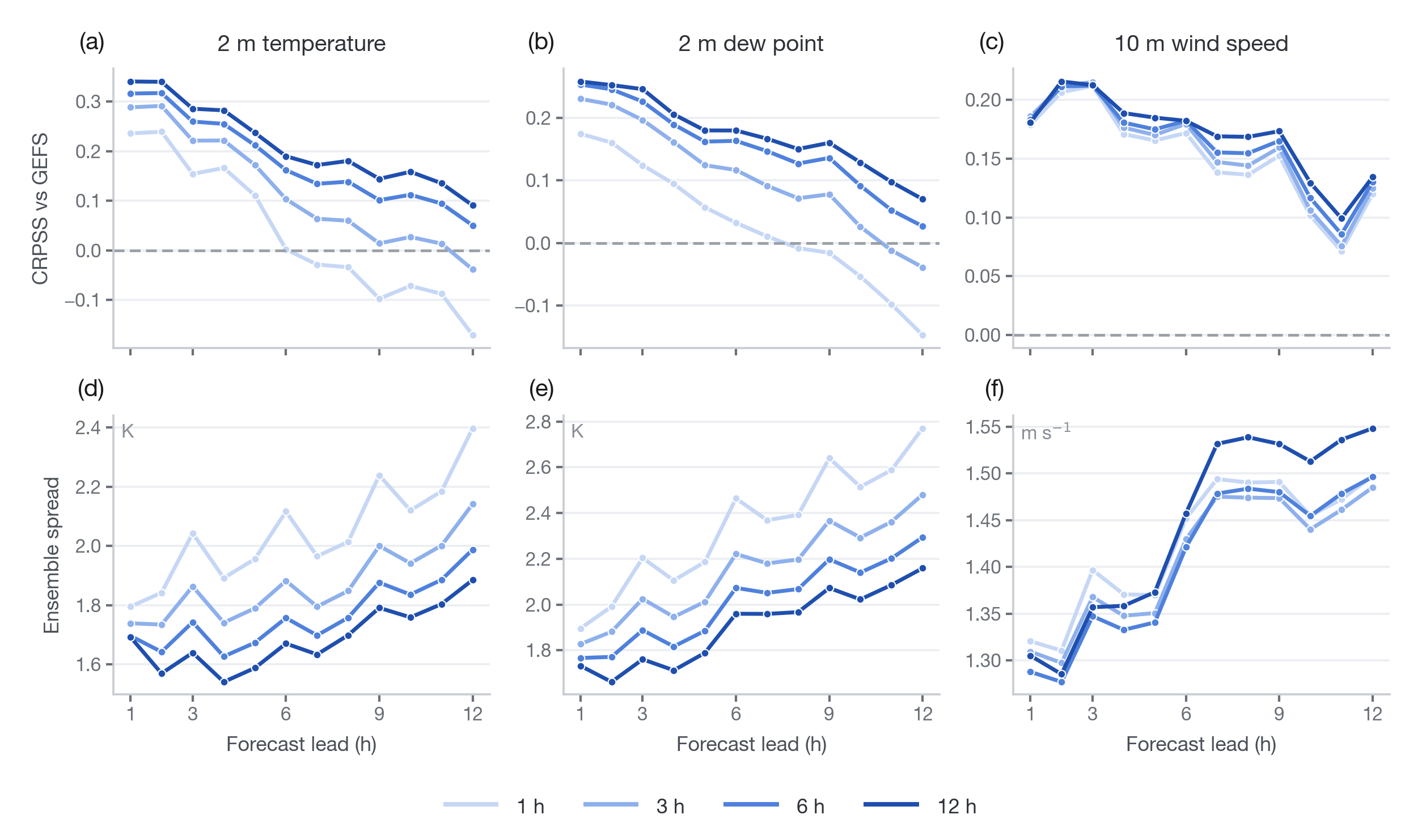}
\caption{
\textbf{Forecast skill as a function of assimilation-window length.}
WeatherDreamer ensemble forecasts initialized after assimilating 1, 3, 6 or 12\,h of observations (light to dark lines), verified at the 534 held-out ISD stations for the eight set-A initializations marked in Table~S9.
(a--c) Continuous ranked probability skill score relative to GEFS, $\mathrm{CRPSS} = 1 - \mathrm{CRPS}/\mathrm{CRPS}_{\mathrm{GEFS}}$, for (a) 2\,m temperature, (b) 2\,m dew point and (c) 10\,m wind speed.
Positive values are more skillful than GEFS, and the dashed line marks parity.
(d--f) Ensemble spread for the same variables. 
For temperature and dew point, spread decreases as the window lengthens, in step with the lower error; for wind speed it is nearly independent of window length.
}
\label{fig:warmup}
\end{figure} 

\begin{figure}[t!]
\centering
\includegraphics[width=\textwidth]{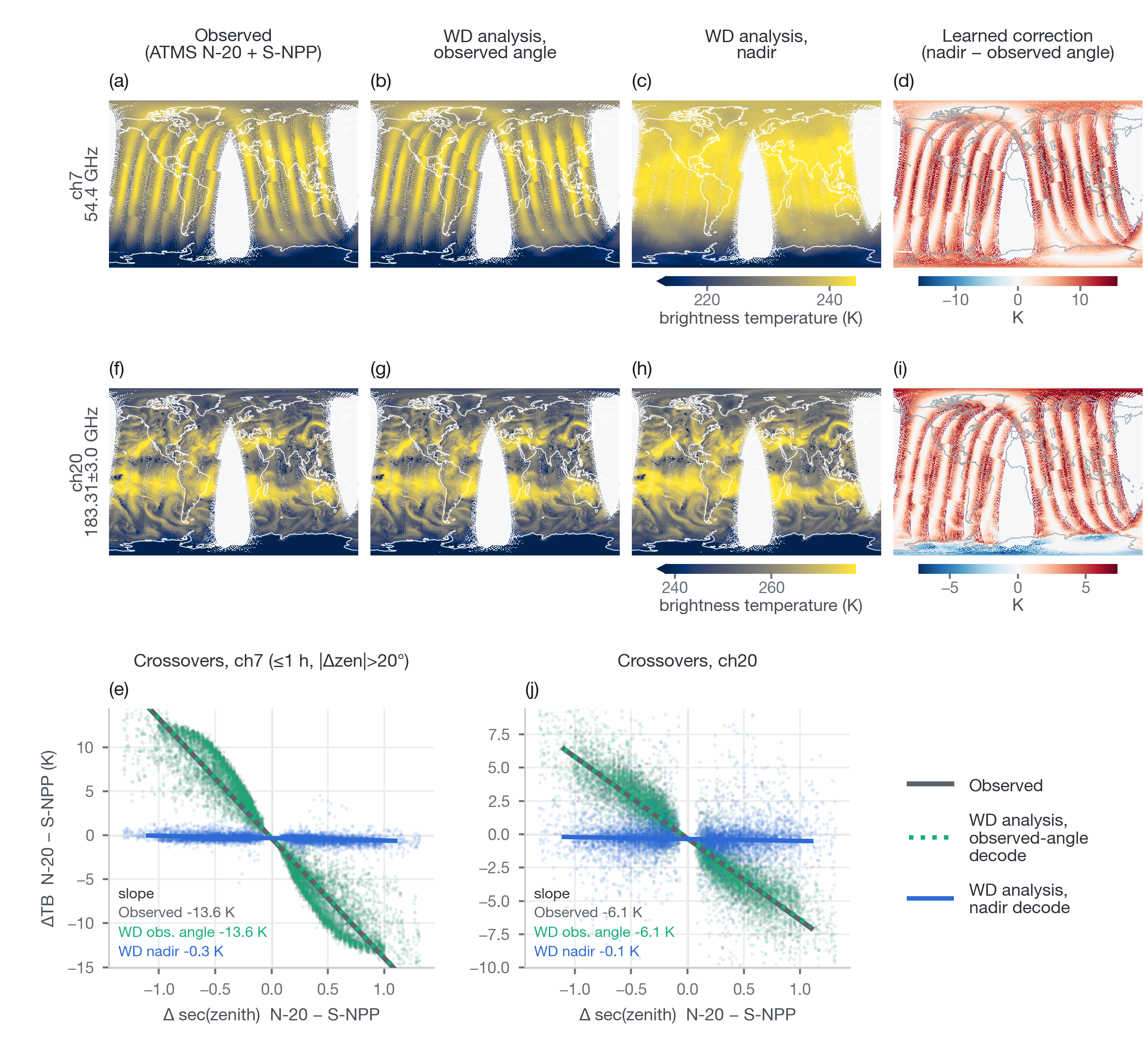}
\caption{
\textbf{Learned viewing-angle correction for microwave sounders.}
(a, f) Twelve-hour composite (2024-08-20, 00–12 UTC) of observed brightness temperatures from ATMS on NOAA-20 and Suomi-NPP, channel 7 (54.4\,GHz, top row) and channel 20 (183.31\,$\pm$\,3.0\,GHz, bottom row).
(b, g) The WeatherDreamer analysis decoded at each pixel's observed viewing angle.
(c, h) The same analysis decoded at nadir.
(d, i) The learned correction, nadir minus observed-angle decode. 
(e, j) Crossovers where NOAA-20 and Suomi-NPP view the same location within 1\,h at viewing angles differing by more than 20$^\circ$: NOAA-20 minus Suomi-NPP brightness temperature against the difference in $\sec\theta$, with fitted slopes (K per unit $\sec\theta$) for the observations, the observed-angle decode and the nadir decode.
}
\label{fig:mwlimb}
\end{figure}

\clearpage

\begin{table}[htbp]
\centering
\scriptsize
\setlength{\tabcolsep}{3pt}
\caption{ATMS on NOAA-20 and Suomi-NPP.}
\label{tab:setb_atms}
\begin{tabular*}{\textwidth}{@{\extracolsep{\fill}}llrrrrrrrrrrrr}
\toprule
 & & \multicolumn{6}{c}{All observations} & \multicolumn{6}{c}{Assimilated by GSI} \\
\cmidrule(lr){3-8}\cmidrule(lr){9-14}
 & & & \multicolumn{2}{c}{GSI} & \multicolumn{2}{c}{WD} & & & \multicolumn{2}{c}{GSI} & \multicolumn{2}{c}{WD} & \\
\cmidrule(lr){4-5}\cmidrule(lr){6-7}\cmidrule(lr){10-11}\cmidrule(lr){12-13}
Ch. & GHz & $N$ & Bias & SD & Bias & SD & Ratio & $N$ & Bias & SD & Bias & SD & Ratio \\
\midrule
\multicolumn{14}{l}{\textit{NOAA-20}} \\
1 & 23.8 & 202,437 & +0.81 & 4.66 & $-$0.10 & 6.27 & 1.35 & 173,133 & +0.43 & 3.03 & $-$0.76 & 5.25 & 1.73 \\
2 & 31.4 & 202,163 & +0.54 & 6.56 & $-$0.30 & 6.57 & 1.00 & 168,939 & +0.07 & 3.17 & $-$1.15 & 4.55 & 1.44 \\
3 & 50.3 & 202,552 & +0.03 & 3.94 & +0.12 & 3.83 & 0.97 & 173,809 & $-$0.13 & 2.62 & $-$0.31 & 2.94 & 1.12 \\
4 & 51.76 & 202,552 & $-$0.74 & 2.11 & +0.15 & 2.18 & 1.03 & 175,810 & $-$0.87 & 1.59 & $-$0.09 & 1.77 & 1.11 \\
5 & 52.8 & 202,552 & $-$0.53 & 0.74 & +0.11 & 0.99 & 1.35 & 176,072 & $-$0.53 & 0.52 & +0.08 & 0.86 & 1.67 \\
6 & 53.596$\pm$0.115 & 202,552 & $-$0.08 & 0.34 & +0.10 & 0.67 & 1.98 & 176,312 & $-$0.04 & 0.18 & +0.12 & 0.61 & 3.37 \\
7 & 54.40 & 202,552 & $-$0.36 & 0.18 & +0.15 & 0.51 & 2.82 & 176,384 & $-$0.34 & 0.16 & +0.16 & 0.49 & 3.13 \\
8 & 54.94 & 202,552 & $-$0.67 & 0.14 & +0.17 & 0.47 & 3.24 & 202,496 & $-$0.67 & 0.14 & +0.17 & 0.47 & 3.26 \\
9 & 55.50 & 202,552 & $-$0.86 & 0.13 & +0.18 & 0.59 & 4.55 & 202,528 & $-$0.86 & 0.13 & +0.18 & 0.59 & 4.57 \\
10 & 57.29 & 202,552 & $-$0.59 & 0.19 & $-$0.04 & 0.71 & 3.64 & 202,494 & $-$0.59 & 0.19 & $-$0.04 & 0.71 & 3.71 \\
11 & 57.29$\pm$0.217 & 202,552 & $-$0.45 & 0.21 & $-$0.13 & 0.67 & 3.19 & 202,479 & $-$0.45 & 0.21 & $-$0.13 & 0.67 & 3.20 \\
12 & 57.29$\pm$0.322$\pm$0.048 & 202,552 & $-$0.40 & 0.25 & $-$0.06 & 0.68 & 2.73 & 202,260 & $-$0.40 & 0.24 & $-$0.06 & 0.68 & 2.77 \\
13 & 57.29$\pm$0.322$\pm$0.022 & 202,552 & $-$0.62 & 0.38 & +0.11 & 0.97 & 2.59 & 199,033 & $-$0.61 & 0.35 & +0.11 & 0.96 & 2.70 \\
14 & 57.29$\pm$0.322$\pm$0.010 & 202,552 & $-$0.59 & 0.57 & +0.14 & 1.16 & 2.04 & 200,959 & $-$0.57 & 0.55 & +0.13 & 1.16 & 2.10 \\
15 & 57.29$\pm$0.322$\pm$0.0045 & 202,552 & $-$0.91 & 1.05 & +0.04 & 1.45 & 1.38 & 200,996 & $-$0.88 & 1.02 & +0.07 & 1.44 & 1.41 \\
16 & 88.2 & 202,537 & $-$0.26 & 6.56 & +0.16 & 6.44 & 0.98 & 150,772 & +0.10 & 3.64 & $-$0.78 & 5.22 & 1.43 \\
17 & 165.5 & 201,996 & $-$0.19 & 5.05 & $-$0.02 & 5.05 & 1.00 & 160,654 & +0.05 & 2.20 & +0.24 & 3.59 & 1.63 \\
18 & 183.31$\pm$7.0 & 202,326 & $-$0.53 & 3.88 & $-$0.06 & 3.73 & 0.96 & 161,528 & $-$0.16 & 1.52 & +0.37 & 2.37 & 1.56 \\
19 & 183.31$\pm$4.5 & 202,445 & $-$0.30 & 3.23 & $-$0.03 & 3.34 & 1.04 & 162,203 & +0.09 & 1.57 & +0.32 & 2.35 & 1.50 \\
20 & 183.31$\pm$3.0 & 202,464 & $-$0.22 & 2.73 & +0.00 & 3.01 & 1.10 & 162,101 & +0.10 & 1.65 & +0.25 & 2.40 & 1.46 \\
21 & 183.31$\pm$1.8 & 202,476 & $-$0.03 & 2.36 & $-$0.02 & 2.80 & 1.19 & 161,385 & +0.19 & 1.75 & +0.16 & 2.49 & 1.42 \\
22 & 183.31$\pm$1.0 & 202,469 & +0.56 & 2.23 & $-$0.05 & 2.74 & 1.23 & 160,513 & +0.69 & 1.82 & +0.06 & 2.55 & 1.40 \\
\midrule
\multicolumn{14}{l}{\textit{Suomi-NPP}} \\
1 & 23.8 & 202,931 & +1.15 & 4.66 & $-$0.16 & 6.33 & 1.36 & 173,853 & +0.74 & 3.03 & $-$0.84 & 5.33 & 1.76 \\
2 & 31.4 & 202,683 & +0.99 & 6.54 & $-$0.11 & 6.71 & 1.03 & 169,692 & +0.47 & 3.21 & $-$1.17 & 4.54 & 1.41 \\
3 & 50.3 & 203,065 & +0.29 & 3.94 & +0.17 & 3.82 & 0.97 & 174,519 & +0.12 & 2.65 & $-$0.32 & 2.90 & 1.10 \\
4 & 51.76 & 203,065 & $-$0.42 & 2.08 & +0.10 & 2.17 & 1.05 & 176,523 & $-$0.52 & 1.62 & $-$0.10 & 1.73 & 1.07 \\
5 & 52.8 & 203,065 & $-$0.47 & 0.74 & +0.11 & 1.00 & 1.36 & 176,768 & $-$0.45 & 0.53 & +0.08 & 0.83 & 1.56 \\
6 & 53.596$\pm$0.115 & 203,065 & $-$0.02 & 0.35 & +0.11 & 0.67 & 1.93 & 176,954 & +0.02 & 0.19 & +0.13 & 0.61 & 3.12 \\
7 & 54.40 & 203,065 & $-$0.09 & 0.23 & +0.13 & 0.51 & 2.25 & 177,043 & $-$0.07 & 0.21 & +0.14 & 0.50 & 2.37 \\
8 & 54.94 & 203,065 & $-$0.67 & 0.15 & +0.15 & 0.47 & 3.09 & 203,047 & $-$0.67 & 0.15 & +0.15 & 0.47 & 3.09 \\
9 & 55.50 & 203,065 & $-$0.61 & 0.17 & +0.16 & 0.59 & 3.54 & 203,021 & $-$0.61 & 0.17 & +0.16 & 0.59 & 3.54 \\
10 & 57.29 & 203,065 & $-$0.35 & 0.21 & $-$0.04 & 0.73 & 3.54 & 203,036 & $-$0.35 & 0.21 & $-$0.04 & 0.73 & 3.54 \\
11 & 57.29$\pm$0.217 & 203,065 & $-$0.31 & 0.22 & $-$0.18 & 0.67 & 3.07 & 203,025 & $-$0.31 & 0.22 & $-$0.18 & 0.67 & 3.07 \\
12 & 57.29$\pm$0.322$\pm$0.048 & 203,065 & $-$0.35 & 0.26 & $-$0.06 & 0.70 & 2.67 & 202,749 & $-$0.35 & 0.26 & $-$0.06 & 0.70 & 2.69 \\
13 & 57.29$\pm$0.322$\pm$0.022 & 203,065 & $-$0.46 & 0.39 & +0.12 & 0.96 & 2.47 & 198,770 & $-$0.45 & 0.37 & +0.11 & 0.95 & 2.57 \\
14 & 57.29$\pm$0.322$\pm$0.010 & 203,064 & $-$0.39 & 0.59 & +0.15 & 1.22 & 2.07 & 201,444 & $-$0.38 & 0.57 & +0.14 & 1.21 & 2.15 \\
15 & 57.29$\pm$0.322$\pm$0.0045 & 203,065 & $-$0.81 & 1.07 & +0.04 & 1.49 & 1.39 & 201,702 & $-$0.78 & 1.07 & +0.06 & 1.47 & 1.38 \\
16 & 88.2 & 203,058 & $-$0.40 & 6.53 & +0.26 & 6.39 & 0.98 & 151,437 & $-$0.06 & 3.63 & $-$0.72 & 5.15 & 1.42 \\
17 & 165.5 & 202,554 & +0.00 & 5.14 & $-$0.03 & 5.21 & 1.01 & 161,133 & +0.26 & 2.19 & +0.30 & 3.51 & 1.60 \\
18 & 183.31$\pm$7.0 & 202,865 & $-$0.31 & 3.89 & $-$0.11 & 3.74 & 0.96 & 162,106 & +0.08 & 1.54 & +0.33 & 2.37 & 1.54 \\
19 & 183.31$\pm$4.5 & 202,981 & $-$0.03 & 3.29 & $-$0.05 & 3.29 & 1.00 & 162,930 & +0.36 & 1.59 & +0.26 & 2.32 & 1.46 \\
20 & 183.31$\pm$3.0 & 202,996 & +0.07 & 2.74 & $-$0.04 & 2.96 & 1.08 & 162,833 & +0.40 & 1.68 & +0.21 & 2.39 & 1.42 \\
21 & 183.31$\pm$1.8 & 203,001 & +0.25 & 2.38 & $-$0.08 & 2.76 & 1.16 & 162,104 & +0.45 & 1.77 & +0.12 & 2.49 & 1.41 \\
22 & 183.31$\pm$1.0 & 202,999 & +0.81 & 2.26 & $-$0.07 & 2.74 & 1.21 & 161,251 & +0.89 & 1.85 & +0.02 & 2.60 & 1.41 \\
\bottomrule
\end{tabular*}
\end{table}

\begin{table}[htbp]
\centering
\scriptsize
\setlength{\tabcolsep}{3pt}
\caption{AMSU-A on MetOp-B and MetOp-C. Only observations present in WeatherDreamer's input data in the same hour are used (45--83\% of GSI's MetOp-B and 58--78\% of its MetOp-C observations per cycle; see text). Channels 3, 6 and 15 are not used by WeatherDreamer.}
\label{tab:setb_amsua}
\begin{tabular*}{\textwidth}{@{\extracolsep{\fill}}llrrrrrrrrrrrr}
\toprule
 & & \multicolumn{6}{c}{All observations} & \multicolumn{6}{c}{Assimilated by GSI} \\
\cmidrule(lr){3-8}\cmidrule(lr){9-14}
 & & & \multicolumn{2}{c}{GSI} & \multicolumn{2}{c}{WD} & & & \multicolumn{2}{c}{GSI} & \multicolumn{2}{c}{WD} & \\
\cmidrule(lr){4-5}\cmidrule(lr){6-7}\cmidrule(lr){10-11}\cmidrule(lr){12-13}
Ch. & GHz & $N$ & Bias & SD & Bias & SD & Ratio & $N$ & Bias & SD & Bias & SD & Ratio \\
\midrule
\multicolumn{14}{l}{\textit{MetOp-B}} \\
1 & 23.8 & 99,508 & +2.44 & 5.04 & +0.52 & 5.95 & 1.18 & -- & -- & -- & -- & -- & -- \\
2 & 31.4 & 99,273 & +1.23 & 6.88 & +0.10 & 6.26 & 0.91 & -- & -- & -- & -- & -- & -- \\
4 & 52.8 & 99,524 & +0.02 & 0.76 & +0.28 & 0.97 & 1.28 & -- & -- & -- & -- & -- & -- \\
5 & 53.596$\pm$0.115 & 99,524 & +0.31 & 0.44 & +0.34 & 0.68 & 1.56 & -- & -- & -- & -- & -- & -- \\
7 & 54.94 & 98,751 & $-$0.64 & 0.21 & +0.42 & 0.51 & 2.47 & -- & -- & -- & -- & -- & -- \\
8 & 55.50 & 99,524 & $-$0.10 & 0.23 & +0.39 & 0.59 & 2.52 & 99,322 & $-$0.10 & 0.23 & +0.39 & 0.59 & 2.53 \\
9 & 57.29 & 99,524 & $-$0.20 & 0.25 & +0.20 & 0.66 & 2.67 & 99,308 & $-$0.20 & 0.25 & +0.20 & 0.66 & 2.68 \\
10 & 57.29$\pm$0.217 & 99,524 & +0.03 & 0.29 & $-$0.03 & 0.62 & 2.11 & 99,479 & +0.03 & 0.29 & $-$0.03 & 0.62 & 2.11 \\
11 & 57.29$\pm$0.322$\pm$0.048 & 99,524 & +0.04 & 0.33 & +0.15 & 0.65 & 1.98 & 99,451 & +0.04 & 0.33 & +0.15 & 0.65 & 1.98 \\
12 & 57.29$\pm$0.322$\pm$0.022 & 99,040 & +0.02 & 0.46 & +0.48 & 0.79 & 1.71 & 98,733 & +0.02 & 0.46 & +0.48 & 0.79 & 1.72 \\
13 & 57.29$\pm$0.322$\pm$0.010 & 99,260 & $-$0.05 & 0.65 & +0.34 & 1.09 & 1.68 & 98,778 & $-$0.05 & 0.65 & +0.34 & 1.08 & 1.67 \\
14 & 57.29$\pm$0.322$\pm$0.0045 & 99,226 & $-$0.30 & 1.17 & +0.17 & 1.50 & 1.28 & 98,633 & $-$0.27 & 1.16 & +0.17 & 1.49 & 1.28 \\
\midrule
\multicolumn{14}{l}{\textit{MetOp-C}} \\
1 & 23.8 & 111,507 & +1.58 & 4.92 & $-$0.50 & 6.10 & 1.24 & 96,881 & +1.22 & 3.92 & $-$0.73 & 5.47 & 1.39 \\
2 & 31.4 & 111,296 & +1.31 & 6.59 & +0.01 & 6.48 & 0.98 & 97,719 & +0.85 & 5.02 & $-$0.33 & 5.16 & 1.03 \\
4 & 52.8 & 111,527 & $-$0.49 & 0.92 & $-$0.20 & 1.10 & 1.19 & 99,542 & $-$0.45 & 0.84 & $-$0.19 & 1.03 & 1.24 \\
5 & 53.596$\pm$0.115 & 111,530 & $-$0.24 & 0.46 & $-$0.21 & 0.68 & 1.47 & 101,759 & $-$0.19 & 0.34 & $-$0.17 & 0.62 & 1.84 \\
7 & 54.94 & 111,530 & $-$1.31 & 0.22 & $-$0.26 & 0.48 & 2.21 & 111,251 & $-$1.31 & 0.21 & $-$0.26 & 0.48 & 2.22 \\
8 & 55.50 & 108,656 & $-$0.58 & 0.38 & $-$0.11 & 0.64 & 1.67 & 103,652 & $-$0.58 & 0.36 & $-$0.10 & 0.63 & 1.75 \\
9 & 57.29 & 111,530 & $-$0.75 & 0.24 & $-$0.35 & 0.64 & 2.70 & 111,372 & $-$0.75 & 0.24 & $-$0.35 & 0.64 & 2.71 \\
10 & 57.29$\pm$0.217 & 111,338 & $-$0.35 & 0.26 & $-$0.45 & 0.61 & 2.36 & 111,314 & $-$0.35 & 0.26 & $-$0.45 & 0.61 & 2.36 \\
11 & 57.29$\pm$0.322$\pm$0.048 & 111,530 & $-$0.42 & 0.32 & $-$0.24 & 0.63 & 1.95 & 111,417 & $-$0.42 & 0.32 & $-$0.24 & 0.63 & 1.95 \\
12 & 57.29$\pm$0.322$\pm$0.022 & 111,518 & $-$0.68 & 0.42 & $-$0.27 & 0.83 & 1.95 & 111,244 & $-$0.68 & 0.42 & $-$0.27 & 0.83 & 1.95 \\
13 & 57.29$\pm$0.322$\pm$0.010 & 111,475 & $-$0.63 & 0.65 & $-$0.21 & 1.06 & 1.63 & 111,028 & $-$0.63 & 0.64 & $-$0.21 & 1.05 & 1.64 \\
14 & 57.29$\pm$0.322$\pm$0.0045 & 111,511 & $-$0.65 & 1.12 & $-$0.32 & 1.42 & 1.27 & 110,528 & $-$0.62 & 1.09 & $-$0.29 & 1.41 & 1.30 \\
\bottomrule
\end{tabular*}
\end{table}

\begin{table}[htbp]
\centering
\scriptsize
\setlength{\tabcolsep}{3pt}
\caption{HIRS/4 on NOAA-19. GSI monitors but does not assimilate these radiances, so there is no assimilated subset; its departures include cloud-affected scenes, which its clear-sky radiative transfer cannot represent. Input-data coverage 90--100\% per cycle.}
\label{tab:setb_hirs}
\begin{tabular*}{\textwidth}{@{\extracolsep{\fill}}llrrrrrrrrrrrr}
\toprule
 & & \multicolumn{6}{c}{All observations} & \multicolumn{6}{c}{Assimilated by GSI} \\
\cmidrule(lr){3-8}\cmidrule(lr){9-14}
 & & & \multicolumn{2}{c}{GSI} & \multicolumn{2}{c}{WD} & & & \multicolumn{2}{c}{GSI} & \multicolumn{2}{c}{WD} & \\
\cmidrule(lr){4-5}\cmidrule(lr){6-7}\cmidrule(lr){10-11}\cmidrule(lr){12-13}
Ch. & \textmu m & $N$ & Bias & SD & Bias & SD & Ratio & $N$ & Bias & SD & Bias & SD & Ratio \\
\midrule
\multicolumn{14}{l}{\textit{NOAA-19}} \\
2 & 14.71 & 143,655 & $-$0.13 & 1.16 & +0.07 & 1.27 & 1.10 & -- & -- & -- & -- & -- & -- \\
3 & 14.49 & 143,655 & $-$0.11 & 0.61 & +0.11 & 0.74 & 1.21 & -- & -- & -- & -- & -- & -- \\
4 & 14.22 & 143,655 & $-$1.54 & 1.41 & +0.03 & 1.26 & 0.89 & -- & -- & -- & -- & -- & -- \\
5 & 13.97 & 143,654 & $-$2.49 & 3.14 & +2.13 & 2.58 & 0.82 & -- & -- & -- & -- & -- & -- \\
6 & 13.64 & 143,562 & $-$2.50 & 4.84 & +2.04 & 4.03 & 0.83 & -- & -- & -- & -- & -- & -- \\
7 & 13.35 & 143,345 & $-$4.24 & 6.14 & +2.62 & 5.24 & 0.85 & -- & -- & -- & -- & -- & -- \\
8 & 11.11 & 142,087 & $-$5.00 & 8.58 & +5.36 & 7.64 & 0.89 & -- & -- & -- & -- & -- & -- \\
9 & 9.71 & 143,436 & $-$3.39 & 6.13 & +3.04 & 5.08 & 0.83 & -- & -- & -- & -- & -- & -- \\
10 & 12.47 & 142,548 & $-$4.75 & 8.05 & +5.00 & 7.16 & 0.89 & -- & -- & -- & -- & -- & -- \\
11 & 7.33 & 143,582 & $-$1.59 & 4.33 & +1.96 & 4.21 & 0.97 & -- & -- & -- & -- & -- & -- \\
12 & 6.52 & 143,654 & +0.00 & 2.66 & +0.82 & 2.82 & 1.06 & -- & -- & -- & -- & -- & -- \\
13 & 4.57 & 143,073 & $-$2.11 & 6.97 & +5.12 & 6.03 & 0.87 & -- & -- & -- & -- & -- & -- \\
14 & 4.52 & 143,468 & $-$2.87 & 5.11 & +1.45 & 4.51 & 0.88 & -- & -- & -- & -- & -- & -- \\
15 & 4.47 & 143,603 & $-$0.72 & 4.48 & +4.01 & 3.75 & 0.84 & -- & -- & -- & -- & -- & -- \\
16 & 4.45 & 143,654 & $-$0.73 & 2.60 & +0.68 & 2.34 & 0.90 & -- & -- & -- & -- & -- & -- \\
17 & 4.13 & 142,922 & $-$4.18 & 6.72 & +3.85 & 5.91 & 0.88 & -- & -- & -- & -- & -- & -- \\
18 & 4.00 & 142,700 & $-$3.39 & 7.49 & +3.92 & 6.71 & 0.90 & -- & -- & -- & -- & -- & -- \\
19 & 3.76 & 143,151 & $-$1.84 & 7.70 & +3.07 & 6.76 & 0.88 & -- & -- & -- & -- & -- & -- \\
\bottomrule
\end{tabular*}
\end{table}

\begin{table}[htbp]
\centering
\scriptsize
\setlength{\tabcolsep}{3pt}
\caption{GOES-16 ABI at GSI's clear-sky superobservation locations only; WeatherDreamer's all-sky forecast is penalized wherever it places cloud. GSI assimilates only bands 8 (6.2\,\textmu m) and 10 (7.3\,\textmu m); the other bands are monitored.}
\label{tab:setb_abi}
\begin{tabular*}{\textwidth}{@{\extracolsep{\fill}}llrrrrrrrrrrrr}
\toprule
 & & \multicolumn{6}{c}{All observations} & \multicolumn{6}{c}{Assimilated by GSI} \\
\cmidrule(lr){3-8}\cmidrule(lr){9-14}
 & & & \multicolumn{2}{c}{GSI} & \multicolumn{2}{c}{WD} & & & \multicolumn{2}{c}{GSI} & \multicolumn{2}{c}{WD} & \\
\cmidrule(lr){4-5}\cmidrule(lr){6-7}\cmidrule(lr){10-11}\cmidrule(lr){12-13}
Ch. & \textmu m & $N$ & Bias & SD & Bias & SD & Ratio & $N$ & Bias & SD & Bias & SD & Ratio \\
\midrule
\multicolumn{14}{l}{\textit{GOES-16}} \\
7 & 3.9 & 80,983 & +0.62 & 2.71 & +3.85 & 4.37 & 1.61 & -- & -- & -- & -- & -- & -- \\
8 & 6.2 & 80,986 & +0.53 & 1.59 & +0.42 & 1.96 & 1.23 & 78,520 & +0.39 & 1.28 & +0.37 & 1.88 & 1.46 \\
9 & 6.9 & 80,986 & +1.22 & 1.53 & +0.54 & 2.24 & 1.46 & -- & -- & -- & -- & -- & -- \\
10 & 7.3 & 80,986 & +0.14 & 1.19 & +0.80 & 2.27 & 1.91 & 32,652 & +0.01 & 0.94 & +0.45 & 1.63 & 1.73 \\
11 & 8.4 & 80,982 & $-$0.41 & 0.59 & +3.20 & 4.59 & 7.72 & -- & -- & -- & -- & -- & -- \\
12 & 9.6 & 80,982 & $-$0.53 & 0.49 & +2.46 & 2.69 & 5.51 & -- & -- & -- & -- & -- & -- \\
13 & 10.3 & 80,986 & $-$0.26 & 0.66 & +3.48 & 4.98 & 7.53 & -- & -- & -- & -- & -- & -- \\
14 & 11.2 & 80,985 & $-$0.04 & 0.74 & +3.49 & 5.22 & 7.08 & -- & -- & -- & -- & -- & -- \\
15 & 12.3 & 80,986 & $-$0.10 & 0.77 & +3.09 & 5.08 & 6.64 & -- & -- & -- & -- & -- & -- \\
16 & 13.3 & 80,986 & $-$0.55 & 0.57 & +1.98 & 3.53 & 6.23 & -- & -- & -- & -- & -- & -- \\
\bottomrule
\end{tabular*}
\end{table}

\begin{table}[htbp]
\centering
\scriptsize
\setlength{\tabcolsep}{3pt}
\caption{Radiosondes by pressure level, interpolated in log-pressure from the 11 decoded levels; $N$ is pooled over the 30 cycles.}
\label{tab:setb_raob}
\begin{tabular*}{\textwidth}{@{\extracolsep{\fill}}llrrrrrrrrrrrr}
\toprule
 & & \multicolumn{6}{c}{All observations} & \multicolumn{6}{c}{Assimilated by GSI} \\
\cmidrule(lr){3-8}\cmidrule(lr){9-14}
 & & & \multicolumn{2}{c}{GSI} & \multicolumn{2}{c}{WD} & & & \multicolumn{2}{c}{GSI} & \multicolumn{2}{c}{WD} & \\
\cmidrule(lr){4-5}\cmidrule(lr){6-7}\cmidrule(lr){10-11}\cmidrule(lr){12-13}
Variable & hPa & $N$ & Bias & SD & Bias & SD & Ratio & $N$ & Bias & SD & Bias & SD & Ratio \\
\midrule
\multicolumn{14}{l}{\textit{Virtual temperature (K)}} \\
 & 1000 & 17,666 & +1.79 & 2.36 & +0.27 & 2.82 & 1.19 & 17,528 & +1.79 & 2.34 & +0.27 & 2.82 & 1.20 \\
 & 925 & 34,193 & +1.50 & 1.87 & +0.41 & 2.78 & 1.49 & 33,879 & +1.50 & 1.82 & +0.43 & 2.67 & 1.47 \\
 & 850 & 55,634 & +1.13 & 1.36 & +0.35 & 2.45 & 1.80 & 55,147 & +1.13 & 1.31 & +0.36 & 2.44 & 1.86 \\
 & 700 & 95,560 & +0.63 & 1.14 & +0.54 & 2.03 & 1.77 & 94,713 & +0.64 & 1.09 & +0.54 & 2.01 & 1.84 \\
 & 500 & 93,622 & +0.32 & 0.97 & +0.65 & 2.20 & 2.25 & 92,895 & +0.36 & 0.91 & +0.66 & 2.10 & 2.32 \\
 & 400 & 77,321 & +0.13 & 0.87 & +0.57 & 2.17 & 2.50 & 76,747 & +0.14 & 0.84 & +0.57 & 2.16 & 2.57 \\
 & 300 & 65,314 & $-$0.05 & 0.95 & +0.19 & 2.26 & 2.38 & 64,822 & $-$0.04 & 0.84 & +0.19 & 2.23 & 2.67 \\
 & 250 & 54,307 & $-$0.08 & 1.04 & $-$0.13 & 2.46 & 2.35 & 53,905 & $-$0.08 & 0.96 & $-$0.15 & 2.39 & 2.49 \\
 & 200 & 62,858 & $-$0.14 & 1.26 & +0.04 & 2.91 & 2.31 & 62,425 & $-$0.15 & 1.13 & +0.00 & 2.84 & 2.51 \\
 & 150 & 74,471 & $-$0.19 & 1.34 & +0.47 & 2.54 & 1.90 & 74,022 & $-$0.20 & 1.17 & +0.47 & 2.52 & 2.16 \\
 & 100 & 46,149 & $-$0.20 & 1.29 & +0.35 & 2.39 & 1.84 & 45,837 & $-$0.20 & 1.24 & +0.34 & 2.28 & 1.84 \\
\multicolumn{14}{l}{\textit{Specific humidity (g\,kg$^{-1}$)}} \\
 & 1000 & 17,640 & +0.26 & 1.55 & +0.27 & 1.85 & 1.19 & 17,517 & +0.27 & 1.55 & +0.28 & 1.84 & 1.19 \\
 & 925 & 34,162 & +0.12 & 1.29 & +0.51 & 1.75 & 1.35 & 33,892 & +0.13 & 1.29 & +0.49 & 1.75 & 1.36 \\
 & 850 & 55,586 & $-$0.05 & 1.35 & +0.59 & 1.88 & 1.39 & 55,161 & $-$0.05 & 1.34 & +0.60 & 1.90 & 1.42 \\
 & 700 & 95,476 & $-$0.07 & 1.13 & +0.29 & 1.56 & 1.37 & 94,694 & $-$0.06 & 1.11 & +0.29 & 1.55 & 1.41 \\
 & 500 & 93,541 & $-$0.03 & 0.71 & +0.13 & 0.88 & 1.23 & 92,400 & $-$0.03 & 0.70 & +0.13 & 0.88 & 1.25 \\
 & 400 & 77,189 & $-$0.01 & 0.37 & +0.06 & 0.41 & 1.09 & 73,414 & $-$0.01 & 0.38 & +0.07 & 0.41 & 1.09 \\
 & 300 & 65,081 & $-$0.01 & 0.14 & +0.03 & 0.18 & 1.25 & 36,010 & +0.00 & 0.21 & +0.06 & 0.22 & 1.06 \\
\multicolumn{14}{l}{\textit{$u$ wind (m\,s$^{-1}$)}} \\
 & 1000 & 17,934 & +0.03 & 2.60 & +0.09 & 3.31 & 1.27 & 17,724 & +0.02 & 2.54 & +0.09 & 3.22 & 1.27 \\
 & 925 & 36,808 & +0.10 & 2.68 & $-$0.01 & 4.17 & 1.56 & 36,478 & +0.09 & 2.57 & $-$0.03 & 4.07 & 1.58 \\
 & 850 & 59,203 & +0.06 & 2.62 & $-$0.29 & 4.52 & 1.73 & 58,744 & +0.06 & 2.58 & $-$0.30 & 4.47 & 1.73 \\
 & 700 & 101,199 & +0.08 & 2.74 & $-$0.48 & 5.46 & 1.99 & 100,535 & +0.06 & 2.65 & $-$0.48 & 5.31 & 2.00 \\
 & 500 & 94,894 & +0.02 & 2.87 & $-$0.38 & 6.74 & 2.34 & 94,366 & +0.03 & 2.78 & $-$0.38 & 6.73 & 2.42 \\
 & 400 & 76,610 & +0.08 & 3.13 & $-$0.18 & 8.74 & 2.79 & 76,140 & +0.07 & 3.05 & $-$0.17 & 8.71 & 2.85 \\
 & 300 & 64,564 & +0.03 & 3.26 & +0.16 & 10.62 & 3.26 & 64,115 & +0.05 & 3.25 & +0.16 & 10.63 & 3.27 \\
 & 250 & 54,985 & +0.05 & 3.31 & +0.33 & 11.62 & 3.51 & 54,576 & +0.06 & 3.23 & +0.33 & 11.60 & 3.59 \\
 & 200 & 64,678 & +0.07 & 3.44 & +0.75 & 11.23 & 3.26 & 64,165 & +0.10 & 3.33 & +0.70 & 11.23 & 3.38 \\
 & 150 & 83,051 & +0.22 & 3.35 & +1.77 & 9.44 & 2.81 & 82,426 & +0.23 & 3.29 & +1.78 & 9.44 & 2.87 \\
 & 100 & 52,966 & +0.31 & 3.16 & +1.61 & 7.65 & 2.42 & 52,527 & +0.39 & 3.14 & +1.58 & 7.60 & 2.42 \\
\multicolumn{14}{l}{\textit{$v$ wind (m\,s$^{-1}$)}} \\
 & 1000 & 17,934 & $-$0.21 & 2.57 & +0.36 & 3.66 & 1.42 & 17,724 & $-$0.16 & 2.52 & +0.35 & 3.41 & 1.35 \\
 & 925 & 36,815 & +0.01 & 2.59 & +0.46 & 4.17 & 1.61 & 36,478 & +0.01 & 2.50 & +0.49 & 4.15 & 1.66 \\
 & 850 & 59,205 & +0.03 & 2.55 & +0.30 & 4.39 & 1.72 & 58,744 & +0.01 & 2.49 & +0.28 & 4.32 & 1.73 \\
 & 700 & 101,206 & $-$0.01 & 2.75 & $-$0.15 & 5.30 & 1.92 & 100,535 & +0.00 & 2.60 & $-$0.14 & 5.22 & 2.01 \\
 & 500 & 94,899 & +0.05 & 2.66 & $-$0.52 & 6.53 & 2.46 & 94,366 & +0.05 & 2.64 & $-$0.52 & 6.48 & 2.46 \\
 & 400 & 76,610 & +0.03 & 2.97 & $-$0.83 & 8.17 & 2.75 & 76,140 & +0.03 & 2.92 & $-$0.83 & 8.16 & 2.79 \\
 & 300 & 64,564 & $-$0.03 & 3.24 & $-$0.97 & 10.24 & 3.16 & 64,115 & $-$0.06 & 3.19 & $-$0.97 & 10.23 & 3.21 \\
 & 250 & 54,985 & +0.01 & 3.40 & $-$0.74 & 11.17 & 3.28 & 54,576 & +0.01 & 3.30 & $-$0.76 & 11.21 & 3.40 \\
 & 200 & 64,677 & +0.01 & 3.31 & $-$1.18 & 10.63 & 3.21 & 64,165 & +0.01 & 3.30 & $-$1.17 & 10.64 & 3.22 \\
 & 150 & 83,051 & +0.09 & 3.42 & $-$1.12 & 8.42 & 2.46 & 82,426 & +0.08 & 3.35 & $-$1.11 & 8.37 & 2.50 \\
 & 100 & 52,966 & +0.12 & 3.21 & $-$1.08 & 6.87 & 2.14 & 52,527 & +0.13 & 3.17 & $-$1.08 & 6.83 & 2.15 \\
\bottomrule
\end{tabular*}
\end{table}

\begin{table}[htbp]
\centering
\scriptsize
\setlength{\tabcolsep}{3pt}
\caption{GPS-RO fractional bending-angle departures at four impact heights.}
\label{tab:setb_gpsro}
\begin{tabular*}{\textwidth}{@{\extracolsep{\fill}}llrrrrrrrrrrrr}
\toprule
 & & \multicolumn{6}{c}{All observations} & \multicolumn{6}{c}{Assimilated by GSI} \\
\cmidrule(lr){3-8}\cmidrule(lr){9-14}
 & & & \multicolumn{2}{c}{GSI} & \multicolumn{2}{c}{WD} & & & \multicolumn{2}{c}{GSI} & \multicolumn{2}{c}{WD} & \\
\cmidrule(lr){4-5}\cmidrule(lr){6-7}\cmidrule(lr){10-11}\cmidrule(lr){12-13}
Satellite & km & $N$ & Bias & SD & Bias & SD & Ratio & $N$ & Bias & SD & Bias & SD & Ratio \\
\midrule
MetOp-B & 5 & 4,261 & $-$1.00 & 8.92 & $-$1.36 & 10.01 & 1.12 & 4,073 & $-$0.40 & 6.81 & $-$0.82 & 9.17 & 1.35 \\
 & 10 & 4,331 & $-$0.17 & 2.10 & +0.01 & 3.10 & 1.48 & 4,233 & $-$0.21 & 1.74 & $-$0.05 & 3.00 & 1.72 \\
 & 20 & 4,331 & $-$0.06 & 1.87 & +0.62 & 3.12 & 1.67 & 4,289 & $-$0.13 & 1.48 & +0.49 & 2.66 & 1.79 \\
 & 30 & 4,331 & $-$0.17 & 1.77 & +0.90 & 2.74 & 1.55 & 4,295 & $-$0.11 & 1.46 & +0.97 & 2.39 & 1.64 \\
\midrule
MetOp-C & 5 & 4,095 & $-$1.38 & 9.19 & $-$1.53 & 10.05 & 1.09 & 3,909 & $-$0.54 & 6.72 & $-$1.28 & 8.81 & 1.31 \\
 & 10 & 4,174 & $-$0.25 & 2.16 & $-$0.03 & 3.23 & 1.49 & 4,055 & $-$0.23 & 1.70 & $-$0.06 & 2.93 & 1.72 \\
 & 20 & 4,174 & +0.06 & 2.50 & +0.55 & 3.39 & 1.36 & 4,136 & $-$0.11 & 1.49 & +0.43 & 2.52 & 1.68 \\
 & 30 & 4,174 & $-$0.22 & 1.69 & +0.76 & 2.57 & 1.52 & 4,141 & $-$0.16 & 1.50 & +0.82 & 2.40 & 1.60 \\
\bottomrule
\end{tabular*}
\end{table}

\begin{table}[htbp]
\centering
\scriptsize
\setlength{\tabcolsep}{3pt}
\caption{Surface stations (held-out and in-sample) and ASCAT ocean surface winds.}
\label{tab:setb_surface}
\begin{tabular*}{\textwidth}{@{\extracolsep{\fill}}llrrrrrrrrrrrr}
\toprule
 & & \multicolumn{6}{c}{All observations} & \multicolumn{6}{c}{Assimilated by GSI} \\
\cmidrule(lr){3-8}\cmidrule(lr){9-14}
 & & & \multicolumn{2}{c}{GSI} & \multicolumn{2}{c}{WD} & & & \multicolumn{2}{c}{GSI} & \multicolumn{2}{c}{WD} & \\
\cmidrule(lr){4-5}\cmidrule(lr){6-7}\cmidrule(lr){10-11}\cmidrule(lr){12-13}
Variable & Stations & $N$ & Bias & SD & Bias & SD & Ratio & $N$ & Bias & SD & Bias & SD & Ratio \\
\midrule
\multicolumn{14}{l}{\textit{Virtual temperature (K)}} \\
 & held-out & 95,357 & +1.55 & 2.37 & $-$0.23 & 2.47 & 1.04 & 70,042 & +1.50 & 2.25 & $-$0.11 & 2.40 & 1.07 \\
 & in-sample & 1,736,058 & +1.67 & 2.60 & $-$0.21 & 2.65 & 1.02 & 1,298,333 & +1.56 & 2.21 & $-$0.12 & 2.32 & 1.05 \\
\multicolumn{14}{l}{\textit{Specific humidity (g\,kg$^{-1}$)}} \\
 & held-out & 93,162 & $-$0.06 & 1.37 & +0.25 & 1.39 & 1.02 & 67,891 & $-$0.03 & 1.39 & +0.29 & 1.38 & 1.00 \\
 & in-sample & 1,692,988 & $-$0.05 & 1.45 & +0.19 & 1.41 & 0.97 & 1,268,343 & $-$0.01 & 1.39 & +0.21 & 1.37 & 0.99 \\
\multicolumn{14}{l}{\textit{$u$ wind (m\,s$^{-1}$)}} \\
 & held-out & 104,411 & +0.12 & 2.07 & +0.02 & 2.27 & 1.10 & 2,944 & +0.02 & 1.81 & $-$0.21 & 1.93 & 1.07 \\
 & in-sample & 1,931,490 & +0.05 & 2.12 & $-$0.09 & 2.14 & 1.01 & 61,111 & +0.23 & 2.39 & $-$0.10 & 2.39 & 1.00 \\
\multicolumn{14}{l}{\textit{$v$ wind (m\,s$^{-1}$)}} \\
 & held-out & 104,411 & $-$0.05 & 2.10 & +0.04 & 2.39 & 1.14 & 2,944 & $-$0.33 & 1.88 & +0.08 & 2.04 & 1.08 \\
 & in-sample & 1,931,490 & $-$0.07 & 2.09 & +0.02 & 2.28 & 1.09 & 61,111 & $-$0.12 & 2.22 & +0.16 & 2.36 & 1.06 \\
\multicolumn{14}{l}{\textit{Wind speed (m\,s$^{-1}$)}} \\
 & held-out & 104,411 & +0.12 & 2.05 & +0.97 & 2.22 & 1.08 & 2,944 & $-$0.62 & 1.44 & +1.31 & 1.49 & 1.04 \\
 & in-sample & 1,931,490 & +0.11 & 2.14 & +0.87 & 2.12 & 0.99 & 61,111 & $-$0.17 & 2.19 & +0.84 & 2.23 & 1.02 \\
\midrule
\multicolumn{14}{l}{\textit{ASCAT ocean surface wind (m\,s$^{-1}$)}} \\
 & $u$ & 3,318,962 & +0.06 & 1.54 & +0.01 & 2.70 & 1.76 & 437,941 & +0.03 & 1.35 & $-$0.02 & 2.67 & 1.97 \\
 & $v$ & 3,318,962 & $-$0.01 & 1.60 & $-$0.03 & 2.88 & 1.80 & 437,941 & +0.01 & 1.45 & +0.00 & 2.81 & 1.94 \\
 & speed & 3,318,962 & +0.47 & 1.31 & +0.70 & 2.16 & 1.65 & 437,941 & +0.45 & 1.22 & +0.69 & 2.12 & 1.74 \\
\bottomrule
\end{tabular*}
\end{table}

\begin{table}[ht!]
\caption{Verification data. (a) Reference observations, treated as truth.
(b) Baselines against which WeatherDreamer is compared.}
\label{tab:verif}
\centering
\small

\textbf{(a) Reference observations}\\[3pt]
\begin{tabular}{p{5cm} p{8cm}}
\toprule
Dataset & Variables \\
\midrule
% GPS-RO (MetOp-B) & Bending angle \\
% ISD held-out stations & $T_{\mathrm{2m}}$, $T_{d,\mathrm{2m}}$, $u_{10}$, $v_{10}$, $p_{\mathrm{msl}}$ \\
% IMERG Final V07B & 1 h liquid precipitation \\
USCRN, SOLRAD  & GHI, DNI, DHI \\
EarthCARE \texttt{AC\_TC\_2B} & Cloud-top height \\
GPM MergIR & 11 \textmu m brightness temperature \\
% GOES-East/West ABI collocations & Clear-ocean brightness temperature (4 bands) \\
\bottomrule
\end{tabular}

\vspace{12pt}

\textbf{(b) Baselines}\\[3pt]
\begin{tabular}{p{5cm} p{8cm}}
\toprule
Dataset & Variables \\
\midrule
NCEP GSI [GSI 3D-Var] & First-guess departures (O$-$B) \\
GFS 0.25\textdegree & Surface variables, precipitation, DSWRF, simulated IR \\
GEFS 0.25\textdegree{} (31 members) & Surface variables \\
IFS ENS 0.25\textdegree{} (51 members) & Surface variables \\
GOES-16/18 ABI DSR & GHI \\
NSRDB & GHI, DNI, DHI \\
HRRR & GHI, cloud-top height \\
GOES ABI ACHA (East, West) & Cloud-top height \\
\bottomrule
\label{tab:refs}
\end{tabular}
\end{table}

\begin{table}[htbp]
\centering
\footnotesize
\caption{Set-A initialization times (UTC). Each run assimilates the preceding 12\,h of observations with operational latency and forecasts 12\,h ahead. GOES-East is GOES-16 until 7 April 2025 and GOES-19 thereafter. $^*$: used in the assimilation-window experiment.}
\label{tab:setA_inits}
\begin{tabular*}{\textwidth}{@{\extracolsep{\fill}}rlrlrl}
\toprule
\# & Initialization & \# & Initialization & \# & Initialization \\
\midrule
1  & 2024-08-20 00$^*$ & 18 & 2024-12-22 08     & 35 & 2025-04-25 16 \\
2  & 2024-08-25 09     & 19 & 2025-01-02 20     & 36 & 2025-05-04 04 \\
3  & 2024-08-28 12     & 20 & 2025-01-08 15$^*$ & 37 & 2025-05-11 23 \\
4  & 2024-09-04 07     & 21 & 2025-01-14 03     & 38 & 2025-05-19 11 \\
5  & 2024-09-15 19     & 22 & 2025-01-22 22     & 39 & 2025-05-26 06$^*$ \\
6  & 2024-09-20 14     & 23 & 2025-02-02 10     & 40 & 2025-06-02 18 \\
7  & 2024-09-25 02     & 24 & 2025-02-06 05     & 41 & 2025-06-10 13 \\
8  & 2024-10-05 21$^*$ & 25 & 2025-02-12 17     & 42 & 2025-06-15 01 \\
9  & 2024-10-14 09     & 26 & 2025-02-22 12$^*$ & 43 & 2025-06-26 20 \\
10 & 2024-10-21 04     & 27 & 2025-02-26 00     & 44 & 2025-07-02 08 \\
11 & 2024-10-24 16     & 28 & 2025-03-04 21     & 45 & 2025-07-12 03$^*$ \\
12 & 2024-11-01 23     & 29 & 2025-03-10 19     & 46 & 2025-07-20 15 \\
13 & 2024-11-10 11     & 30 & 2025-03-19 07     & 47 & 2025-07-27 10 \\
14 & 2024-11-16 18$^*$ & 31 & 2025-03-25 02     & 48 & 2025-08-06 22 \\
15 & 2024-11-24 06     & 32 & 2025-04-03 14     & 49 & 2025-08-11 17 \\
16 & 2024-12-06 01     & 33 & 2025-04-09 09$^*$ & 50 & 2025-08-15 05 \\
17 & 2024-12-14 13     & 34 & 2025-04-13 21     &    & \\
\bottomrule
\end{tabular*}
\end{table}

\begin{table}[htbp]
\centering
\footnotesize
\caption{Set-B GSI cycle times (UTC), all in 2024. For each cycle, WeatherDreamer is initialized 6\,h earlier, assimilates the preceding 12\,h of observations without latency and forecasts 9\,h ahead. The comparison uses the background and analysis diagnostics of the NOAA 3D-Var coupled reanalysis scout run v2 (GSI) for that cycle.}
\label{tab:gsi_cycles}
\begin{tabular}{rlrlrl}
\toprule
\# & Cycle & \# & Cycle & \# & Cycle \\
\midrule
1  & 2024-01-03 00 & 11 & 2024-04-16 06 & 21 & 2024-07-24 00 \\
2  & 2024-01-10 00 & 12 & 2024-04-25 18 & 22 & 2024-08-02 00 \\
3  & 2024-01-25 06 & 13 & 2024-05-11 00 & 23 & 2024-08-08 12 \\
4  & 2024-02-17 12 & 14 & 2024-05-20 12 & 24 & 2024-08-20 06 \\
5  & 2024-02-21 12 & 15 & 2024-05-28 18 & 25 & 2024-09-01 12 \\
6  & 2024-02-25 18 & 16 & 2024-06-10 12 & 26 & 2024-09-08 18 \\
7  & 2024-03-10 00 & 17 & 2024-06-15 00 & 27 & 2024-09-11 18 \\
8  & 2024-03-15 00 & 18 & 2024-06-26 18 & 28 & 2024-10-01 18 \\
9  & 2024-03-25 06 & 19 & 2024-07-06 06 & 29 & 2024-10-18 06 \\
10 & 2024-04-08 06 & 20 & 2024-07-19 12 & 30 & 2024-10-29 12 \\
\bottomrule
\end{tabular}
\end{table}

\clearpage

\bibliography{references}